%% file: main-arxiv.tex
\documentclass{article}

\usepackage{microtype}
\usepackage{graphicx}
\usepackage{subfigure}
\usepackage{booktabs} % for professional tables
\usepackage{hyperref}
\usepackage{amsmath}
\usepackage{amssymb}
\usepackage{mathtools}
\usepackage{amsthm}
\usepackage{mathtools}
\usepackage{bbm}
\usepackage{blkarray}
\usepackage{bm}
\usepackage{setspace}
\usepackage{multicol}
\usepackage[capitalize,noabbrev]{cleveref}
\usepackage[textsize=tiny]{todonotes}
\usepackage{thm-restate}
\usepackage{subcaption}
\usepackage{mwe}
\DeclarePairedDelimiter\ceil{\lceil}{\rceil}

\DeclareMathOperator*{\argmin}{arg\,min}

\newcommand{\calf}{\mathcal{F} }
\newcommand{\calc}{\mathcal{C} }

\newcommand{\cala}{\mathcal{A} }

\newcommand{\calr}{\mathcal{R} }

\newcommand{\nat}{\mathbb{N} }
\newcommand{\poly}{\mathrm{poly}}
\newcommand{\eps}{\epsilon}
\newcommand{\caly}{\mathcal{Y}}
\newcommand{\calx}{\mathcal{X}}
\newcommand{\crasp}{\textsf{C-RASP}{}}
\newcommand{\ind}{\mathbbm{1}}
\newcommand{\ps}{\mathtt{ps}}

\newcommand{\xax}{\mathsf{x}}
\newcommand{\yax}{\mathsf{y}}
\newcommand{\clo}{\operatorname{cl}}
\newcommand{\ri}{\operatorname{ri}}
\newcommand{\intr}{\operatorname{int}}
\newcommand{\conv}{\operatorname{conv}}

\newcommand{\act}{\mathbb{A}}
\newcommand{\mci}{\cala_{\textsf{mci}}}
\newcommand{\mcirc}{\cala_{\textsf{mci}}^{\calr,\calc}}
\newcommand{\dfa}{\textsf{DFA}}
\newcommand{\cfg}{\textsf{CFG}}
\newcommand{\lcfg}{\textsf{L-CFG}}

\theoremstyle{plain}
\newtheorem{theorem}{Theorem}[section]

\newtheorem{lemma}[theorem]{Lemma}
\newtheorem{corollary}[theorem]{Corollary}
\theoremstyle{definition}
\newtheorem{definition}[theorem]{Definition}

\theoremstyle{remark}
\newtheorem{remark}[theorem]{Remark}

\usepackage{natbib}
\usepackage[noend]{algorithmic}
\usepackage{algorithm}
\usepackage{fancyhdr}
\usepackage[margin=1in]{geometry}
\usepackage{changepage}
\title{Non-Asymptotic Length Generalization}

\author{%
Thomas Chen \\
Stanford University\\
\texttt{tchen@cs.stanford.edu} \\
\and
Tengyu Ma \\
Stanford University \\
\texttt{tengyuma@stanford.edu}\\
\and
Zhiyuan Li \\
Toyota Technological Institute at Chicago\\
\texttt{zhiyuanli@ttic.edu} 
}

\date{}

\begin{document}

\maketitle

\begin{abstract}

\input{sections/abstract-page}

\end{abstract}

\input{sections/new_intro}

\input{sections/related-work}

\input{sections/foundations-setup}

\input{sections/main-results}

\input{sections/proof-sketch-new}

\input{sections/conclusion}

\input{sections/acknowledgements}

\newpage
\bibliography{bibliography}
\bibliographystyle{icml2025}

%%%%%%%%%%%%%%%%%%%%%%%%%%%%%%%%%%%%%%%%%%%%%%%%%%%%%%%%%%%%%%%%%%%%%%%%%%%%%%%
%%%%%%%%%%%%%%%%%%%%%%%%%%%%%%%%%%%%%%%%%%%%%%%%%%%%%%%%%%%%%%%%%%%%%%%%%%%%%%%
% APPENDIX
%%%%%%%%%%%%%%%%%%%%%%%%%%%%%%%%%%%%%%%%%%%%%%%%%%%%%%%%%%%%%%%%%%%%%%%%%%%%%%%
%%%%%%%%%%%%%%%%%%%%%%%%%%%%%%%%%%%%%%%%%%%%%%%%%%%%%%%%%%%%%%%%%%%%%%%%%%%%%%%
\newpage
\appendix
\onecolumn
\input{sections/appendix}

%%%%%%%%%%%%%%%%%%%%%%%%%%%%%%%%%%%%%%%%%%%%%%%%%%%%%%%%%%%%%%%%%%%%%%%%%%%%%%%
%%%%%%%%%%%%%%%%%%%%%%%%%%%%%%%%%%%%%%%%%%%%%%%%%%%%%%%%%%%%%%%%%%%%%%%%%%%%%%%

\end{document}

%% file: sections/abstract-page.tex
Length generalization is the ability of a learning algorithm to learn a hypothesis which generalizes to longer inputs than the inputs in the training set. In this paper, we provide provable guarantees of length generalization for various classes of functions in an idealized setting. First, we formalize the framework of non-asymptotic length generalization, which requires a computable upper bound for the minimum input length that guarantees length generalization, as a function of the complexity of ground-truth function under some given complexity measure. We refer to this minimum input length to length generalize as length complexity. We show the Minimum-Complexity Interpolator learning algorithm achieves optimal length complexity. We further show that whether a function class admits non-asymptotic length generalization is equivalent to the decidability of its language equivalence problem, which implies that there is no computable upper bound for the length complexity of Context-Free Grammars. On the positive side, we show that the length complexity of Deterministic Finite Automata is $2n - 2$ where $n$ is the number of states of the ground-truth automaton. Our main results are upper bounds of length complexity for a subset of a transformer-related function class called \crasp{} \citep{yang2024countingliketransformerscompiling}. We show that the length complexity of 1-layer \crasp{} functions is  $O(T^2)$ when the ground-truth function has precision $T$, and that the length complexity of 2-layer \crasp{} functions is $O(T^{O(K)})$ when the ground-truth function has precision $T$ and $K$ heads.

%% file: sections/new_intro.tex
\section{Introduction}
\label{sec:intro}

The generalization of a trained model from shorter inputs seen during training time to longer inputs seen at inference time is a phenomenon called length generalization. The question of when length generalization is possible is an important question in the area of language modeling and reasoning with Large Language Models (LLMs) \citep{brown2020languagemodelsfewshotlearners}. Real-world language understanding often requires handling longer contexts that exceed training-time input lengths, such as long Chain-of-Thought for reasoning problems, multi-turn conversations, lengthy documents, or complex code. Many factors limit the ability to train on long sequences directly, such as the increasing computational cost and memory requirement of training on longer sequences and the fact that long sequences are rare. Successful length generalization does not only allow efficient training for good long-context performance, but also is a natural test on whether the model has robustly learned the underlying language or reasoning patterns.

% \cite{bhattamishra2020abilitylimitationstransformersrecognize}, \cite{shaw2021compositionalgeneralizationnaturallanguage}, and \cite{ruoss2023randomizedpositionalencodingsboost}

%\cite{bhattamishra2020abilitylimitationstransformersrecognize}, 

Prior works study length generalization in a controlled setting where Transformers \citep{vaswani2023attentionneed} are trained on algorithmic tasks from scratch. In this context, length generalization has been empirically observed for certain algorithmic tasks but not others, and it appears that transformers cannot length generalize for most tasks. Tasks that have been tested are arithmetic tasks like integer addition \citep{nogueira2021investigatinglimitationstransformerssimple, nye2021workscratchpadsintermediatecomputation, anil2022exploringlengthgeneralizationlarge, zhou2024transformersachievelengthgeneralization}, formal-language tasks like $\textsc{parity}$ \citep{anil2022exploringlengthgeneralizationlarge}, algorithmic tasks in the Chomsky Hierarchy \citep{shaw2021compositionalgeneralizationnaturallanguage, delétang2023neuralnetworkschomskyhierarchy, ruoss2023randomizedpositionalencodingsboost}, copying tasks, tasks involving a sequence of integers such as sorting or finding the mode \citep{zhou2023algorithmstransformerslearnstudy}, and deducing the end-assignment of a variable in a block of code \citep{anil2022exploringlengthgeneralizationlarge}. Some of these tasks do not exhibit length generalization in their vanilla form, but do exhibit length generalization if certain modifications are made to the learning setup. These modifications include modifying the input and output format \citep{zhou2023algorithmstransformerslearnstudy, zhou2024transformersachievelengthgeneralization}, adding positional embeddings \citep{jelassi2023lengthgeneralizationarithmetictransformers}, and adding access to a scratchpad \citep{nye2021workscratchpadsintermediatecomputation}. 

On the theoretical side, \cite{gold} provides a foundational result in the theory of length generalization, which is originally stated in a more broad context of language identification. Let $\Sigma$ be a finite alphabet, let a hypothesis be a mapping from all strings $\Sigma^*$ to $\{0, 1\}$, and let $\mathcal{F}$ be the hypothesis class. \citet{gold} roughly says that there is a learning algorithm $\cala$ such that for all hypotheses in $\calf$, $\cala$ will eventually learn the ground-truth hypothesis, when the training set inputted to $\cala$ contains (string, label) pairs for all the strings up to a sufficiently large length. Gold's result holds for all hypothesis classes $\mathcal{F}$ satisfying the mild assumption that $\mathcal{F}$ can be enumerated by a Turing machine, including Regular languages, Context-Free languages, decision problems that can be solved by finite-precision transformers with or without Chain-of-Thought, etc. The learning algorithm $\cala$ simply returns the first hypothesis in the enumeration of $\calf$ that correctly labels all training examples. When this enumeration lists the functions in $\calf$ in order of increasing complexity for some complexity measure, then we call this learning algorithm the Minimum-Complexity Interpolator, denoted by $\mci$.

% Unfortunately, in practice, transformers struggle to length generalize on even very simple arithmetic and logic tasks,such as computing parities, integer addition, and variable assignment [Anil et al., 2022; Kazemnejad et al., 2023], despite achieving perfect accuracy on short training examples. The gap between positive theoretical result by Gold, 1967 and poor practical performance could of course be explained by the fact that the learning algorithm is not $\mci$ but training transformers using gradient-based methods, which leads to issues as non-realizability (the ground truth hypothesis is cannot be expressed by transfomers uniformly [cite preetum RASP-L and follow-up]) or returning interpoloting hypothesis with high complexity. 

The gap between the wide range of function classes which Gold's theory predicts length generalization for and the limited classes of functions that transformers empirically length generalize on presents an opportunity to develop a more predictive theory of length generalization. There are many potential reasons for this gap, such as the discrepancy between the learning algorithm $\mci$ and the gradient-based learning algorithms used in practice. However, a more fundamental issue of Gold's result is that the guarantee is inherently \emph{asymptotic} --- it does not provide any information on the minimum input length required to achieve length generalization.  It is completely possible that even if gradient-based training methods like SGD are able to find the minimum-complexity interpolating hypothesis, length generalization still only happens when the training set contains impractically long length inputs. In this case, it may appear empirically that length generalization does not occur at all, despite the asymptotic theoretical guarantee. 

Intuitively, the minimum input length required to achieve length generalization should be a function of the complexity of the ground truth hypothesis, and better length generalization is expected for simpler hypotheses. To both get a more useful length generalization guarantee and better understand the limit of length generalization, the goal of this paper is to answer the following question: 
\begin{quote}
\emph{What is the minimum input length required to achieve length generalization as a function of complexity of ground-truth hypothesis, assuming we have infinite computational resources?}
\end{quote}

\paragraph{Our Contributions:} This paper provides a more fine-grained analysis of length generalization, by providing a \emph{non-asymptotic} guarantee on the minimum input length required to achieve length generalization, as a function of complexity of ground-truth. More specifically, we make the following contributions:
\begin{itemize}
    \item In \Cref{sec:foundations}, we introduce the framework of \emph{non-asymptotic length generalization} (\Cref{defn:nonasympt-lgen}), which requires a non-asymptotic upper bound for the minimum input length that guarantees length generalization, given the complexity of ground-truth function under some given complexity measure $\calc$. The latter we call the length complexity under $\calc$ (\Cref{defn:length-complexity}). We show that the Minimum-Complexity Interpolator learning algorithm $\mci$, instantiated with complexity measure $\calc$, is the optimal with respect to the length complexity under $\calc$. As a concrete example, we show that $\mci$ only needs inputs up to length $2c-2$ to learn any ground-truth Deterministic Finite Automata (DFA) of $c$ states.
    \item In \Cref{sec:characterization}, we show whether a hypothesis class (with a default encoding system) admits non-asymptotic length generalization is equivalent to whether the problem of deciding equivalence of two finite descriptions of hypotheses is decidable. As a consequence, Context-Free Grammars (CFGs) do not admit non-asymptotic length-generalization, though they admit (asymptotic) length-generalization in the limit. % This implies non-asymptotic length generalization is a strictly stronger property than asymptotic length generalization (or length generalization in the limit).\Cref{??}. In other words, length generalization for CFG eventually happens, but there does not exist a computable upper bound for the minimum required length of training data for length generalization to happen.
    \item In \Cref{sec:crasp}, we prove non-asymptotic length generalization for a subset of the transformer-related hypothesis class, \crasp{} \citep{yang2024countingliketransformerscompiling}. \crasp{} is a super-set of functions that can be expressed by transformers. Variants of the RASP function class, like $\textup{RASP-L}$ and \crasp{}, have been recently shown to have a good prediction power on the length generalization performance of transformers \citep{zhou2023algorithmstransformerslearnstudy, huang2024formalframeworkunderstandinglength}. We study two subclasses of \crasp{}, $\crasp{}^1$ (\Cref{thm:crasp1}) and $\crasp{}^2$ (\Cref{thm:main-weaker}), which are of depth $1$ and $2$ respectively. 
\end{itemize}

\iffalse 
\zhiyuan{@thomas consider removing theorem 1.1 and 1.2}
\begin{theorem} (Informal Version of Theorem \ref{thm:crasp1}) There exists a learning algorithm which learns the class of $1$-layer \crasp{}  programs in a length generalizable way. In particular, let $\calc$ be a mapping which takes in a $1$-layer \crasp{}  program $p$ and returns the precision of $p$'s parameters. If the ground-truth function's parameters has precision $T$, then given all inputs of length at most $O(T^2)$ and their labels, the Minimum-Complexity Interpolator equipped with complexity measure $\calc$ can learn a hypothesis which length generalizes to arbitrary length inputs.
\end{theorem}

\begin{theorem} (Informal Version of Theorem \ref{thm:main-weaker}) There exists a learning algorithm which learns the class of $2$-layer \crasp{}  programs in a length generalizable way. In particular, let $\calc$ be a mapping which takes in a $2$-layer \crasp{}  program $p$ and returns the precision of $p$'s parameters exponentiated by the number of heads in $p$. If the ground-truth function has $K$ heads and precision $T$ parameters, then given all inputs of length at most $O(T^{O(K)})$ and their labels, the Minimum-Complexity Interpolator equipped with complexity measure $\calc$ can learn a hypothesis which length generalizes to arbitrary length inputs.
\end{theorem}
\fi 
\begin{table}[ht]
    \centering
    \begin{tabular}{|c|c|c|c|}
      \hline
      Function Class & Complexity of Ground-Truth Function & Length of Training Data Sufficient to Generalize\\ \hline
      DFAs & number of states, $c$ & $2c - 2$ (Proposition \ref{prop:dfa-len-gen})  \\ \hline
      CFGs & description length, $c$ & no computable bound in $c$ exists (Proposition \ref{prop:cfg-len-gen})\\ \hline
      $\crasp{}^1$ & precision, $T$ & $O(T^2)$ (Theorem \ref{thm:crasp1}) \\ \hline
      $\crasp{}^2$ &  precision, $T$, and number of heads, $K$ & $O(T^{O(K)})$ (Theorem \ref{thm:main-weaker}) \\ \hline
    \end{tabular}
    \caption{Summary of results: upper bounds on minimum length of binary strings in training data which suffices for length generalization. C-RASP is a class of functions which is a superset of functions expressible by finite-precision transformers. For $L \in \{ 1,2 \}$, $\crasp{}^L$ is a subclass of depth-$L$ $\crasp{}$ functions. }
  \end{table}

%[Papers related to improving length generalization by ensuring realizability]

%[Papers related to improving length generalization by reducing the complexity of hypothesis class]\tc{not sure what these two are}

%% file: sections/related-work.tex
\section{Related Work}

\citet{solomonoff} proposed Bayesian-inference-based algorithms which when given a sequence of symbols, predict the next symbol according to the posterior distribution computed from the Solomonoff-Levin prior distribution. \citet{gold} introduced Identification-in-the-Limit as an asymptotic notion of learnability and proved that many classes of functions can be learned in this sense. 

\citet{zhou2023algorithmstransformerslearnstudy} propose the RASP-L Conjecture, which says that whether a transformer length generalizes on a particular ground-truth function $f_*$ is well predicted by whether $f_*$ has a short RASP-L description. \citet{huang2024formalframeworkunderstandinglength} formulate the problem of length generalization--and the RASP-L Conjecture--formally as a version of Identification-in-the-Limit and prove asymptotic results for identifying languages expressible by Limit Transformers. None of the works above provide non-asymptotic guarantees. We distinguish this work from previous work by providing non-asymptotic bounds on the length of the training data required in order to guarantee that the learner outputs a single hypothesis which exhibits perfect length generalization. %We are interested in the setting where the learner can take in the ground-truth complexity $c$ and know in advance what length of data it needs to length generalize, as opposed a setting where the learner outputs an infinite sequence of hypotheses for each training length, without knowing at which point the hypotheses will start to length generalize.

\cite{weiss2021thinkingliketransformers} \cite{zhou2023algorithmstransformerslearnstudy}, \cite{yang2024countingliketransformerscompiling}, and \cite{shaw2024altacompilerbasedanalysistransformers} study programming languages which capture the set of functions which transformers can express (like RASP). % \cite{zhou2023algorithmstransformerslearnstudy} proposes and provides empirical evidence for the RASP-L Conjecture, which says that ground-truth functions which are expressible by short RASP-L programs are also learnable in a length generalizable way.

Regarding theoretical works for length generalization not related to transformers, \citet{marsden2024provablelengthgeneralizationsequence} prove length generalization for learning linear dynamical systems with SGD.  \citet{abbe2024generalizationunseenlogicreasoning} prove out-of-domain generalization for boolean functions of a fixed input size. 

% \zhiyuan{mention asymptotic guarantess in RASP papers}

% Regarding theoretical work on the expressivity and limitations of transformers, there is \cite{hahn-2020-theoretical}, \cite{chiang2022overcomingtheoreticallimitationselfattention}, \cite{merrill2023logicexpressinglogprecisiontransformers}, \cite{liu2023transformerslearnshortcutsautomata}, \cite{sanford2024transformersparallelcomputationlogarithmic}, \cite{li2024chainthoughtempowerstransformers}, and \cite{hahn-rofin-2024-sensitive}. 

% Additionally, \citet{shaw2021compositionalgeneralizationnaturallanguage} learn Synchronous CFGs on various tasks via a hybrid neural and grammar based learning approach.

% We came up with our results independently from \cite{huang2024formalframeworkunderstandinglength}, and we were not aware of their results until late in our project. \tc{does this violate anonymity?}

%% file: sections/foundations-setup.tex
\section{Non-Asymptotic Length Generalization}\label{sec:foundations}

\paragraph{Notation.} For simplicity, we will fix the alphabet $\Sigma = \{ 0,1\}$ throughout the paper. We define computable (recursive) functions as functions which can be computed by a Turing Machine, which halts on all inputs. We say a function $f:\nat \to \nat$ is \emph{computably bounded} (recursively bounded) if there exists a computable (recursive) function $g:\nat \to \nat$ such that $f(n) \leq g(n)$ for all $n \in \nat$. Let $\langle M \rangle \in \{0,1\}^*$ denote the binary string encoding of $M$. Let $[N] := \{ 1,2,3,\ldots, N - 1, N\}$. Denote $\{ 0,1\}^{n}$ as the set of binary strings of length $n$, $\{ 0,1\}^{\leq n} := \bigcup_{j = 0}^{n} \{ 0,1\}^{j}$, and $\{ 0,1\}^{*} := \bigcup_{j \geq 0} \{ 0,1\}^{j}$. Note that $\eps$, the empty string of length $0$, is included in $\{ 0,1\}^{*}$. Denote $\ind [\cdot]$ as the indicator function and $\clo(A)$ as the closure of set $A \subset \mathbb{R}^d$. 

\paragraph{Representation of Functions.} We are interested in learning subsets of computable functions mapping $\{ 0,1\}^*$ to $\{ 0,1\}$, denoted by $\calf$. Because a learning algorithm can only return a finite description of the hypothesis it selects, we need to define the representation of the hypothesis and the corresponding encoding system below in \Cref{defn:encoding-system}.

\begin{definition}[Encoding System]\label{defn:encoding-system}
    An \emph{encoding system} is a Turing Machine $\calr$ which on input of a finite string $p$ (which can be thought as the code or description of a function), outputs the Turing Machine description of the computable function $f:\{0,1\}^*\to \{0,1\}$ represented by $p$. Here we denote the description of the Turing Machine as $\langle \calr(p) \rangle$ and the function $f$ as $\calr(p)$.
    We use $\calf^\calr$ to denote the function class implicitly defined by the encoding system $\calr$, i.e. $\calf^\calr = \{ \calr(p) : p \in \{ 0,1\}^*\}$.
\end{definition}

Often the standard encoding system is only defined on valid inputs. For convenience we define the encoding system $\calr$ for all inputs in $\{ 0,1\}^*$ and map invalid inputs to the empty language.

As examples of encoding systems, DFAs and CFGs are two encoding systems for regular languages and context-free languages, respectively.

% \begin{restatable}{definition}{dfa}
% An $n$-state Deterministic Finite Automata (DFA) is a function $\{ 0,1\}^* \to \{ 0,1\}$ specified by a start state $q_0 \in Q$, accepting states $A \subset Q$, and a transition function $\delta : Q \times \{ 0,1\} \to Q$. Denote $\delta^* : Q \times \{ 0,1\}^* \to Q$ as the extension of $\delta$ to any finite string. A string $s$ is accepted iff $\delta^*(q_0, s) \in A$. 
% \end{restatable}

\begin{restatable}{definition}{dfadefi}\label{def:dfa}
    An $n$-state Deterministic Finite Automaton (DFA) is a tuple $M = (Q = [n], \Sigma = \{0,1\}, \delta, q_0, F)$ where $Q = [n]$ is the set of states, $\Sigma = \{0,1\}$ is the input alphabet, $\delta: Q \times \Sigma \to Q$ is the transition function, $q_0 \in Q$ is the start state, and $F \subseteq Q$ is the set of accepting states. The encoding system $\calr_\dfa$ is a Turing Machine that reads $\langle M \rangle$ and outputs the description of a Turing Machine that simulates $M$. For any input string $x \in \{0,1\}^*$, the language characterized by $M$ is where $\calr_\dfa(\langle M \rangle)(x) = 1$ if and only if $\delta^*(q_0, x) \in F$, where $\delta^*: Q \times \{0,1\}^* \to Q$ is the natural extension of $\delta$ to strings defined recursively as $\delta^*(q,\eps) = q$ and $\delta^*(q,sa) = \delta(\delta^*(q,s),a)$ for any $q \in Q$, $s \in \{ 0,1\}^*$, and $a\in \{ 0,1\}$.
\end{restatable}

\begin{restatable}{definition}{cfgdefi}\label{defn:cfg}
    A Context-Free Grammar (CFG) is a tuple $G = (N, T = \{0,1\}, P, S)$ where $N$ is a finite set of nonterminal symbols, $T = \{0,1\}$ is the terminal alphabet, $P$ is a finite set of production rules of the form $A \to \alpha$ where $A \in N$ and $\alpha \in (N \cup T)^*$, and $S \in N$ is the start symbol. Let $\langle G \rangle \in \{0,1\}^*$ denote the binary string encoding of $G$. The encoding system $\calr_\cfg$ is a Turing Machine that reads $\langle G \rangle$ and outputs the description of a Turing Machine that simulates $G$. For any input string $x \in \{0,1\}^*$, the language characterized by $G$ is $\calr_\cfg(\langle G \rangle)(x) = 1$ if and only if $x$ can be derived from $S$ by applying a finite sequence of production rules from $P$. The membership problem for CFGs (determining whether $x$ can be derived from $S$) is decidable and can be solved efficiently using the Cocke-Younger-Kasami (CYK) algorithm \citep{hopcroft-ullman}, which runs in time $O(|x|^3\cdot |G|)$ for CFGs in Chomsky Normal Form.
\end{restatable}

Often we are only interested in learning a subset of all programs $A$ given a certain encoding system $\calr$, where learning can be significantly easier. Our setup can cover this case by considering the new encoding system $\calr_A$ which is defined as $\calr_A(p) = \calr(p)$ for those $p \in A$ and $\calr_A(p)$ be the Turing Machine that outputs the empty language for those $p \notin A$. We use linear CFG as a concrete example below.

\begin{restatable}{definition}{lcfgdefi}\label{defn:lcfg}
    A Linear CFG is a CFG $G = (N,T=\{0,1\},P,S)$ where each production rule in $P$ has at most one nonterminal symbol on its right-hand side. We denote the encoding system for linear CFGs as $\calr_\lcfg$, which is defined as $\calr_\lcfg(p) = \calr_\cfg(p)$ for those $p$ representing a linear CFG $G$, and $\calr_\lcfg(p)$ be the Turing Machine that outputs the empty language for those $p$ representing a non-linear CFG or non-CFG. We denote the function class for linear CFGs as $\calf_\lcfg$.
\end{restatable}

% $\calr$ is a mapping from a string $p \in \{ 0,1\}^*$ to the language computed by the CFG that $p$ represents, with invalid strings $p$ mapping to the empty language. 

\paragraph{Learning Setup.} With the ground truth function being denoted $f^*$ or $\calr(p)$, we define the labeled training dataset consisting all data up to length $N$ as below, for all $N\in\mathbb{N}$:
\begin{align*}
   D_N(f^*) := \{ (x, f^*(x)) : x \in \{ 0,1\}^*, |x| \leq N\} \text{ and } D_N(p) := D_N(\calr(p)).
\end{align*}

With $\calr$ as the encoding system, an adversary picks any ground-truth function $f^* \in \calf^\calr$. A learning algorithm is a Turing Machine $\cala$ which takes as input a training set $D_N(f^*)$ and outputs some $\hat{p} \in \{ 0,1\}^*$. We say a learning algorithm $\cala$ \emph{length-generalizably learns} a function $f^*$ at input length $N$ w.r.t.  encoding system $\calr$ iff $\calr(\cala(D_N(f^*))) = f^*$. Now, define the following asymptotic notion of learnability. % We may omit the encoding system $\calr$ when it is clear from the context.

% For instance, one way of assigning a finite description to a regular language $L$ is with the transition table of the minimal DFA which computes $L$. In this way, CFGs are also finite descriptions of Context-Free Languages (CFLs). Regarding $\crasp{}^1$ (resp. $\crasp{}^2$), one natural finite description are the parameters of the $\crasp{}^1$ (resp. $\crasp{}^2$) program, defined in Definitions \ref{defn:crasp1} (resp. \ref{defn:crasp21}). It is important to make this distinction because the learning algorithms can only output a finite description of the hypothesis it selects, not the hypothesis itself. 

\begin{definition}[Length Generalization in the Limit, adapted from \citet{gold}]
     A function class $\calf\subseteq \calf^\calr$ admits \emph{length generalization in the limit} w.r.t.  encoding system $\calr$ if there exists a learning algorithm $\cala$ such that for all $f^* \in \calf$, there exists a natural number $N$ such that for all $N' \geq N$, $\cala$ length-generalizably learns $f^*$ at input length $N'$.
\end{definition}

The above definition of length generalization in the limit is a special case of the so-called \textit{identification in the limit} in the informant model in \cite{gold}. The major difference is that \cite{gold} requires the function class $\calf$ to be learnable in arbitrary order of data presentation, while in our case, we are only interested in a particular order of data presentation, namely the order of increasing input length. 

\begin{restatable}[Adapted from Theorem I.4 of \citet{gold}]{proposition}{lengeninlimit}
\label{prop:asympt-id}
For all encoding systems $\calr$, the function class $\calf^\calr$ admits length generalization in the limit, and thus so does any function class $\calf\subseteq \calf^\calr$.
\end{restatable}

This is not in contradiction with results in \cite{gold}, since the conditions on $\calr$ excludes identifiability in the limit for the set of all recursive functions. For completeness, we provide a proof of the above proposition in \Cref{appen:length-gen-in-the-limit}. 

\paragraph{Complexity Measure.} We are interested in understanding the minimum length of training data which suffices for length generalization. To this end, a complexity measure for the functions in $\calf$ is necessary. Our approach is to assume a complexity measure $\calc: \{ 0,1\}^*\to \nat$ which assigns a complexity to each representation $p$, and define the complexity of the function $f$ as the minimum complexity of any representation of $f$ in $\calr$, namely $\calc^\calr(f) := \min_{p \in \{ 0,1\}^*, \calr(p) = f} \calc(p)$. We might drop the superscript $\calr$ and just use $\calc$ for function complexity when $\calr$ is clear from the context. We will make the following mild assumption that $\calc$ is reasonably simple throughout the paper, unless otherwise stated.

\begin{restatable}[]{assumption}{asmptEexists}\label{asmpt:E-exists}
$\calc$ is computable and there exists a Turing Machine $E$ which can enumerate programs $p \in \{ 0,1\}^*$ in an non-decreasing order of $\calc(p)$. In particular,  the range of $E$ is $\{ 0,1\}^*$ and $\forall i \leq i'$, $\calc(E(i)) \leq \calc(E(i'))$. In addition, $\calc$ is such that for each $c \in \nat$, $|\{ p \in \{ 0,1\}^* : \calc(p) \leq c\}| < \infty$.
\end{restatable}

% \zhiyuan{@thomas: I changed DFA CFG notation to align with our encoding system}

This assumption is easily satisfied by a standard choice of $\calc$, namely where $\calc(p)$ returns the length of $p \in \{ 0,1\}^*$. There is little loss in generality in just thinking of $\calc$ as this standard complexity measure. Having a general $\calc$ provides some extra flexibility and makes the results easier to understand.  % \zhiyuan{leaves this to introduce MCI and say it becomes MDL}

\begin{definition}[Complexity Measures for DFAs and CFGs]\label{defn:calc-dfa-cfg}
    The complexity measure $\calc_\dfa$ for DFAs maps a DFA $M=(Q=[n], \Sigma, \delta, q_0, F)$ (represented by $\langle M \rangle$) to $n$, the number of states in $Q$. The number of states of any DFA is within a logarithmic factor of the length of representation of that DFA in bits ($n\log n$). For CFGs, given a CFG $G = (N, T, P, S)$, let $|P|$ be the total length of the production rules in $P$. The complexity measure of CFG $G$ is $\calc_\cfg(\langle G \rangle) = |P| + |N| + |T|$, which is within a constant factor of the length of $\langle G \rangle$ in bits.
\end{definition}

Now we are ready to define the notion of non-asymptotic length generalization.

\begin{restatable}[Non-Asymptotic Length Generalization]{definition}{nonasymptoticlgen}\label{defn:nonasympt-lgen}
A function class $\calf\subseteq \calf^\calr$ admits \emph{non-asymptotic length generalization} w.r.t.  encoding system $\calr$ and complexity measure $\calc$ if there exists a learning algorithm $\cala$ and a computable function $\hat{N}^{\calr, \calf}_\cala: \nat \to \nat$ such that for all $f^* \in \calf$ and for all $N' \geq \hat{N}^{\calr,\calf}_\cala(\calc^\calr(f^*))$, $\cala$ length-generalizably learns $f^*$ at input length $N'$.
\end{restatable}

\paragraph{Length Complexity.} For notation simplicity, we define the length complexity of a function $f^*$ for a learning algorithm $\cala$ w.r.t. encoding system $\calr$, $N^{\calr}_\cala(f^*)$, as the minimum length of training data which suffices for the learning algorithm $\cala$ to length generalize on $f^*$.
\begin{align}\label{eq:N-cala-f}
    N^{\calr}_\cala(f^*) &:= \min\{ N \geq 0 : \forall n \geq N, \calr(\cala(D_n(f^*))) = f^*\} 
\end{align}

This quantity is $\infty$ if there is no such $N$ where $\forall n \geq N$, $\calr(\cala(D_n(f^*))) = f^*$.  We also define the length complexity of functions in $\calf \subset \calf^\calr$ up to complexity $c$ for a learning algorithm $\cala$ w.r.t. encoding system $\calr$, $N^{\calr, \calf}_\cala(c)$, as the maximum length complexity of any function in $\calf \subset \calf^\calr$ with complexity at most $c$.
\begin{align}\label{eq:N-cala-f-calf}
    N^{\calr, \calf}_\cala(c) &:= \max_{f^* \in \calf \text{ s.t. }  \calc^\calr(f^*) \leq c} N_\cala^{\calr}(f^*) = \max_{p \in \{ 0,1\}^* \text{ s.t. }  \calc(p) \leq c \wedge \calr(p)\in\calf} N^{\calr}_\cala(\calr(p)).
\end{align}
In particular, we denote $N^{\calr, \calf^\calr}_\cala(c)$ by $N^{\calr}_\cala(c)$ for convenience. We omit the superscript $\calr$ when it is clear from context.

The definition of length generalization in the limit and non-asymptotic length generalization can be rewritten using length complexity as follows.
\begin{itemize}\setlength{\itemsep}{0pt}\setlength{\parskip}{0pt}
\item \textbf{Length generalization in the limit:} $\exists$ learning algorithm $\cala$ such that $\forall f^* \in \calf$, $  N^{\calr}_\cala(f^*) <\infty$.
\item \textbf{Non-asymptotic length generalization}: $\exists$ learning algorithm $\cala$ such that $ N^{\calr,\calf}_\cala$ is upper bounded by some computable function from $\mathbb{N}$ to $ \mathbb{N}$, \emph{i.e.}, $N^{\calr,\calf}_\cala$ is computably bounded.
\end{itemize}

As a concrete example, \Cref{prop:dfa-len-gen} shows that DFAs admits non-asymptotic length generalization w.r.t. the standard encoding system $\calr_\dfa$ and complexity measure $\calc_\dfa$ returning the number of states in the input DFA.  Its proof is in \Cref{appen:chomsky}.
% \zhiyuan{@thomas: mention where we can find the proof and the high-level idea/the main existing technical result behind it.}

% For instance, the class of regular languages with encoding system $\calr$ given by the DFA encoding system and complexity measure given by the number of states of the DFA satisfies \Cref{defn:nonasympt-lgen} as follows.

\begin{restatable}[Non-Asymptotic Length Generalization for DFAs]{proposition}{dfalgen}
\label{prop:dfa-len-gen}
Let $\calr_\dfa$ be the DFA encoding system defined in \Cref{def:dfa}, and let $\calc_\dfa$ be the number of states in DFA. Regular languages $\calf^\calr$ admits non-asymptotic length generalization w.r.t. encoding system $\calr_\dfa$ and complexity measure $\calc_\dfa$. More specifically, there exists a learning algorithm $\cala$ such that $N^{\calr_\dfa}_\cala(c) \leq 2c - 2$ for all $c \in \nat$.
\end{restatable}

% Because there are two parameters $K$ and $T$ for each such $2$-layer $\crasp{}$ program, it is not immediately clear how to define a complexity measure for $\crasp{}^{2}$, as a complexity measure must map functions in $\crasp{}^{2}$ to natural numbers. We choose $\calc(f) = T(f)^{K(f)}$ for proving our main results, though other complexity measures are possible. We note that the description length of a $\crasp^{1,T}$ program is $O(\log T)$ while the description length of a $\crasp^{2,K,T}$ program is $O(K \log T)$, so that the complexity measures we chose for $\crasp{}^1$ and $\crasp{}^2$ are the exponential of a constant times the upper bound on the description length.

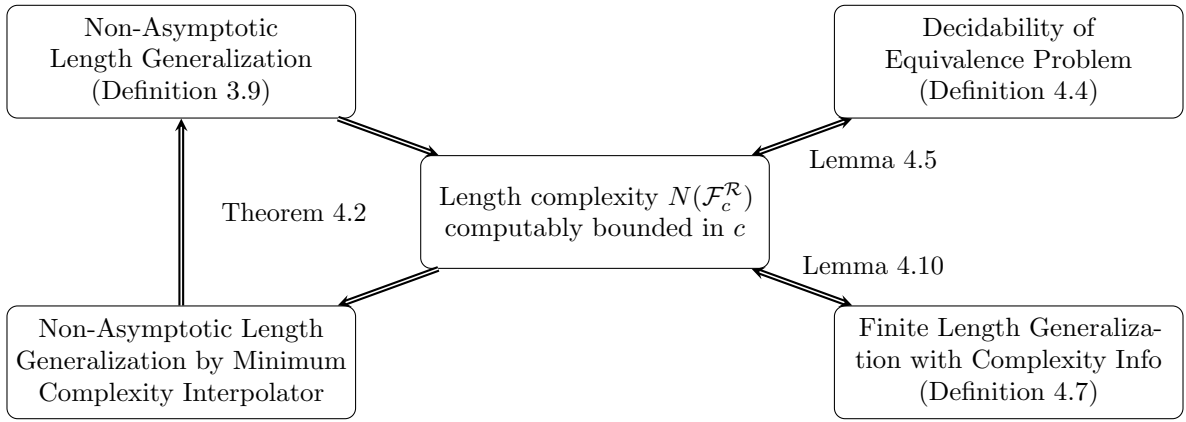
\begin{figure}[t]
    \centering
    \vspace{-0.8cm}
    \begin{tikzpicture}[
        node distance=4.5cm,
        box/.style={draw, rounded corners, text width=4.4cm, align=center, minimum height=1.5cm}
    ]
        % Nodes
        \node[box] (nonasympt) at (-5.5,2) {Non-Asymptotic Length Generalization\\(\Cref{defn:nonasympt-lgen})};
        \node[box] (mci) at (-5.5,-2) {Non-Asymptotic Length Generalization by Minimum Complexity Interpolator};
        \node[box] (decidable) at (5.5,2) {Decidability of\\ Equivalence Problem\\(\Cref{defn:lang-equiv-prob})};
        \node[box] (finite) at (5.5,-2) {Finite Length Generalization with Complexity Info\\ (\Cref{defn:finite-identification})};
        \node[box] (nfc) at (0,0)  {Length complexity $N(\calf^\calr_c)$ computably bounded in $c$};

        % Add text for Lemma reference
        \node at (-4,0) {\Cref{lem:opt-learner}};
        \node at (3.7,0.7) {\Cref{lem:equiv-decidability}};
        \node at (3.7,-0.7) {\Cref{lem:equiv-finite-identification}};

        % Arrows to/from middle node
        \draw[<->, double, thick, >=stealth] (nfc) -- (finite);
        \draw[<->, double, thick, >=stealth] (nfc) -- (decidable);
        \draw[->, double, thick, >=stealth] (nfc) -- (mci);
        \draw[->, double, thick, >=stealth] (mci) -- (nonasympt);
        \draw[->, double, thick, >=stealth] (nonasympt) -- (nfc);

    \end{tikzpicture}
    \caption{Summary of equivalence results between different characterizations of length generalization in \Cref{sec:characterization} under some mild simplicity assumptions on the complexity measure $\calc$~(\Cref{asmpt:E-exists}). Each arrow represents an implication proven by the corresponding theorem.}
    \label{fig:equiv-relationships}
\end{figure}

\section{Characterization of Non-Asymptotic Length Generalization}\label{sec:characterization}

% \zhiyuan{Have a figure showing the equivalence relationships between the different characterizations of length generalization. around each of arrow mention the theorem that proves the equivalence.}

In \Cref{sec:characterization}, we characterize the conditions under which a function class admits non-asymptotic length generalization w.r.t. an encoding system $\calr$ and complexity measure $\calc$ satisfying \Cref{asmpt:E-exists}. In \Cref{subsec:mci-optimality}, we define the Minimum Complexity Interpolator $\mcirc$ and show its optimality in terms of length complexity. Therefore, a function class $\calf^\calr$ admits non-asymptotic length generalization  if and only if the function class can be length-generalizably learned by $\mcirc$. In \Cref{subsec:equiv-to-decidability}, we show that a function class admits non-asymptotic length generalization if and only if its language equivalence problem for encoding system $\calr$ is decidable. In \Cref{subsec:finite-identification}, we show equivalence between non-asymptotic length generalization to a variant of ``finite identification" proposed in \citet{gold}. The results are summarized in \Cref{fig:equiv-relationships}.

Before diving into the details of the characterization, we first introduce Definition \ref{defn:length-complexity}, an algorithm-independent version of length complexity. This contrasts with Equation \ref{eq:N-cala-f-calf}, which is dependent on the learning algorithm $\cala$. We introduce this second notion of length complexity as we will show in \Cref{lem:opt-learner} that it coincides with the optimal length complexity, over all learning algorithms $\cala$.

\begin{definition}[Length Complexity of Function Class]\label{defn:length-complexity}
    Given a function class $\calf$, we define the length complexity of $\calf$ as the minimum input length that can distinguish any two functions in $\calf$:
    \begin{align*}
        N(\calf) := \min \{ n \in \nat : \forall f\neq f' \in \calf,\exists x \in \{ 0,1\}^{\leq n} \textrm{ s.t. } f(x) \neq f'(x)\}.
    \end{align*} 
\end{definition}

It is easy to see that $N(\calf)$ is finite if and only if $\calf$ is finite. Let $\calf^\calr_{c} := \{ f \in \calf : \calc^\calr(f) \leq c\}$ be the set of functions in $\calf^\calr$ of complexity at most $c$. The length complexity of $\calf^\calr_{c}$, $N(\calf^\calr_{c})$ will be the core quantity we will study in this section. As we will see in \Cref{lem:opt-learner}, $N(\calf^\calr_{c})$ is the optimal length complexity over all learning algorithms $\cala$, $\min_\cala N^{\calr}_\cala(c)$. We further note that if $\calc$ satisfies \Cref{asmpt:E-exists} then $\calf^\calr_{c}$ is a finite set for each $c \in \nat$.

\subsection{Minimum Complexity Interpolator and Its Optimality}\label{subsec:mci-optimality}

\begin{algorithm}[tb]
   \caption{Minimum-Complexity Interpolator ($\mcirc$)}
   \label{alg:min-c-interp}
\begin{algorithmic}
   \STATE {\bfseries Hyperparameters:} Complexity Measure $\calc$, encoding system $\calr$
%    \STATE {\bfseries Input:} Training Data $D_N(f^*) := \{ (x, f^*(x)) : \forall |x| \leq N\}$ for some $N \in \nat$ and unknown function $f^*:\{0,1\}^*\to \{0,1\}$.
%    \STATE {\bfseries Output:} $\argmin_{p \in \{ 0,1\}^* : \forall (x,y) \in D_N(f^*), y = \calr(p)(x) }  \calc(p)$
   \STATE {\bfseries Input:} Finite Training Dataset  $S \subseteq \{ (x, y)\mid x \in \{ 0,1\}^*,y\in\{0,1\} \}$.
   \STATE {\bfseries Output:} $\argmin_{p \in \{ 0,1\}^* : \forall (x,y) \in D_N(f^*), y = \calr(p)(x) }  \calc(p)$
\end{algorithmic}
\end{algorithm}

The Minimum-Complexity Interpolator (\Cref{alg:min-c-interp}) is the main learning algorithm which we study in this work. Although it would be ideal to study a learning algorithm closer to what is used empirically to train transformers, we study the Minimum-Complexity Interpolator to abstract away the complex training dynamics of gradient-based methods like SGD, which are very non-trivial even for training 2 layer neural networks \citep{mahankali2023ntkvanillagradientdescent}.  We also acknowledge the limitation that the Minimum-Complexity Interpolator is computationally intractable for general function classes and encoding systems. For instance, it was shown that the problem of finding the minimum-state DFA for a regular language is \textsf{NP}-hard \citep{min-interpolator-np-complete}.

We denote the Minimum-Complexity Interpolator learning algorithm by $\mcirc$ for short when the encoding system is $\calr$ and complexity measure is $\calc$. When the complexity measure $\calc$ is the length of the program, $\mcirc$ is just the famous Minimum Description Length (MDL) algorithm \citep{Rissanen1978ModelingBS}. $\mcirc$ has the nice property that it is the best possible algorithm over all learning algorithms in minimizing $N^{\calr, \calf}_\cala(c)$, in the following sense.

\begin{restatable}[Optimality of Minimum-Complexity Interpolator]{theorem}{optmci}
\label{lem:opt-learner}
Given any encoding system $\calr$ and complexity measure $\calc$\footnote{Note here \Cref{asmpt:E-exists} is not needed for the optimality of $\mcirc$ to hold. When \Cref{asmpt:E-exists} is not satisfied, all three quantities are infinity}, for all $c \in \nat$, it holds that $N^{\calr}_{\mcirc}(c) = \min_\cala N^{\calr}_\cala(c) = N(\calf^\calr_{c})$. % \zhiyuan{@thomas: it is fine for all three quantities to be $\infty$. Also this does not assume \Cref{asmpt:E-exists} on the existence of $E$. feel free to remove this comment if you also think it is ok here.}

As a consequence, the following three statements are equivalent:
\begin{itemize}\itemsep0em
    \item Function class $\calf^\calr$ admits non-asymptotic length generalization;
    \item Function class $\calf^\calr$ admits non-asymptotic length generalization, via learning algorithm $\mcirc$;
    \item For all $c \in \nat$, length complexity of $\calf^\calr_{c}$, $N(\calf^\calr_{c})$, is computably bounded in $c$.
\end{itemize}
\end{restatable}
% \zhiyuan{@thomas: we need to use the fact that $\calf_c$ is finite. I add the condition that $n_c<\infty$. Please check proof sketch and the proof.}

\Cref{lem:opt-learner} follows from two observations. First, $N^{\calr}_{\mcirc}(c) \ge \min_\cala N^{\calr}_\cala(c) \ge  N(\calf^\calr_{c})$. That is, no algorithm can perfectly length generalize using only inputs of length strictly smaller than $N(\calf^\calr_{c})$, because there will be some pair $f \neq f' \in \calf^\calr_{c}$ that are both consistent with the training data. The learner cannot deduce which one is the actual ground truth. Whichever it guesses, it will be wrong in the case that the other function is the actual ground truth. Second, $N^{\calr}_{\mcirc}(c)\le N(\calf^\calr_{c})$. That is, $\mcirc$ can perfectly length generalize given all inputs of length $N(\calf^\calr_{c})$. This is because every function $f \in \calf^\calr_{c}$ is distinguishable from every other function $f' \in \calf^\calr_{c}$ on some input of length at most $N(\calf^\calr_{c})$. In particular, $f^*$ will be distinguished from every function of lower or equal complexity on the training set. The proof of \Cref{lem:opt-learner} is in \Cref{appen:foundations}.

Though the above optimality result of $\mcirc$ holds for all complexity measures $\calc$, $\mcirc$ might actually not be computable (implementable by some Turing Machine) without any restriction on the complexity measure. For example, it is well-known that Kolmogorov complexity is not computable \citep{k-complexity}. \cref{lem:mci_computable} shows that $\mcirc$ is indeed computable under \Cref{asmpt:E-exists}. 
\begin{lemma}\label{lem:mci_computable}
    Under \Cref{asmpt:E-exists}, \Cref{alg:min-c-interp} is computable and thus a valid learning algorithm. 
\end{lemma}
\textit{Proof of \Cref{lem:mci_computable}.}
By \Cref{asmpt:E-exists}, we use TM $E$ to enumerate all programs $p \in \{ 0,1\}^*$ in non-decreasing order of $\calc(p)$ and at each iteration check if $\calr(p)$ is consistent with the training data. If it is, we output $p$ and stop. Otherwise, we continue to the next program. Both the enumeration and checking-consistency procedures are computable, as assumed by \Cref{asmpt:E-exists} and \Cref{defn:encoding-system}. $\blacksquare$

In the remaining sections, we may omit $\calr$ and $\calc$ in the superscripts of $\cala_{\textup{mci}}^{\calr, \calc}$ when they are clear from context for convenience. 

\subsection{Equivalence to Decidability of Language Equivalence Problem}\label{subsec:equiv-to-decidability}

The main goal of this paper is to understand when length generalization is possible, namely seeking for concrete upper bounds for the minimum length of training data which suffices for length generalization, the length complexity $N^{\calr}_\cala(c)$. Surprisngly, even though such upper bounds always exist (length generalization in the limit $\implies N^{\calr}_\cala(c)<\infty, \forall c\in \nat $), such upper bounds are not always computable. In this section, we show that there exist computable upper bounds on the length complexity if and only if the language equivalence problem for encoding system $\calr$ is decidable. 

\begin{restatable}[Language Equivalence Problem]{definition}{langequivprob}\label{defn:lang-equiv-prob}
The \textit{Language Equivalence Problem} for encoding system $\calr$ is the computational problem where given any $p,q \in \{ 0,1\}^*$, determine whether $\calr(p) = \calr(q)$.
\end{restatable}

\begin{restatable}[Equivalence to Decidability of Language Equivalence Problem]{lemma}{equivweak}\label{lem:equiv-decidability}
    For any encoding system $\calr$ and complexity measure $\calc$ satisfying \Cref{asmpt:E-exists},  the Language Equivalence problem for $\calr$ is decidable if and only if length complexity of $\calf^\calr_c$, $N(\calf^\calr_c)$, is computably bounded in $c$. Thus it is also equivalent to the property that $\calf^\calr$ admits non-asymptotic length generalization.
\end{restatable}

The proof of \Cref{lem:equiv-decidability} is in \Cref{appen:foundations}. As a consequence, we have the following impossibility result for non-asymptotic length generalization for a special case of CFGs: linear CFGs. Comparing \Cref{prop:asympt-id} and \Cref{prop:cfg-len-gen}, we see that CFG serves as a concrete example for the separation between length generalization in the limit and non-asymptotic length generalization. 

\begin{restatable}[(linear) CFGs only admit length generalization in the limit]{proposition}{cfglgen}
\label{prop:cfg-len-gen}
Recall $\calr_\lcfg$ is the encoding system for linear CFGs defined in \Cref{defn:lcfg} and $\calc_\cfg(\langle G \rangle)$ is the complexity measure that maps a CFG $G = (N,T,P,S=\{0,1\})$ to $|N| + |T| + |P|$. Then  for any learning algorithm $\cala$, the length complexity, $ N_{\cala}^{\calr_{\lcfg}}$, is not computably bounded. That is, linear CFGs do not admit non-asymptotic length generalization (w.r.t. standard CFG encoding system $\calr_\cfg$), and neither does the set of all CFGs.
\end{restatable}

Below we give a proof sketch of \Cref{prop:cfg-len-gen} to illustrate the general idea of why non-decidability of language equivalence problem implies that non-asymptotic length generalization is not possible. The full proof of \Cref{prop:cfg-len-gen} is in \Cref{appen:chomsky}.

% \zhiyuan{@thomas: I think the proof sketch is already formal and complete. if you agree we can remove the proof in the appendix, which is kind of broken now. }
\textit{Proof sketch of \Cref{prop:cfg-len-gen}.} Determining the equivalence of two linear CFGs is well-known to be undecidable~\citep{baker-book-linear-cfg}. Suppose there is a learning algorithm $\cala$ with its length complexity $ N_{\cala}^{\calr_{\lcfg}}$ upper bounded by computable function $N$, this would give a computable algorithm to decide whether two given Linear CFGs $G_1, G_2$ have the same language: let $c := \max(\calc_\cfg(G_1), \calc_\cfg(G_2))$ and check whether $G_1, G_2$ agree on all inputs of length at most $N(c)$. If they do not completely agree, clearly $G_1$ and $G_2$ have different languages. If $G_1, G_2$  agree on all inputs of length at most $N(c)$, by \Cref{lem:opt-learner}, $N(\calf^{\calr_{\lcfg}}_{c}) = \min_{\cala'} N^{\calr_{\lcfg}}_{\cala'}(c) \le  N_{\cala}^{\calr_{\lcfg}}(c) \le N(c)$, which implies that $G_1$ and $G_2$ must correspond to the same language.

\iffalse
\begin{proof}[Proof sketch of \Cref{prop:cfg-len-gen}]
    \zhiyuan{@thomas: a one-line or two-line sketch is useful here in arixv version.  (or maybe teo-line full proof?) Suppoer algorithm A and computable upper bound exists, check whether two functions agree on all inputs of length less than the upper bound. If not, obviously they are different. if they agree, they are the same function otherwise A cannot distingush them. }
\end{proof}
\fi 

% \subsection{Connection to Finite Identification, \cite{gold}}\label{subsec:finite-identification}
\subsection{Connection to Finite Identification, \citep{gold}}\label{subsec:finite-identification}
Our notion of non-asymptotic length generalization is related to Gold's notion of Finite Identification~\citep{gold}, which is restated as follows in the context of length generalization. 

\begin{restatable}[Finite Length Generalization (``Identification'',~\cite{gold})]{definition}{finitelgen}\label{defn:finite-identification}
    A function class $\calf\subseteq \calf^\calr$ admits \emph{Finite Length Generalization} w.r.t. encoding system $\calr$ if there exists a Turing Machine (TM) $\cala$, where for any $f_* \in \calf$, on input of a training set $D_N(f_*)$ for some $N\in \nat$, $\cala$ satisfies that:
    \begin{enumerate}
        \item $\cala$ can only output either ``pass" or some program $\hat{p} \in \{ 0,1\}^*$. $\cala$ outputs $\hat{p}$ at least for one $N\in\nat$.
        \item Whenever $\cala$ outputs some program $\hat{p} \in \{ 0,1\}^*$, it must be correct in the sense that $\calr(\hat{p}) = f_*$.
    \end{enumerate}
\end{restatable}

It is clear that Finite Length Generalization implies Length Generalization in the limit --- one can replace ``pass'' by arbitrary program $\hat{p}$ and sticks to the same (correct) output $\hat{p}$ once $\cala$ outputs any program $\hat{p}$. Intuitively, a function class admits Finite Length Generalization if there exists a learning algorithm which can perfectly length generalize at some finite input length (and where the learning algorithm knows the length at which it length generalizes), rather than only generalizing in the limit.
Thus, Finite Length Generalization is a very desirable property. However, it is in general too good to be true. We give a characterization of Finite Length Generalization in \Cref{lem:char-finitelgen}. As a consequence of \Cref{lem:char-finitelgen}, functions classes as simple as the set of all languages that are finite do not admit Finite Length Generalization.

\begin{restatable}[Characterization of Finite Length Generalization]{lemma}{charfinitelgen}\label{lem:char-finitelgen}
    A function class $\calf^\calr$ admits Finite Length Generalization w.r.t. encoding system $\calr$ if and only if for any $f_* \in \calf^\calr$, there exists a natural number $N$ such that $f^*$ is the only function that is consistent with the training set $D_N(f_*)$.
\end{restatable}

The proof of \Cref{lem:char-finitelgen} is straightforward and omitted.

Since Finite Length Generalization is in general too restrictive, we relax the definition to allow the learning algorithm to output a program $\hat{p}$ with some information on the complexity measure of ground truth $f_*$, namely some $c \ge \calc^\calr(f_*)$. This leads to the definition of Finite Length Generalization with Complexity Information in \Cref{defn:prop1}. Interestingly, \Cref{defn:prop1} is equivalent to non-asymptotic length generalization (\Cref{defn:finite-identification}).

\begin{restatable}[Finite Length Generalization with Complexity Information]{definition}{finiteidentprop}\label{defn:prop1}
    A function class $\calf\subseteq \calf^\calr$ admits \emph{Finite Length Generalization} w.r.t. encoding system $\calr$ and complexity measure $\calc$ if there exists a Turing Machine (TM) $\cala$, which for any $f_* \in \calf$, on input of a training set $D_N(f_*)$ for some $N\in \nat$ and \textbf{a natural number $c \ge \calc^\calr(f_*)$}, it satisfies that:
    \begin{enumerate}
        \item $\cala$ can only output either ``pass" or some program $\hat{p} \in \{ 0,1\}^*$. $\cala$ outputs $\hat{p}$ at least for one $N\in\nat$.
        \item Whenever $\cala$ outputs some program $\hat{p} \in \{ 0,1\}^*$, it must be correct in the sense that $\calr(\hat{p}) = f_*$.
    \end{enumerate}
\end{restatable}

\begin{restatable}[Equivalence of Finite Length Generalization Definitions]{lemma}{equivfinitelgen}\label{lem:equiv-finite-identification}
    For any encoding system $\calr$ and complexity measure $\calc$ satisfying \Cref{asmpt:E-exists}, function class $\calf^\calr$ admits Finite Length Generalization with Complexity Information w.r.t. $\calr$ and $\calc$  if and only if length complexity of $\calf^\calr_c$, $N(\calf^\calr_c)$, is computably bounded in $c$. Thus it is also equivalent to $\calf^\calr$ admits non-asymptotic length generalization.
\end{restatable}

We defer the proof of \Cref{lem:equiv-finite-identification} to \Cref{appen:foundations}.

% \Cref{defn:nonasympt-lgen} is also equivalent to \Cref{defn:prop1} in \Cref{appen:foundations}, which says that there exists an algorithm which takes in a training set $D_N(f_*)$ and the true ground-truth complexity $\calc^\calr(f_*)$. It is only allowed to output ``pass" or some $\hat{p} \in \{ 0,1\}^*$, but when it outputs $\hat{p} \in \{ 0,1\}^*$, it must be that $\calr(\hat{p}) = f_*$. Thus, such an algorithm must be able to decide when the input training set has data of sufficiently long length in order for it to deduce $f_*$ correctly from $\calf^\calr$. This is related to \citet{gold}'s notion of Finite Identification. 

% The proof of Lemma \ref{lem:equiv-decidability} and the equivalence to \Cref{defn:prop1} is in \Cref{appen:foundations}, and the proof of Proposition \ref{prop:cfg-len-gen}  is in \Cref{appen:chomsky}.

%% file: sections/main-results.tex
\section{Main Results: Non-Asymptotic Length Generalization of $\crasp{}$}\label{sec:crasp}

\subsection{Recap: Definition of $\crasp{}$} Our main results pertain to a class of functions called \crasp{}. \crasp{} is a variant of RASP that, with alphabet $\Sigma = \{ 0,1\}$, defines a class of functions from $\{0,1\}^*$ to $ \{0,1\}$, where only certain sequence-to-sequence operations are permitted \citep{yang2024countingliketransformerscompiling}. \crasp{} was shown to be a subset to the class of functions expressible by transformers with infinite-precision activations and a superset to that expressible by finite-precision transformers \citep{yang2024countingliketransformerscompiling}.

A sequence is an element of $\nat^*$. A \crasp{} program consists of a finite number of operations, where each operation is an $m$-ary sequence-to-sequence mapping ($m \in \nat$), which take $m$ input sequences $\nat^*$ and returns a single output sequence $ \nat^*$. For any operation, the length of its input and output sequences are always the same. The entire \crasp{} program must be a mapping $\{0,1\}^*$ to $ \{0,1\}^*$. As a convention, we will take the last bit of the output sequence as the output bit, yielding a function $\{0,1\}^*$ to $ \{0,1\}$. Each intermediate sequence of the $\crasp{}$ program can be either boolean-valued $(\in \{ 0,1\}^*)$ or count-valued $(\in \nat^*)$. \Cref{defn:crasp} lists the operations which are allowed in a \crasp{} program.

\newcommand{\cttn}[2]{\ensuremath{\textsc{\textbf{\#}} \left[ #1 \right] \; #2}}
\newcommand{\cttntwo}[2]{\ensuremath{\textsc{\textbf{\#}$_2$} \left[ #1 \right] \; #2}}
\newcommand{\cif}[3]{\ensuremath{#1\;\mathbf{?}\; #2\; \textbf{:} \;#3}}
\newcommand{\cifop}{\ensuremath{\mathbf{?}}}

% \zhiyuan{@thomas: I copied this from Yang et al. 2024. Still needs to argue why our CRASP class is significant subclass. Maybe split operations into two types: lcoal and non-local(ps). Say describe depth in a intuitive way, but more clearly than it is now. focus on depth equal to 1 and 2. When depth is 2, width arise as another parameter. I feel like it does not make sense to define width for a general CRASP, also too hard. }
\begin{definition}[$\crasp{}$, \citep{yang2024countingliketransformerscompiling}]\label{defn:crasp}
    A $\crasp{}$ program over alphabet $\Sigma = \{ 0,1\}$ is defined as a series of $n < \infty$ $\crasp{}$ operations. We denote the $h^{(1)},\ldots, h^{(n)}$ as the output sequences of each $\crasp{}$ operation. Denote $x \in \{ 0,1\}^*$ as the input sequence to the \crasp{} program. Denote $h^{(i)}_j$ as the $j$th element of the sequence $h^{(i)}$, where $i \in [n]$ and $j \in [|x|]$. There are two types of operations:
   
   \begin{center}
   \begin{tabular}{p{0.45\textwidth}p{0.45\textwidth}}
   \begin{center}
       \textbf{Boolean-Valued Operations}
   \end{center}
   \begin{tabular}{ll}
       \toprule
        \textbf{Initial} & $h^{(i)}_j:=\ind[x_j = a]$ for $a\in\{ 0,1\}$ \\
        \midrule
        \textbf{Boolean} & $h^{(i)}_j:=\lnot h^{(i')}_j$ \\
        & $h^{(i)}_j:=h^{(i')}_j\land  h^{(i'')}_j$\\
        \midrule
        \textbf{Sign} & $h^{(i)}_j:= \ind [h^{(i')}_j > 0]$\\
        \midrule
        \textbf{Constant} & $h^{(i)}_j:=1$\\
        \bottomrule
   \end{tabular}
   &
   \begin{center}
       \textbf{Count-Valued Operations}
   \end{center}
   \begin{tabular}{ll}
       \toprule
        \textbf{Partial Sum} & $h^{(i)}_j:=\ps(h^{(i')})_j$ \\
        \midrule
       \textbf{Conditional} & $h^{(i)}_j:=\cif{h^{(i')}_j}{h^{(i'')}_j}{h^{(i''')}_j}$\\
       \midrule
       \textbf{Addition} & $h^{(i)}_j:=h^{(i')}_j+h^{(i'')}_j$\\
       \midrule
       \textbf{Subtraction} & $h^{(i)}_j:=h^{(i')}_j-h^{(i'')}_j$\\
       \midrule
       \textbf{Min/Max} & $h^{(i)}_j:=\min(h^{(i')}_j,h^{(i'')}_j)$\\
            & $h^{(i)}_j:=\max(h^{(i')}_j,h^{(i'')}_j)$\\
       \midrule
       \textbf{Constant} & $h^{(i)}_j:=1$\\
       \bottomrule
   \end{tabular}
   \end{tabular}
   \end{center}
   Where the partial-sum operator $\ps : \{ 0,1\}^* \to \nat^*$ is defined as: $\forall h \in \{ 0,1\}^*, \forall j \in [|x|], \ps(h)_j = \sum_{l \in [j]}h_l.$
\end{definition}

There are several natural parameters that contribute to a measure of complexity of $\crasp{}$ functions. First, we study the impact of the following notion of precision of the parameters of $\crasp{}$ functions on the difficulty of length generalization.

\begin{restatable}[$p$-precision]{definition}{precision}
\label{defn:precision}
An integer of absolute value at most $p$ is of $p$-precision. A rational number between $[0,1]$ is of $p$-precision if in simplest form, where the numerator and denominator are relatively prime, its denominator is at most $p$ in magnitude. A tuple of rational numbers in $[0,1]$ is precision $p$ if the least common denominator of its entries is at most $p$ in magnitude. 
\end{restatable}

This notion of precision makes sense in the context of $\crasp{}$ since $\crasp{}$ only allows integer combinations of previously computed variables instead of arbitrary linear combinations of previously computed variables.

We also study the impact of depth of \crasp{} programs on the difficulty of length generalization where depth is, loosely, the maximum number of sequentially-applied $\ps$ operations along any part of the $\crasp{}$ program. We study depth-1 and depth-2 programs, which we will also refer to as 1-layer and 2-layer programs. Note that for convenience, we will use functions $f \in \calf^\calr$ and descriptions $p \in \{ 0,1\}^*$ interchangeably in the following exposition.

We will first prove length generalization results pertaining to the following subset of 1-layer $\crasp{}$, which we call $\crasp{}^1$.

\begin{restatable}[$\crasp{}^1$]{definition}{onelayercrasp}
\label{defn:crasp1}
With integer $T$, let $\crasp{}^{1, T}$ denote the set of programs of the following form. Each program $f$ has parameters $a,b,d \in [-T, T]$, $a > 0$. For any $n > 0$, on input $x \in \{ 0,1\}^n$, $f$ computes: 
    \begin{align*}
        f(x) = \ind [a \cdot \ps (x)_n -b \cdot n - d > 0]
    \end{align*}
    Then, $\crasp{}^1 = \bigcup_{T \geq 1} \crasp{}^{1, T}$. Given a function $f \in \crasp{}^{1}$, the complexity measure $\calc(f) = \max(|a|, |b|, |d|)$, the precision of $f$'s parameters in the sense of \Cref{defn:precision}.
\end{restatable}

Here, the encoding system $\calr$ maps a string $p$ to a $\crasp{}^1$ function. The string $p$ is interpreted by $\calr$ to encode $3$ parameters, each taking $\Theta(\log T)$ bits to encode. Invalid encodings are mapped to a default $\crasp{}^1$ function. Thus, the complexity measure proposed above, which takes in $p$ and returns the maximum precision of its parameters, returns an integer which is roughly exponential in the length of $p$.

Our main length generalization result will apply to a subset of 2-layer $\crasp{}$, which we call $\crasp{}^{2}$. For 2 layer programs, the width $K$ of the first layer becomes a natural parameter to study.

\begin{restatable}[$\crasp{}^2$]{definition}{twolayercrasp}
\label{defn:crasp21}
With integers $T$ and $1 \leq K \leq T^2$, let $\crasp{}^{2,K,T}$ be the set of programs of the following form. Each program $f$ has parameters $0 < z \leq T$, $\forall i \in [K]$, $a^{(i)}, b^{(i)}, \lambda_i \in \{ -T, \ldots, T\}$, with $a^{(i)} > 0$. We require that for all $i \in [K]$, $\frac{b^{(i)}}{a^{(i)}} \in (0,1)$ and is distinct from $\frac{b^{(i')}}{a^{(i')}}$ for $i' \neq i$. We also require $\sum_{i \in [K]}\lambda_i > z$.
    
    For any $n > 0$, on input $x \in \{ 0,1\}^n$, the first layer computes the values of $K$ heads, $\{h^{(i)}\}_{i \in [K]}$, on the $n$ prefixes of $x$: $\{ \{ x_1\}, \{ x_1, x_2\}, \ldots, \{ x_1, \ldots, x_n\}\}$. Subscript $j$ indicates the value of a quantity on the $j$th prefix of $x$.
    \begin{align*}
        \forall j \in [n], \forall i \in [K], h_j^{(i)} &= \ind [\ps(x)_j > \frac{b^{(i)}}{a^{(i)}} j]\\
    \end{align*}
    The second layer computes the output, which is the $n$th bit of the final sequence.
    \begin{align*}  
        f(x) &= \ind [\sum_{i \in [K]} \lambda_i \ps(h^{(i)})_n > z \cdot n]
    \end{align*}
    Then, $\crasp{}^2 = \bigcup_{1 \leq T, 1 \leq K \leq T^2} \crasp{}^{2, K, T}$. Given a function $f \in \crasp{}^2$, let $K(f)$ be the number of heads, $h^{(i)}$, in the first layer of $f$ and let $T(f) := \max(\max_{i \in [K(f)]} |a^{(i)}|, \max_{i \in [K(f)]} |b^{(i)}|, \max_{i \in [K(f)]} |\lambda_i|, |z|)$. The complexity measure is then $\calc(f) = T(f)^{K(f)}$, the precision of the function's parameters to the power of the number of heads.
\end{restatable}

We will refer to functions in $\crasp^{2,K,T}$  as a $2$ layer, $K$-head, $T$-precision program, where each intermediate variable $h^{(i)}$ is a head. 

Here, the encoding system $\calr_{\crasp{}^2}$ maps a string $p$ to a $\crasp{}^2$ function. The string $p$ is interpreted by $\calr_{\crasp{}^2}$ to encode $\Theta(K)$ parameters, each taking $\Theta(\log T)$ bits to encode, so that the entire encoding of a $\crasp{}^{2,K,T}$ function is roughly $\Theta(K \log T)$ bits long. Invalid encodings are mapped to a default $\crasp{}^2$ function. Thus, the complexity measure $\calc(f) = T(f)^{K(f)} = \exp(K(f) \log T(f))$ returns an integer which is roughly exponential in the length of $p$.

% We will now state our main results. Recall that two functions $f, f'$ are not equal if there exists at least one string $x$ that distinguishes them: $f(x) \neq f'(x)$. 

\subsection{Non-Asymptotic Length Generalization of $\crasp{}^{1}$ and $\crasp{}^{2}$} The following result states that in order to identify the ground-truth function from the set $\crasp{}^{1}$ where the precision of parameters is at most $T$, it suffices for the Minimum-Complexity Interpolator to receive $(\textup{string}, \textup{label})$ pairs for all strings of length at most $O(T^2)$.

\begin{restatable}[$\crasp{}^1$ Length Generalization]{theorem}{thmonelayer}
\label{thm:crasp1}
Let $\calf = \crasp{}^1$ and $\calc(f) = \max(|a|, |b|, |d|)$, defined in \Cref{defn:crasp1}. Then $\forall T \in \nat,$ we have $N_{\mci}(T) \leq O(T^2)$. That is, the Minimum-Complexity Interpolator, with complexity $\calc$ and function class $\calf$, can length generalize given inputs of length $O(T^2)$ when the ground-truth has complexity $T$.
\end{restatable}

% Theorem \ref{thm:crasp1} shows that the set of $1$-layer, $T$-precision \crasp{} programs can be learned in a length generalizable way with inputs of length $O(T^2)$.

The following is our main length generalization result. It states that in order to identify the ground-truth function from the set $\crasp{}^{2}$ where the precision of parameters is at most $T$ and number of heads is at most $K$, it suffices for the Minimum-Complexity Interpolator to receive $(\textup{string}, \textup{label})$ pairs for all strings of length at most $O(T^{O(K)})$. 

\begin{restatable}[$\crasp{}^2$ Length Generalization, Main Result]{theorem}{thmmainweaker}
\label{thm:main-weaker}
Let $\calf = \crasp{}^2$ and $\calc(f) = T(f)^{K(f)}$, defined in \Cref{defn:crasp21}. Then if the ground-truth function $f_*$ has $T(f_*) \leq T$ and $K(f_*) \leq K$, then the Minimum-Complexity Interpolator, with complexity $\calc$ and function class $\calf$, can length generalize given inputs of length $O(T^{O(K)})$.
\end{restatable}

By Theorem \ref{thm:main-weaker}, $\crasp^{2,K,T}$ can be learned in a length generalizable way with inputs of length $O(T^{O(K)})$. By Theorem \ref{thm:crasp1}, $\crasp^{1,T}$ can be learned in a length generalizable way with inputs of length $O(T^2)$. Our upper bound on the minimum length required to learn $\crasp{}^{2, K, T}$ is much larger than that of $\crasp{}^{1, T}$. We don't have a lower bound on the length of inputs required to identify $\crasp{}^{2, K, T}$, but our best guess is that one of the form $\Omega(T^K)$ exists in the regime $K \leq \Theta(\log T)$.

The proof of Theorem \ref{thm:crasp1} can be found in \Cref{appen:crasp1}. Below we sketch the proof of Theorem \ref{thm:main-weaker}.

% \tc{put a table summarizing the 4 results}

%% file: sections/proof-sketch-new.tex
\section{Proof Sketch of Main Result, \Cref{thm:main-weaker}.}

Theorem \ref{thm:main-weaker} follows as a Corollary to the following, stronger Theorem \ref{thm:main}, which says that the Minimum Complexity Interpolator with complexity measure $\calc(f) = T(f)^{K(f)}$ will length generalize if it receives inputs of length $N_{\mci}(\alpha) \leq O(\alpha^{O(1)})$, when the ground-truth function has complexity at most $\alpha$.

\begin{restatable}{theorem}{thmmain}
\label{thm:main}
Let $\calf = \crasp{}^2$ and $\calc(f) = T(f)^{K(f)}$, defined in \Cref{defn:crasp21}. Then $\forall \alpha \in \nat,$ we have $N_{\mci}(\alpha) \leq O(\alpha^{O(1)})$. 
\end{restatable}

\textit{Proof of Theorem \ref{thm:main-weaker}.} Every function $f$ in $\crasp{}^{2,K,T}$ has at most $K$ heads and $T$ precision, and so $\calc(f) \leq T^K =: \alpha_*$. Thus, Theorem \ref{thm:main} implies that inputs of length at most $N_{\mci}(\alpha_*)\leq O(\alpha_*^{O(1)}) = O(T^{O(K)})$ suffices to learn $\crasp{}^{2,K,T}$. $\blacksquare$

It remains to prove Theorem \ref{thm:main}. The full proof of Theorem \ref{thm:main} is in \Cref{appen:full-pf-thm}.  Below, we sketch the proof Lemma \ref{lem:main}, a weaker version of Theorem \ref{thm:main}, which says that in the same context as in Theorem \ref{thm:main}, $N_{\mci}(\alpha) \leq O(\alpha^{O(\log^2 \alpha)})$. Although weaker, the proof sketch of \Cref{lem:main} will still illustrate the main ideas of the full proof.

\begin{restatable}{lemma}{lemmain}
\label{lem:main}
Let $\calf = \crasp{}^2$ and $\calc(f) = T(f)^{K(f)}$, defined in \Cref{defn:crasp21}. Then $\forall \alpha \in \nat,$ we have $N_{\mci}(\alpha) \leq O(\alpha^{O(\log^2 \alpha)})$. 
\end{restatable}

To prove Lemma \ref{lem:main}, it suffices to prove Lemma \ref{lem:pf-sketch-suff-cond}. In particular, we will upper bound, for any  $T$ and $K$, the minimum length of inputs required to distinguish any two unequal $\crasp{}^2$ functions that have at most $K$ heads and $T$ precision. 

\begin{restatable}[Length Bound on $\crasp{}^{2,K,T}$]{lemma}{suffcondforpfsketch}\label{lem:pf-sketch-suff-cond}
For any $1 \leq T, 1 \leq K \leq T^2$, for all $f, f' \in \crasp{}^{2,K,T}$ such that $f \neq f'$, there exists a string $x_* \in \{ 0,1\}^*$ of length at most $O(T^{O(K^2)})$ such that $f(x_*) \neq f'(x_*)$.
\end{restatable}

\textit{Proof of Lemma \ref{lem:main}.} By \Cref{lem:opt-learner}, it suffices to show that for any two un-equal functions $f, f' \in \crasp^2$ such that $T(f)^{K(f)} \leq \alpha$ and $T(f')^{K(f')} \leq \alpha$, there is a string of length at most $ O(\alpha^{O(\log^2 \alpha)})$ which distinguishes them. If $T(f)^{K(f)} \leq \alpha$ and $T(f')^{K(f')} \leq \alpha$, then $T(f), T(f')$ are upper bounded as $T(f), T(f') \leq \alpha$, maximized when $K(f), K(f') = 1$. Meanwhile, $K(f), K(f')$ are upper bounded as $K(f), K(f') \leq \log \alpha$,  maximized when $K(f) = T(f)^2, K(f') = T(f')^2$. Thus, $f, f' \in \crasp{}^{2, \log \alpha, \alpha}$. Applying Lemma \ref{lem:pf-sketch-suff-cond} implies that there is a string of length $O(\alpha^{O(\log^2 \alpha)})$ distinguishing $f$ and $f'$. $\blacksquare$

The rest of the proof sketch will describe how to prove Lemma \ref{lem:pf-sketch-suff-cond}, and it will contain all the essential ideas to prove Theorem \ref{thm:main} directly. To do this, given two arbitrary $f, f' \in \crasp{}{}^{2,K,T}$ that are not equal, we show the existence of a \textit{short} string $x_*$ which distinguishes $f$ and $f'$, in the sense that $f(x_*) \neq f'(x_*)$. We will call such a string $x_*$ a distinguisher for $f, f'$.

\subsection{Key Definitions.}

Suppose $f, f' \in \crasp{}{}^{2,K,T}$, with parameters $( a^{(i)})_{i \in [K]}, ( b^{(i)})_{i \in [K]}, ( \lambda_i)_{i \in [K]}, z$ and $( (a^{(i)})')_{i \in [K]}$, $( (b^{(i)})')_{i \in [K]}$, $( \lambda_i')_{i \in [K]}$, $z'$, respectively. 

Suppose the set of unique numbers $R := \{ \frac{b^{(i)}}{a^{(i)}}\}_{i \in [K]} \cup \{ \frac{(b^{(i)})'}{(a^{(i)})'}\}_{i \in [K]} \subset (0,1)$ between the first layer of $f$ and $ f'$ has size $k,$ where $K \leq k \leq 2K$. We will refer to these numbers as ``slopes." We will denote $R = \{ s_j\}_{j \in [k]} \subset (0,1)$, where $s_1$ is the largest slope and $s_k$ is the smallest slope, and the slopes are sorted in descending order so that $s_1 > \ldots > s_k$. For $i \in [K]$, let $\textup{ord}(1, i) : [K] \to [k]$ be the index within $R$ of the $i$th slope of $f$, $\frac{b^{(i)}}{a^{(i)}}$. Let $\textup{ord}(2, i) : [K] \to [k]$ be the index within $R$ of the $i$th slope of $f'$, $\frac{(b^{(i)})'}{(a^{(i)})'}$.  In the following exposition, we will refer to ``line $j$" as the homogeneous, 2D line $\yax = s_j \xax$, with slope $s_j, j\in [k]$. We will denote ``line $j$" by the symbol $l_j$. We will call the set of $k \leq 2K$ unique slopes $\{ s_i\}_{i \in [k]}$ a configuration.

\begin{restatable}[Configuration]{definition}{configuration}
\label{defn:configuration}
A $(k,T)$-configuration is a set of $k$ distinct $T$-precision rational numbers $\{ s_i\}_{i \in [k]} \subset (0,1)$.
\end{restatable}

% Call a string $x \in \{ 0,1\}^*$ a distinguisher of \crasp{} programs $f, f' \in \calf$ if $f(x) \neq f'(x)$. 

We will refer to strings in $\{ 0,1\}^*$ synonymously as \textit{discrete test-functions}, because there is a one-to-one correspondence between $x$ and the sequence of 2D points $\{ (j, \ps(x)_j)\}_{j \in [|x|]} \subset \mathbb{R}^2$. The latter set of points acts as a ``test" of whether two programs $f, f' \in \crasp{}^{2,K,T}$ are different or not.

\begin{restatable}[Discrete Test-Function]{definition}{disctestfn}
\label{defn:disc-test-fn}
Given a $(k,T)$-configuration $\{ s_i\}_{i \in [k]}$, a discrete test-function $\calx$, with respect to $\{ s_i\}_{i \in [k]}$ and of  length $n < \infty$, is a function $\{ 0, 1, \ldots, n\} \to \{ 0, 1, \ldots, n\}$ where $\calx(0) = 0$ and $\forall j  \in [n]$, $\calx(j) = \calx(j - 1)$ or $\calx(j) = \calx(j - 1) + 1$. The induced activations $(B_1(\calx), \ldots, B_k(\calx))$ of $\calx$ with respect to the $(k,T)$-configuration are defined as:
\begin{align*}
    \forall i \in [k], \quad B_i(\calx) := \frac{1}{n}\sum_{j = 1}^n \ind [\calx(j) > s_i \cdot j]
\end{align*}
\end{restatable}

In our proof sketch, we will need to analyze properties of a continuous analog to discrete test-functions, which can be thought of as corresponding to infinite-length strings. 

\begin{restatable}[Continuous Test-Function]{definition}{cntstestfn}
\label{defn:cnts-test-fn}
Given a $(k,T)$-configuration $\{ s_i\}_{i \in [k]}$, a continuous test-function $\caly$, with respect to $\{ s_i\}_{i \in [k]}$, is a $1$-Lipschitz, monotone non-decreasing continuous function $[0,1] \to [0,1]$, with $\caly(0) = 0$. Continuous test-functions can only intersect the $k$ lines $\{ l_i\}_{i \in [k]}$ of slopes given by $\{ s_i\}_{i \in [k]}$ at finitely many points. The induced activations $(B_1(\caly), \ldots, B_k(\caly))$ of $\caly$ with respect to $\{ s_i\}_{i \in [k]}$ are defined as:
    \begin{align*}
        \forall i \in [k], \quad B_i(\caly) := \int_{0}^1 \ind [\caly(j) > s_i \cdot j] dj
    \end{align*}
\end{restatable}

To motivate this definition of continuous test-functions, note that there is a natural approximation property between discrete and continuous test-functions. For every discrete test-function $\calx$ of length $n$ w.r.t. some $(k,T)$-configuration, there is a continuous test-function $\caly$ whose activations are $O(\frac{\poly(k,T)\log n}{n})$-close in $L_\infty$-norm to $(B_i(\calx))_{i \in [k]}$.

\begin{align*}
    ||(B_i(\caly))_{i \in [k]} - (B_i(\calx))_{i \in [k]}||_\infty \leq O(\frac{\poly(k,T)\log n}{n})
\end{align*}

Conversely, for any ``$p$-precision" continuous test-function $\caly$ w.r.t. some $(k,T)$-configuration and for any $n \in \mathbb{Z}$ which is a sufficiently large multiple of $p$, there is a discrete test-function $\calx$ of length $n$ whose activations are $O(\frac{\poly(k,T)}{n})$-close in $L_\infty$-norm to $(B_i(\caly))_{i \in [k]}$.

The aforementioned approximation guarantees, in particular the second one (formalized in Lemma \ref{lem:disc-approx-informal}), are useful for our proof since activations of a continuous (resp. discrete) test-functions are a key property of continuous (resp. discrete) test-functions, as each $\crasp{}^2$ function can be thought of as a linear threshold function over the induced activations of the input string (test-function) with respect to the $(k,T)$-configuration given by the parameters of the first layer of the $\crasp{}^2$ function. We study continuous test-functions as a proxy for discrete test-functions, since the set possible activations $(B_i(\caly))_{i \in [k]}$ induced by continuous test-functions has a nice characterization (see Lemma \ref{lem:completeness-informal}).

The importance of the activations induced by continuous test-functions motivates the following ``decomposition" of continuous test-functions. Given a $(k,T)$-configuration $\{ s_i\}_{i \in [k]}$, we can think of a continuous test-function as a concatenation of partial functions which are defined on domains whose end-points are at the points where the continuous test-function intersects the $k$ lines: $\{ \yax = s_i \xax : i \in [k]\}$. We refer to each partial function as a \textit{segment} of the continuous test-function. Importantly, the lengths of the domains of these partial functions are sufficient to determine the activations that the continuous test-function induces. A segment is defined formally below.

\begin{restatable}[Segment]{definition}{segment}
    Given any $(k,T)$-configuration $\{ s_i\}_{i \in [k]} \subset (0,1)$, a \textit{segment} is a restricted test-function $S : [a,b] \to [0,1]$ where $[a,b] \subset [0,1]$ which maps a continuous subset $[a,b]$ to $[0,1]$. $S$ is $1$-Lipschitz and monotone non-decreasing. The segment's start-point $(a, S(a)) $ and the end-point $(b,S(b))$ each lie on one of the $k$ lines, in the sense that there exists some $i,j \in [k] $ where $S(a) = s_i \cdot a$ and $S(b) = s_j \cdot b$, where $i = j$ or $|i - j| = 1$. No other points $(x, S(x))$, $x \in (a,b)$ can lie on a line $l_1,\ldots, l_k$. 
\end{restatable}

Two different test-functions can share the same general ``blue-print," if one considers the subset of the $k$ lines the test-function crosses and the order in which the test-functions cross the $k$ lines. We define such ``blue-prints" formally as schema. For each schema, there are many test-functions which are of that schema.

\begin{restatable}[Schema]{definition}{schema}
Given any $(k,T)$-configuration $\{ s_i\}_{i \in [k]} \subset (0,1)$, a \textit{schema}  $Y$ is a blueprint for a continuous test-function, specifying a sequence of lines $\{ l_i\}_{i \in [k]}$ that any test-function of the schema must cross. It consists of an integer $0 < M < \infty$ and two tuples $\{ \textup{idx}(i)\}_{i \in [M]} \subset [k]^M, \{ \textup{sec}_i\}_{i \in [M]} \subset [k + 1]^M$, where $|\textup{idx}(i) - \textup{idx}(i + 1)| \leq 1$ for all $i \in [M - 1]$. If $|\textup{idx}(i) - \textup{idx}(i + 1)| = 1$, then $\textup{sec}_{i + 1}$ is unique and must be $\max(\textup{idx}(i), \textup{idx}(i + 1))$. If $\textup{idx}(i) = \textup{idx}(i + 1)$, then $\textup{sec}_{i + 1}$ can be either $\textup{idx}(i + 1)$ or $\textup{idx}(i + 1) + 1$. $\textup{sec}_1$ can be either $\textup{idx}(1)$ or $\textup{idx}(1) + 1$.

Any continuous test-function of schema $Y = (\{ \textup{idx}(i)\}_{i \in [M]}, \{ \textup{sec}_i\}_{i \in [M]})$ consists of exactly $M$ segments $S_1, S_2, \ldots, S_M$ whose domains are a partition of $[0,1]$. For each $i \in [M]$, the $i$th segment $S_i$'s end-point lies on $l_{\textup{idx}(i)}$. For $i > 1$, $S_i$'s start-point lies on line $l_{\textup{idx}(i - 1)}$, and $S_1$'s start-point is the origin, $(0,0)$. In addition, the $i$th segment must be contained in $\textup{Sector}_{\textup{sec}_i}$, where $\textup{Sector}_1$ is the subset of the positive quadrant of the $2$D plane which lies above $\yax = s_1 \xax$, $\textup{Sector}_{k + 1}$ is the subset of the positive quadrant of the $2$D plane which lies below $\yax = s_k \xax$, and for $i \in \{ 2, \ldots, k\}$, $\textup{Sector}_{i}$ is the subset of the positive quadrant of the $2$D plane which lies below $\yax = s_{i - 1} \xax$ and above $\yax = s_{i} \xax$.

For $i \in [M]$, we denote $n_i \in [0,1]$ as the length of the $i$th segment $S_i$. Note that different test-functions of the same schema may have different segment lengths $( n_i)_{i \in [M]}$, subject to some constraints described in Lemma \ref{lem:schema-constraints}. 
\end{restatable}

Figure \ref{fig:test-fn} depicts a continuous test-function, whose schema consists of four segments. Figure \ref{fig:four-segments} shows four generic types of segments.

The formal version of the aforementioned approximation property between discrete and continuous test-functions is described below. We will use this Lemma in the proof sketch.

\begin{restatable}[Discrete Approximation to Continuous Test-Function, weaker version of Lemma \ref{lem:low-prec-activ-to-string}]{lemma}{discapprox}
\label{lem:disc-approx-informal}
Suppose we are given a $(k,T)$-configuration $\{ s_i\}_{i \in [k]}$. For any schema $Y$ of $M$ segments, suppose $\caly$ is any continuous test-function of schema $Y$, with segment lengths $(\overline{n}_1(\caly), \ldots, \overline{n}_M(\caly)) \in [0,1]^M$ (where we assumed WLOG that $\sum_{j} \overline{n}_j(\caly) = 1$). Suppose every $\overline{n}_j(\caly)$ is a rational number and that the common denominator of all $( \overline{n}_j(\caly))_{j \in [M]}$ is $p$.
    
Then there exists an $n_0 \leq O(p \cdot T^k)$ so that for any positive integer multiple $n$ of $n_0$, there exists a discrete test-function $\calx$ of length $n$ so that 

\begin{align*}
    \forall i \in [k], |B_i(\caly) - B_i(\calx)| \leq \frac{T^2 + M}{n}
\end{align*}
\end{restatable}

We mentioned before that the set possible activations $(B_i(\caly))_{i \in [k]}$ induced by continuous test-functions has a nice characterization. More precisely, we are referring to the following set $\act(\{ s_i\}_{i \in [k]}) \subset [0,1]^k$.

\begin{align*}
    \act (\{ s_i\}_{i \in [k]} ) &:= \{ (B_1(\caly), \ldots, B_k(\caly) ) :\\ & \caly \text{ continuous test-function w.r.t } \{ s_i\}_{i \in [k]} \}
\end{align*}

Lemma \ref{lem:completeness-informal} gives a characterization of $\act(\{ s_i\}_{i \in [k]})$, which we will describe more in the proof sketch.

Finally, we will need the following notion of margin of a point for a linear inequality, as a notion of how far apart a point and the hyper-plane defined by the linear inequality are.

\begin{restatable}[Margin of a Point for a Linear Inequality]{definition}{margin}
\label{defn:margin}
Given a linear inequality $L$ over $M$ variables and a point $x \in \mathbb{R}^M$, define $L(x)$ as the difference between the left-hand-side and right-hand-side of the inequality when the coordinates of $x$ are plugged into $L$. Notationally, let $L(x) = 0 \iff$ $x$ satisfies the inequalities tightly (i.e. with equality). We say $L(x)$ is the margin of $x$ for $L$. 
\end{restatable}

\subsection{Proof Sketch of Lemma \ref{lem:pf-sketch-suff-cond}.}

\paragraph{Proof Plan.} We want to show that any two \crasp{} programs $f, f' \in \crasp{}^{2,K,T}$ that are not equal must be distinguished by some string $x'$ of length at most $O(T^{O(K^2)})$. For the rest of the proof, consider two \crasp{} programs $f, f' \in \crasp{}^{2,K,T}$, which on input $x \in \{ 0,1\}^n$ compute the following.

\begin{align*}
    \forall j \in [n], i \in [K], h_j^{(i)}(x) &= \ind [\ps(x)_j > \frac{b^{(i)}}{a^{(i)}} j]\\
        f(x) = \ind [\sum_{i \in [K]} &\lambda_i \ps(h^{(i)}(x))_n > z \cdot n]\\
    \forall j \in [n],  i \in [K], (h^{(i)})'_j(x) &= \ind [\ps(x)_j > \frac{(b^{(i)})'}{(a^{(i)})'} j]\\
    f'(x) = \ind [\sum_{i \in [K]}  &\lambda_i' \ps((h^{(i)})'(x))_n > z' \cdot n]\\
\end{align*}

% ( \frac{\ps(h^{(1)}(x))_n}{n}, \frac{\ps(h^{(2)}(x))_n}{n}, \ldots, \frac{\ps(h^{(K)}(x))_n}{n})\\  &\cup ( \frac{\ps((h^{(i)})'(x))_n}{n}, \frac{\ps((h^{(2)})'(x))_n}{n}, \ldots, \frac{\ps((h^{(K)})'(x))_n}{n} )

For any $n > 0$, consider the induced activations by an arbitrary string $x \in \{ 0,1\}^n$, defined as $\{ \frac{\ps(h^{(i)}(x))_n}{n}\}_{i \in [K]} \cup \{ \frac{\ps((h^{(i)})'(x))_n}{n} \}_{i \in [K]}$, per \Cref{defn:disc-test-fn}. With $k \leq 2K$ being the number of unique slopes (i.e. unique values in $\{ \frac{b^{(i)}}{a^{(i)}}\}_{i \in [K]} \cup \{ \frac{(b^{(i)})'}{(a^{(i)})'}\}_{i \in [K]}$) among the heads in the first layers of $f$ and $f'$, denote 

\begin{align*}
    (B_1(x), B_2(x), \ldots, B_k(x)) &:= \{ \frac{\ps(h^{(i)}(x))_n}{n}\}_{i \in [K]} \cup \{ \frac{\ps((h^{(i)})'(x))_n}{n} \}_{i \in [K]}
\end{align*}

To argue that the existence of distinguisher $x_0$ for $f$ and $f'$ implies existence of a short distinguisher for for $f$ and $f'$, it suffices to find an $x \in \{ 0,1\}^{n}, n \leq O((KT)^{O(K^2)})$ that induces activations $( B_i(x))_{i \in [k]}$ where:

\begin{align}
    \text{Either } \sum_{i \in [K]} \lambda_{i} B_{\textup{ord}(1,i)}(x) > z  \text{ and } \sum_{i \in [K]} \lambda_{i}' B_{\textup{ord}(2,i)}(x) \leq z'\label{eq:h1}\\
    \text{Or, }\sum_{i \in [K]} \lambda_{i} B_{\textup{ord}(1,i)}(x) \leq z  \text{ and } \sum_{i \in [K]} \lambda_{i}' B_{\textup{ord}(2,i)}(x) > z'\label{eq:h2}
\end{align}

Define the following halfspaces, $H_1^+, H_1^-, H_2^+, H_2^-$, which are strict versions of those described in Equations (\ref{eq:h1}) and (\ref{eq:h2}).

\begin{align*}
    H_1^+ &:= \{ (B_1,\ldots, B_k) : \sum_{i \in [K]} \lambda_{i} B_{\textup{ord}(1,i)} > z\}\\
    H_2^+ &:= \{ (B_1,\ldots, B_k) : \sum_{i \in [K]} \lambda_{i}' B_{\textup{ord}(2,i)} < z'\} \\
    H_1^- &:= \{ (B_1,\ldots, B_k) : \sum_{i \in [K]} \lambda_{i} B_{\textup{ord}(1,i)} < z\}\\
    H_2^- &:= \{ (B_1,\ldots, B_k) : \sum_{i \in [K]} \lambda_{i}' B_{\textup{ord}(2,i)} > z'\} 
\end{align*}

With this goal in mind, we will use the following proof plan.

\begin{enumerate}
    \item \label{pf-plan-step:1} First, we argue that if $f \neq f'$, then there exists a continuous test-function $\caly_0$ which induces activations $(B_1(\caly_0), \ldots, B_k(\caly_0))$ that is either contained in $H_1^+ \cap H_2^+$ or $H_1^- \cap H_2^-$. This is a non-trivial step which uses a technical Lemma \ref{lem:suffcond-for-asmpt}, whose proof is deferred to \Cref{appen:aux}. WLOG, suppose that $(B_1(\caly_0), \ldots, B_k(\caly_0)) \in H_1^+ \cap H_2^+$. % \footnote{If we instead use Lemma \ref{lem:stronger-suffcond-for-asmpt} in this step, a stronger version of Lemma \ref{lem:suffcond-for-asmpt}, then we can attain the stronger upper bound of $O((KT)^{O(K)})$ instead of $O((KT)^{O(K^2)})$. This is done in the main proof but not done here to keep the sketch cleaner.}
    
    \item \label{pf-plan-step:2} Next, we will show that because the set $\act (\{ s_i\}_{i \in [k]} ) \cap H_1^+ \cap H_2^+$ is non-empty, then $\act (\{ s_i\}_{i \in [k]} ) \cap H_1^+ \cap H_2^+$ must be at least a minimum ``size." We accomplish this via the following steps.
    \begin{itemize}
        \item With Lemma \ref{lem:completeness-informal}, we first characterize the set $\act (\{ s_i\}_{i \in [k]} )$ as the union of a finite number of polytopes, where each polytope is equal to the set of activations which can be induced by test-functions of a particular schema. In this way, each polytope corresponds to a unique schema, and the finite set of schemas corresponding to the finite set of polytopes serves as a ``basis" of all test-functions.

        \item We convert $\caly_0$ into a new test-function which induces the same activations as $\caly_0$, but which follows one of the aforementioned ``basis" schema, $Y$. This new test-function, which we call $\caly_1$, will be specified by a tuple of $M := M(Y) \leq k^2$ numbers $(n_1, n_2, \ldots, n_M) \in [0,1]^M$, which are the lengths of the segments of $\caly_1$ in schema $Y$.

        \item Let $A^{(M)}(Y) \subset [0,1]^M$ denote the polytope of valid settings of the lengths of segments of schema $Y$, as explained in Lemma \ref{lem:schema-constraints}. Then, there are two halfspaces $H_1^{(M)}$ and $H_2^{(M)}$ over $M$ variables such that the polytope $P := \clo(A^{(M)}(Y) \cap H_1^{(M)} \cap H_2^{(M)})$ is the set of settings of the lengths of segments of schema $Y$, which correspond to test-functions whose induced activations are contained in $\clo(H_1^+) \cap \clo(H_2^+)$. By the existence of $\caly_1$, $P \neq \emptyset$. Moreover, $P$ is a polytope whose faces are low-precision, in the sense of \Cref{defn:precision}. 

        \item Denote the set of vertices of $P$ as $V$. Let $c := \frac{1}{|V|} \sum_{v \in V} v \in P$ be the average of the vertices of $P$. We apply Lemma \ref{lem:margin-informal} to $P$  to derive a lower-bound of the margin of $c$ to the faces of $P$, of $\gamma \geq \frac{1}{|V|} \cdot \frac{1}{(\poly(K,T)\sqrt{M})^M}$. Since $c \in P \subset [0,1]^M$ is a valid setting of the lengths of the segments of schema $Y$, there is a  continuous test-function $\caly_2$ of schema $Y$ with segment lengths given by $c$. 
    \end{itemize}
    % We've demonstrated the existence of $c$ (and the corresponding test-function $\caly_2$) with large margin $\gamma$ to the faces of $P$.

    \item \label{pf-plan-step:3} It follows from steps \ref{pf-plan-step:1} and \ref{pf-plan-step:2} of the proof plan that if $f, f'$ are not equal, then there exists a point $c \in  P$ whose margin to the faces of $P$ is at least $\gamma \geq \frac{1}{3M^2} \cdot \frac{1}{(\poly(K,T)\sqrt{M})^M}$, where we used the fact that $|V| \leq 3M^2$, from Lemma \ref{lem:num-vtxs-of-P}.
    
    Using Lemma \ref{lem:l-inf-ball-precision}, we perturb the coordinates of $c \in [0,1]^M$ slightly to attain a point that still has $\frac{\gamma}{2}$ margin to the faces of $P$, but whose coordinates have precision at most $\frac{\poly(K,T)}{\gamma}$, in the sense of \Cref{defn:precision}. We will call this perturbed point $c_* := (n_1^{(c_*)}, n_2^{(c_*)}, \ldots, n_M^{(c_*)}) \in [0,1]^M$. Let $(B_1^*, \ldots, B_k^*)$ be the activations induced by a test-function of schema $Y$ with segment lengths set according to $(n_1^{(c_*)}, n_2^{(c_*)}, \ldots, n_M^{(c_*)})$.

    \item \label{pf-plan-step:4} Finally, we apply Lemma \ref{lem:disc-approx-informal} to the test-function of schema $Y$, with segment lengths set according to $(n_1^{(c_*)}, n_2^{(c_*)}, \ldots, n_M^{(c_*)})$, to attain a discrete test-function $\calx'$ of length $n$ whose induced activations are such that $||(B_1(\calx'), \ldots, B_k(\calx')) - (B_1^*, \ldots, B_k^*)||_\infty \leq O(\frac{\poly(K)\cdot \poly(T)}{n})$. Note that there is a minimal value $n_0 \leq \poly(K,T) \cdot T^{O(K)} \cdot \frac{1}{\gamma}$, which $n$ must be a multiple of, as a requirement of Lemma \ref{lem:disc-approx-informal}.

    Now, we discuss how large $n$ needs to be set so that discrete test-function $\calx'$ distinguishes $f$ and $f'$. First, $n$ must be larger than $n_0$, which was a prerequisite of using Lemma \ref{lem:disc-approx-informal}. There is a second, important requirement. WLOG, suppose that the continuous test-function $\caly_0$ outputted in step \ref{pf-plan-step:1} of the proof plan is such that:

    \begin{align*}
        (B_1(\caly_0), \ldots, B_k(\caly_0)) \in H_1^+ \cap H_2^+
    \end{align*}

    Then, our procedure in steps \ref{pf-plan-step:2} and \ref{pf-plan-step:3} are such that:

    \begin{align*}
        (B_1(\caly_1), \ldots, B_k(\caly_1)) &\in H_1^+ \cap H_2^+\\
        (B_1(\caly_2), \ldots, B_k(\caly_2)) &\in H_1^+ \cap H_2^+\\
        (B^*_1, \ldots, B^*_k) &\in H_1^+ \cap H_2^+
    \end{align*}

    Now, we would like $n$ to be large enough so that $(B_1(\calx'), \ldots, B_k(\calx'))$ is also in $H_1^+ \cap H_2^+$, so that $\calx'$ would distinguish $f$ and $f'$. To do this, we need to ensure the coordinate-wise distance between $(B_1(\calx'), \ldots, B_k(\calx'))$ and $(B^*_1, \ldots, B^*_k)$ is smaller than the margin $\gamma$ of $c_*$ on the faces of $H_1^{(M)} \cap H_2^{(M)}$ divided by the maximum $L_1$ norm of the linear inequalities defining the faces of $\act (\{ s_i\}_{i \in [k]} ) \cap H_1^+ \cap H_2^+$. Here, it is important that the margin of $c_*$ is large, as the smallest value of $n$ which will ensure that $(B_1(\calx'), \ldots, B_k(\calx')) \in H_1^+ \cap H_2^+$ is proportional to $\frac{1}{\gamma}$. In summary, we need $n \gtrapprox \Theta(\max(n_0, \frac{\poly(K,T)}{\gamma}))$, for the resulting discrete test-function $\calx'$ to distinguish $f$ and $f'$. Noting that $M \leq k^2$, $k \leq 2K$, and that $K \leq T^2$, it is sufficient to set $n = (\poly(T))^M = O(T^{O(K^2)})$. The resulting discrete test-function $\calx'$ corresponds to a string $x' \in \{ 0,1\}^{O(T^{O(K^2)})}$ which distinguishes $f$ and $f'$.

\end{enumerate}

% 3 key Lemmas which are about (\#1) The characterization of $\act (\{ s_i\}_{i \in [k]} )$ as the union of a finite number of low-precision polytopes, (\#2) Lower bound on margin of a particular point in $\act (\{ s_i\}_{i \in [k]}) \cap H_1 \cap H_2$ when it is non-empty (\#3) Approximating the activations of a continuous test-function with those of a discrete test-function. Below we describe these Lemmas in more detail and how to stitch them together to get the final result.

% We'll give a high level overview of the proof first. 

% With Lemma (\#1), we will find that $\act (\{ s_i\}_{i \in [k]} )$ is a union of a finite number of $k$-dimensional polytopes, whose faces are low-precision in the sense of \Cref{defn:precision}. Then, for any two unequal functions $f$ and $f'$, the two halfspaces defined in their second-layers (see Equations \ref{eq:h1}, \ref{eq:h2}) will have a non-empty intersection with at least one of these low-precision polytopes, denoted $A \subset [0,1]^k$. We apply Lemma (\#2) to lower bound the ``size" of the polytope given by the intersection of the two halfspaces and $A$. Finally, through Lemma (\#3) we will argue that there exists a short string $x'$ whose discrete test-function's activations are approximately equal to the activations given by the centroid of $A$ intersected with the two halfspaces. The minimum length of such an $x'$ will be inversely proportional to the size of the polytope given by the intersection of the two halfspaces and $A$.

\paragraph{Conclusions of Each Step of Proof Plan.}

Below, we write out the guarantees of each step of the Proof Plan formally. Putting these Lemmas together yields Lemma \ref{lem:pf-sketch-suff-cond}.

\begin{restatable}[Analysis of Step \ref{pf-plan-step:1}]{lemma}{pfplan1}
If $f \neq f' \in \crasp{}^{2,K,T}$, then there exists a continuous test-function $\caly_0$  such that $(B_i(\caly_0))_{i \in [k]} \in (H_1^+ \cap H_2^+) \cup (H_1^- \cap H_2^-)$. WLOG, suppose $(B_i(\caly_0))_{i \in [k]} \in H_1^+ \cap H_2^+$.
\end{restatable}

\begin{restatable}[Analysis of Step \ref{pf-plan-step:2}]{lemma}{pfplan2}
    Given continuous test-function $\caly_0$ where $(B_i(\caly_0))_{i \in [k]} \in H_1^+ \cap H_2^+$,  then there exists a continuous test function $\caly_2$  of a schema $Y$ of $M \leq k^2$ segments, where lengths of the segments of $\caly_2$ is given by $(n_i^{(c)})_{i \in [M]} \in [0,1]^M$. In addition, the $M$-dimensional point $(n_i^{(c)})_{i \in [M]}$ has margin at least $\gamma \geq \frac{1}{3M^2} \frac{1}{(\poly(K,T)\sqrt{M})^M}$ to the faces of polytope $P := \clo(A^{(M)}(Y) \cap H_1^{(M)} \cap H_2^{(M)})$, which is the set of valid settings of segment lengths of schema $Y$ that correspond to test-functions whose induced activations are contained in $H_1^+ \cap H_2^+$. 
\end{restatable} 

Together, these two Lemmas imply that if $f \neq f' \in \crasp{}^{2,K,T}$, then there exists a continuous test function $\caly_2$  of a schema $Y$ of $M \leq k^2$ segments, where lengths of $\caly_2$'s segments is given by $(n_i^{(c)})_{i \in [M]} \in [0,1]^M$ and where the margin of $(n_i^{(c)})_{i \in [M]}$ to the faces of $P$ is at least $\frac{1}{3M^2} \frac{1}{(\poly(K,T)\sqrt{M})^M}$.

\begin{restatable}[Analysis of Step \ref{pf-plan-step:3}]{lemma}{pfplan3}
    Given any point $c := (n_i^{(c)})_{i \in [M]} \in [0,1]^M$ which are segment lengths of a particular test-function of schema $Y$, suppose $c$ has margin at least $\gamma > 0$ to faces of  polytope $P = \clo(A^{(M)}(Y) \cap H_1^{(M)} \cap H_2^{(M)})$. Then, one can perturb the coordinates of  $c$ to get a point  $c_* \in [0,1]^M$ such that 
    \begin{enumerate}
        \item The margin of $c_*$ to the faces of $P$ is at least $\frac{\gamma}{2}$.
        \item The coordinates of $c_*$ have precision at most $\frac{\poly(K,T)}{\gamma}$.
    \end{enumerate}
\end{restatable}

Together, these three Lemmas imply that if $f \neq f' \in \crasp{}^{2,K,T}$, then there exists a point $(n_i^{(c_*)})_{i \in [M]} \in P$ where the margin of $(n_i^{(c_*)})_{i \in [M]}$ to the faces of $P$ is at least $\frac{1}{6M^2} \frac{1}{(\poly(K,T)\sqrt{M})^M}$ and the precision of the coordinates of $(n_i^{(c_*)})_{i \in [M]}$ is at most $O(\poly(K,T,M) \cdot (\poly(K,T)\sqrt{M})^M)$.

\begin{restatable}[Analysis of Step \ref{pf-plan-step:4}]{lemma}{pfplan4}
    If there exists $\gamma > 0, c_* \in P$ such that the margin of $c_*$ to the faces of $P$ is at least $\frac{\gamma}{2}$ and the coordinates of $c_*$ have precision at most $\frac{\poly(K,T)}{\gamma}$, then there exists discrete test-function $\calx'$ such that $(B_i(\calx'))_{i \in [k]} \in H_1^+ \cap H_2^+$ and the length of $\calx'$ is at most $O(\frac{\poly(K,T)}{\gamma}\cdot T^k)$. 
\end{restatable}

Using the facts that $M \leq k^2$, $k \leq 2K$, and $K \leq T^2$, we conclude that if $f \neq f'$, then there exists a discrete test-function $\calx'$, corresponding to string $x'$, such that $f(x') \neq f'(x')$. Moreover, the length of $x'$ and $\calx'$ is at most $O(T^{O(K^2)})$.

\paragraph{Details of Key Steps of Proof Plan.}

We now elaborate on key steps of the proof plan. We will focus on Step \ref{pf-plan-step:2}, because Step \ref{pf-plan-step:1} just applies Lemma \ref{lem:suffcond-for-asmpt}, Step \ref{pf-plan-step:3} just applies Lemma \ref{lem:l-inf-ball-precision}, and Step \ref{pf-plan-step:4} applies Lemma \ref{lem:disc-approx-informal} and analyzes how large $n$ needs to be so that the final discrete test-function distinguishes $f, f'$.

\subparagraph{Details of Step \ref{pf-plan-step:2}.} First, we characterize $\act(\{ s_i\}_{i \in [k]} )$ as the union of a finite number of polytopes.

\begin{restatable}[Characterization of $\act (\{ s_i\}_{i \in [k]})$, combination of Corollary \ref{cor:cleaner-basis-test-function} and Lemma \ref{lem:convexity-of-schema-activations}]{lemma}{pfsketchcompleteness}\label{lem:completeness-informal}
For any $(k,T)$-configuration $\{ s_i\}_{i \in [k]}$, there are a finite number of $k$-dimensional convex polytopes $\{ A_j\}_{j \in [N_k]}$, $N_k < \infty$, such that 
    \begin{align*}
    \act (\{ s_i\}_{i \in [k]}) &= \bigcup_{j \in [N_k]}  A_j
\end{align*}
% Moreover, the faces of each $A_j$ are given by $(k - 1)$-dimensional hyper-planes whose coefficients are $p_{\textup{face}}$-precision, where $p_{\textup{face}} = O( \poly(k) \cdot \poly(T))$
\end{restatable}

Figure \ref{fig:A} pictorally depicts the completeness Lemma for $k = 2$, which shows that $\act (\{ s_i\}_{i \in [2]})$ is the union of two triangles. 

The way Lemma \ref{lem:completeness-informal} is proved is by showing that for any continuous test-function $\caly$, there exists another continuous test-function of one of a finite number of possible schemas which induces the same activations $\{ B_i(\caly)\}_{i \in [k]}$.  We call these $N_k$ schemas ``basis" schema; they are a ``basis" in the sense that any continuous test-function is equivalent to a continuous test-function which follows some basis schema. Figure \ref{fig:basis-schema} depicts a basis schema. We defer the details to \Cref{appen:completeness}. The main take-away is that each polytope $A_j \subset [0,1]^k$ corresponds to a schema $Y_j$, and $A_j$ is the set of activations which can be induced by a test-function of schema $Y_j$.
\begin{align*}
    \forall j \in [N_k], A_j &:= \{ (B_1(\caly),\ldots ,B_k(\caly)) : \caly \text{ valid test-function of schema $Y_j$ }\} \subset [0,1]^k
\end{align*}
\\
\\
Thus, given $\caly_0$ from Step \ref{pf-plan-step:1} such that, WLOG, $(B_i(\caly_0))_{i \in [k]} \in H_1^+ \cap H_2^+$, then there is an equivalent test-function $\caly_1$ of one of the basis schema, $Y \in \{ Y_j\}_{j \in [N_k]}$, such that $(B_i(\caly_1))_{i \in [k]} = (B_i(\caly_0))_{i \in [k]} \in H_1^+ \cap H_2^+$. Denote the set $A(Y)$ as the particular $A_j$ which corresponds to schema $Y$. 

\begin{align*}
    A(Y) &:= \{ (B_1(\caly),\ldots ,B_k(\caly)) : \caly \text{ valid test-function of schema $Y$ }\} \subset [0,1]^k
\end{align*}

Now, suppose that schema $Y$ consists of $M$ segments, whose lengths we denote $(n_1, n_2, \ldots, n_M) \in [0,1]^M$ with $\sum_{i \in [M]}n_i = 1$. Note that not all settings of $(n_1, n_2, \ldots, n_M)$ are valid, because there are linear constraints which must be met by $(n_1, n_2, \ldots, n_M)$, which are described in Lemma \ref{lem:schema-constraints}. Define $A^{(M)}(Y) \subset [0,1]^M$ as the set of all valid settings of $(n_1, n_2, \ldots, n_M)$, per Lemma \ref{lem:schema-constraints}. $A^{(M)}(Y)$ is an $(M - 1)$-dimensional polytope, whose faces are parameterized by linear inequalities whose coefficients are at most $\poly(K,T)$ in magnitude (i.e. $\poly(K,T)$-precision).

\begin{align*}
    A^{(M)}(Y) &:= \{ (n_1,\ldots ,n_M) : \text{ valid segment lengths of schema $Y$ and } \sum_{i \in [M]} n_i = 1\} \subset [0,1]^M
\end{align*}

Further, define $H_1^{(M)}$ and $H_2^{(M)}$ as the analogous halfspaces to $H_1^+$ and $H_2^+$ but in the space of segment lengths as follows. There exists a linear map $L : \mathbb{R}^M \to \mathbb{R}^k$ which maps points in $A^{(M)}(Y)$ to points in $A(Y)$. $L \in \{ 0,1\}^{k \times M}$ is such that $L_{ij} = 1 \iff $ segment $j$ in schema $Y$ lies above line $i$ (that is, for every $\xax $ in the domain of segment $j$, the $\yax$-value of the segment at $\xax$ is at least $s_i \cdot \xax$) and hence contributes to the $i$th activation $B_i(\caly)$ of any test-function $\caly$ of schema $Y$. Using $L$,  we can rewrite the inequalities which characterize $H_1, H_2$ in terms of $(n_1,\ldots, n_M)$.

\begin{align*}
 H_1^{(M)} := \{(n_1,\ldots,n_M) :  \sum_{i = 1}^{K} \lambda_i B_{\textup{ord}(1, i)}  > z\} &= \{(n_1,\ldots,n_M) :  \sum_{i = 1}^{K} \lambda_i \sum_{j \text{ s.t. } L_{\textup{ord}(1, i), j} = 1} n_j  > z\}\\
   H_2^{(M)} := \{(n_1,\ldots,n_M) :  \sum_{i = 1}^{K} \lambda_i' B_{\textup{ord}(2, i)}  < z'\} &= \{(n_1,\ldots,n_M) :  \sum_{i = 1}^{K} \lambda_i' \sum_{j \text{ s.t. }  L_{\textup{ord}(2, i), j} = 1} n_j  < z'\}
\end{align*}

With $P := \clo(A^{(M)}(Y) \cap H_1^{(M)} \cap H_2^{(M)})$, we use the fact that $(B_i(\caly_1)_{i \in [k]} \in H_1^+ \cap H_2^+$ to deduce that $A^{(M)}(Y) \cap H_1^{(M)} \cap H_2^{(M)} \neq \emptyset$. Let $V$ be the set of vertices of $P$. This next Lemma is for lower-bounding the margin of point $c = \frac{1}{|V|}\sum_{v \in V} v \in [0,1]^M$ on the faces of $P$, when the latter is non-empty.

\begin{restatable}[Margin lower bound, slightly informal version of Lemma \ref{lem:margin}]{lemma}{pfsketchmargin}\label{lem:margin-informal}
Consider a nonempty $d$-dimensional polytope $P \subset \mathbb{R}^{d}$ with vertices $V$ and $N$ faces. Suppose the faces of $P$ are each defined by a linear inequality over variables $\{ x_i\}_{i \in [d]}$, with integer coefficients of magnitude at most $p_{\textup{face}}$, where points on the face satisfy the linear inequality with equality. For $j \in [N]$, define $L_j$ as the linear inequality for the $j$th face of $P$. Then, for any $j \in [N]$, for any vertex $x \in V$ which does not lie on the $j$th face of $P$, we have the following lower bound on the margin of $x$ on the $j$th face of $P$.
    \begin{align*}
        L_j(x) \gtrapprox \frac{1}{(p_{\textup{face}}\sqrt{d})^d}
    \end{align*}
\end{restatable}

With $d = M$ and $p_{\textup{face}} = \poly(K,T)$ and using the linearity of the margin, it follows that the margin of $c= \frac{1}{|V|}\sum_{v \in V}v$ on any face of $P$ is at least $\frac{1}{|V|}\frac{1}{(\poly(K,T)\sqrt{M})^M}$. $c$ is a point in $[0,1]^M$, and it is a valid setting of segment lengths for a test-function of schema $Y$, since $c$ has positive margin to the faces of $P$, so it is contained in $P \subset A^{(M)}(Y)$. The activations of the test-function with segment lengths given by $c$ will be contained in $H_1^+ \cap H_2^+$ since $c \in P \subset H_1^{(M)} \cap H_2^{(M)}$. This concludes Step \ref{pf-plan-step:2}.

%% file: sections/conclusion.tex
\section{Conclusion}

We prove guarantees of length generalization for various function classes in an idealized setting. We formalize the framework of non-asymptotic length generalization, which requires a computable upper bound for length complexity. We show the Minimum-Complexity Interpolator learning algorithm achieves optimal length complexity. We show that whether a function class admits non-asymptotic length generalization is equivalent to the decidability of its language equivalence problem, which implies that there is no computable upper bound for the length complexity of CFGs. On the positive side, we show that the length complexity of DFAs is $2n - 2$ where $n$ is the number of states of the ground-truth automaton. We show that the length complexity of 1-layer \crasp{} functions is  $O(T^2)$ when the ground-truth function has precision $T$, and that the length complexity of 2-layer \crasp{} functions is $O(T^{O(K)})$ when the ground-truth function has precision $T$ and $K$ heads.

It is open whether the proof techniques can be extended to 3-Layer \crasp{} programs, or to \crasp{} programs whose layers contain bias terms. It is open how to formalize a weaker notion of partial length generalization which does not entail length generalization to arbitrary length inputs.
%  The latter is interesting as it may allow one to prove guarantees for languages like Dyck-1.

% Third, since our proof technique currently studies properties of continuous test-functions as a proxy for discrete test-functions, one weakness is that it requires the \crasp{} programs to be homogeneous, in the sense that there are no bias (constant) terms in the definition of $\crasp{}^{2,K,T}$. Thus, there is a question of whether one can study discrete test-functions directly and extend these results to a broader class of non-homogeneous \crasp{} functions, which may include interesting functions like Dyck-1. Fourth, there is the higher-level question of whether we can formalize a weaker notion of partial length generalization which does not entail length generalization to arbitrary length inputs.

 % This problem is of importance since it could predict quantitatively how the number of layers, heads, and precision affect how difficult length generalization is.

%% file: sections/acknowledgements.tex
\section*{Acknowledgements}

This work was supported by NSF IIS 2211780 and the Stanford HAI–Google Cloud Credits Program. TC acknowledges funding from an NSF Graduate Research Fellowship. ZL acknowledges funding from an OpenAI Superalignment Grant.

%% file: sections/appendix.tex
\tableofcontents
\section*{Organization of Appendices.}

Section \ref{appen:length-gen-in-the-limit} contains the proof of \Cref{prop:asympt-id}. Section \ref{appen:foundations} contains the proofs of \Cref{lem:opt-learner}, Lemma \ref{lem:equiv-decidability}, and Lemma \ref{lem:equiv-finite-identification}. Section \ref{appen:chomsky} contains the proof of Propositions \ref{prop:dfa-len-gen} and \ref{prop:cfg-len-gen}. Section \ref{appen:crasp1} contains the proof of Theorem \ref{thm:crasp1}. Section \ref{appen:full-pf-thm} contains the full proof of Theorem \ref{thm:main}. It will use Definitions and Lemmas defined in the later sections \ref{appen:completeness} to \ref{appen:aux}. Section \ref{appen:completeness} contains proofs of Lemmas relating to completeness of basis schema. Section \ref{appen:margin} contains proofs of Lemmas relating to lower bounding the margin of the average-of-vertices of a polytope with low-precision faces. Section \ref{appen:disc-approx} contains proofs of Lemmas for approximating a continuous test-function with a discrete test-function. Section \ref{appen:aux} contains auxiliary Lemmas.

\input{sections/appendices/length_gen_in_the_limit}

\input{sections/appendices/foundations}

\input{sections/appendices/chomsky}

\section{Proofs for Length Generalization of \crasp}

\input{sections/appendices/crasp1}

\input{sections/appendices/main-proof}

\input{sections/appendices/completeness}

\input{sections/appendices/margin}

\input{sections/appendices/disc-approx}

\input{sections/appendices/aux}

%% file: sections/appendices/length_gen_in_the_limit.tex
\section{Proof of \Cref{prop:asympt-id}}\label{appen:length-gen-in-the-limit}

Here we give a proof of Proposition \ref{prop:asympt-id}, which is similar to the proof of Theorem I.4 of \cite{gold}.

\lengeninlimit*

\textit{Proof.} Consider a learning algorithm which enumerates functions in $\calf^\calr$. It maintains a current hypothesis, which is initialized to the first element of the enumeration. Given an input training set, the learning algorithm checks whether the current hypothesis interpolates the training data. If it does not, then the current hypothesis is updated to the next function in the enumeration until the current hypothesis interpolates the training data, at which point the current hypothesis is outputted.

    First, because $\calr$ maps $\{ 0,1\}^*$ to $\calf^\calr$, then we can enumerate $\calf^\calr$ simply by enumerating all strings $p \in \{ 0,1\}^*$, say in the order of increasing binary-representations of natural numbers, and then computing $\calr(p)$ for each string $p$. The computation of $\calr(p)$ will always halt by \Cref{defn:encoding-system}. Every function in $\calf^\calr$ appears at some point in this enumeration by the definition of $\calf^\calr$. Suppose that we denote this enumeration by $\mathcal{E} = \{ f_0, f_1, f_2, \ldots\}$, with $\mathcal{E}_c := \{ f_0, f_1, f_2, \ldots, f_c\}$ being the first $c + 1$ elements of the enumeration, for any $c \in \nat$.
    
    Second, because the output of $\calr$ is always a computable function which halts on every input, the learning algorithm described in the first paragraph is computable.
    
    By \Cref{lem:opt-learner} with $\calc(\cdot)$ being the mapping from a binary string to the integer it represents in binary (so that each finite representation is mapped to its order in the aforementioned enumeration), it suffices to show that for any $c \in \nat$, $N(\mathcal{E}_c) := \min\{n : \forall f \neq f' \in \mathcal{E}_c, \exists x \in \{ 0,1\}^{\leq n} \text{ s.t. } f(x) \neq f'(x) \} < \infty$. This follows directly from the fact that $|\mathcal{E}_c| = c < \infty$ and that for any $f \neq f' \in \mathcal{E}_c$, there exists some string $x, |x| < \infty$ where $f(x) \neq f'(x)$. For each pair of unequal functions in $\mathcal{E}_c$, pick one such string $x$ which distinguishes them, and take $N(\mathcal{E}_c)$ to be the maximum of the lengths of these distinguishers. This must be finite. $\blacksquare$

%% file: sections/appendices/foundations.tex
\section{Proofs on Equivalence of Different Non-Asymptotic Length Generalization Definitions}\label{appen:foundations}

We will prove that $\mci$ is optimal with respect to the quantity $N_{\mci}(c)$. Note we will drop the superscripts for $\calf, \calc, \calr$ from the notation for convenience.

\optmci*

\textit{Proof of \Cref{lem:opt-learner}}. For any $c \in \nat$, we have  

\begin{align*}
    N(\calf^\calr_c) := \min \{ n \in \nat : \forall f\neq f' \in \calf \textrm{ with } \calc^\calr(f'), \calc^\calr(f) \leq c,\\
    \textrm{ there exists } x \in \{ 0,1\}^{\leq n} \textrm{ s.t. } f(x) \neq f'(x)\}
\end{align*} 

% Since $|\calf_c| < \infty$, then $n_c < \infty$. By \Cref{asmpt:E-exists}, $\calc$ is such that $\calf_c := \{ f \in \calf : \calc^\calr(f) \leq c\}$ is a finite set. 

We will show separately that $N_{\mci}(c) \geq \min_\cala N_\cala(c) \geq N(\calf^\calr_c)$ and $N_{\mci}(c) \leq N(\calf^\calr_c)$.

\quad (1). $\min_\cala N_\cala(c) \geq N(\calf^\calr_c)$ follows from the fact that for any $N < N(\calf^\calr_c)$, there exists $f\neq f'$ with $\calc^\calr(f'), \calc^\calr(f) \leq c$ that agree on all inputs $x \in \{ 0,1\}^{\leq N}$. Given training set $\{ (x, f(x)) : \forall x \in \{ 0,1\}^{\leq N}\} = \{ (x, f'(x)) : \forall x \in \{ 0,1\}^{\leq N}\}$, any algorithm $\cala$ which outputs an encoding of a function which is not equal to $f$ will be wrong in the case the ground truth is $f$. Meanwhile, any algorithm $\cala$ which outputs an encoding of a function which is not equal to $f'$ will be wrong in the case the ground truth is $f'$. Thus, $N_\cala(f) > N$ or $N_\cala(f') > N$, so $N_\cala(c) > N$.

\quad (2). $N_{\mci}(c) \leq N(\calf^\calr_c)$. For $f$ with $\calc^\calr(f) \leq c$, suppose that on input of $\{ (x, f(x)) : \forall x \in \{ 0,1\}^{\leq N(\calf^\calr_c)}\}$, $\mci$ outputs $p'$, where $\calr(p') \neq f$. This implies that $\calc(\calr(p')) \leq \calc(p') \leq \calc^\calr(f) \leq c$ and that $\calr(p')$ is consistent with $\{ (x, f(x)) : \forall x \in \{ 0,1\}^{\leq N(\calf^\calr_c)}\}$. This is a contradiction of the definition of $N(\calf^\calr_c)$, which says that all pairs of unequal functions $f, f'$ with $\calc^\calr(f), \calc^\calr(f') \leq c$ can be distinguished from each other by some input of length at most $N(\calf^\calr_c)$.$\blacksquare$

Denote $p \equiv p'$ if $\calr(p) = \calr(p')$. The following Lemma is also useful for showing that the same bounds on the length complexity hold whether we think of $\mci$ as operating over finite descriptions $p \in \{ 0,1\}^*$ (which is the actual implementation) or as operating over functions $f \in \calf$ (which is a more abstract way to think of $\mci$).

\begin{lemma}\label{lem:equiv-of-p-and-f}
    Given $(\calf,  \calr, \calc)$, with $\calc^\calr(f) := \min_{p \in \{ 0,1\}^*, \calr(p) = f} \calc(p)$, then for all $c \in \nat$:
    \begin{align*}
        &\min \{ n : \forall f \neq f' \in \calf, \calc^\calr(f), \calc^\calr(f') \leq c, \exists x, |x| \leq n, f(x) \neq f'(x)\}\\
        &= \min \{ n : \forall p \not\equiv p' \in \{ 0,1\}^*, \calc(p), \calc(p') \leq c, \exists x, |x| \leq n, \calr(p)(x) \neq \calr(p')(x)\}
    \end{align*}
\end{lemma}

\textit{Proof.} Every function $f \in \calf$ with $\calc^\calr(f) \leq c$ has some finite description $p$ with $\calr(p) = f$ and $\calc(p) \leq c$. No function $f \in \calf$ with $\calc^\calr(f) > c$ has any finite description $p$ with $\calr(p) = f$ and $\calc(p) \leq c$. $\blacksquare$

Now, we will prove that \Cref{defn:nonasympt-lgen} is equivalent to a few other definitions. First, recall the definition of non-asymptotic length generalization and its equivalent definitions.

\nonasymptoticlgen*
\langequivprob* 
\finiteidentprop*

We now prove the following equivalences.
% \optmci*

\equivweak*

\textit{Proof.} Suppose TM $E$ enumerates elements in $\{ 0,1\}^*$ in non-decreasing order of their complexity according to $\calc$.

\textbf{$N(\calf_c^\calr)$ computably bounded in $c$ $\implies$ Language equivalence problem for $\calr$ decidable.}

Suppose there is a computable procedure $F$ that, for any $c$, receives as input $c$ and outputs an upper bound on $N(\calf_c^\calr)$. We will describe an algorithm that is given any two finite descriptions, $p, q \in \{ 0,1\}^*$, and uses $F$ to determine if $\calr(p) = \calr(q)$. 

Given $p, q \in \{ 0,1\}^*$,  compute $c = \max(\calc(p), \calc(q))$ and generate the training dataset $D_{F(c)}(\calr(q))$. Now, check whether $\forall (x,\calr(q)(x)) \in D_{F(c)}(\calr(q))$, that  $\calr(p)(x) = \calr(q)(x)$. If this is the case, return ``equivalent." Otherwise, return ``non-equivalent".

To argue correctness, if there is some $(x,\calr(q)(x)) \in D_{F(c)}(\calr(q))$ where $\calr(p)(x) \neq \calr(q)(x)$, then clearly these two functions are not equal. If $\forall (x,\calr(q)(x)) \in D_{F(c)}(\calr(q)), \calr(p)(x) = \calr(q)(x)$, then since $\calc(\calr(p)) \leq \calc(p) \leq c$ and $\calc(\calr(q)) \leq \calc(q) \leq c$, then it must be that $\calr(p) = \calr(q)$, or else $F(c)$ is not an upper bound of $N(\calf_c^\calr)$.

\textbf{Language equivalence problem for $\calr$ decidable $\implies$ $N(\calf_c^\calr)$ computably bounded in $c$.} Suppose TM $M$ solves the Language Equivalence Problem for $\calr$. By  \Cref{lem:equiv-of-p-and-f}, it suffices to show that the following quantity is computably bounded in $c \in \nat$.

\begin{align*}
    \min \{ n : \forall p \not\equiv p' \in \{ 0,1\}^*, \calc(p), \calc(p') \leq c, \exists x, |x| \leq n, \calr(p)(x) \neq \calr(p')(x)\}
\end{align*}

\Cref{alg:Nc} computes an upper bound of this quantity.

\begin{algorithm}
  \caption{Computation of $N(\calf_c^\calr)$}\label{alg:Nc}
  \begin{algorithmic}[1]
    \REQUIRE Integer $c\in\mathbb{N}$
    \ENSURE $N(\calf_c^\calr)$

    % ---------- Stage 1: enumerate all pairs ----------
    \STATE $N \leftarrow 0$
    \FORALL{pairs of programs $(p,q)$ (enumerated by $E$) with $\mathcal{C}(p),\mathcal{C}(q)\le  c$}
            \IF{$M(p,q)=\text{``non-equivalent''}$}                     
                    \FORALL{strings $x \in \{0,1\}^{*}$ with non-decreasing length}
                        \IF{ $\calr(p)(x) \neq \calr(q)(x)$}
                            \STATE $N \leftarrow \max(N, |x|)$
                            \STATE \textbf{break}
                        \ENDIF
                    \ENDFOR
            \ENDIF 
    \ENDFOR
    \STATE \textbf{return} $N$

  \end{algorithmic}
\end{algorithm}

\Cref{alg:Nc} always terminates since the for-loop in line 4 only is performed on $p,q \in \{ 0,1\}^*$ which are not equivalent and hence are distinguished by a finite string. Thus, \Cref{alg:Nc} returns the smallest length required to distinguish every two non-equivalent $p, q \in \{ 0,1\}^*$.

It follows from \Cref{lem:opt-learner} that the language equivalence problem for $\calr$ being decidable is equivalent to non-asymptotic length generalization for $(\calf^\calr, \calr, \calc)$ when $\calc$ satisfies \Cref{asmpt:E-exists}. $\blacksquare$

Now we prove the following equivalence.

\equivfinitelgen*

\textit{Proof.} Suppose TM $E$ enumerates elements in $\{ 0,1\}^*$ in non-decreasing order of their complexity according to $\calc$.

\paragraph{$N(\calf_c^\calr)$ computably bounded in $c$ $\implies$ $\calf^\calr$ admits Finite Length Generalization with Complexity Information w.r.t. $\calr$ and $\calc$.}

Let $ F: \nat \to \nat$ be a computable upper bound on $N(\calf_c^\calr)$. Let $\cala$ be given by \Cref{alg:A}, on input $D_N(f_*) $ and $ c \geq \calc^\calr(f_*) \in \nat$.

\begin{algorithm}
  \caption{Learning Algorithm $\mathcal{A}$}\label{alg:A}
  \begin{algorithmic}[1]     % the [1] prints line numbers
    \REQUIRE Dataset $D_N(f_\ast)$; $c\in\mathbb{N}$ with $c \ge \calc^\calr(f_\ast)$
    \ENSURE Either ``\textbf{pass}'' or an program $\hat{p}$

    \IF{$N < F(c)$}
        \STATE \textbf{return} ``\textbf{pass}''
    \ELSE
        \STATE $\hat{p} \leftarrow \mathcal{A}_{\mathrm{mci}}\!\bigl(D_N(f_\ast)\bigr)$
        \STATE \textbf{return} $\hat{p}$
    \ENDIF
  \end{algorithmic}
\end{algorithm}

% We claim that $\cala$ satisfies \Cref{defn:prop1} with $M_\cala(\cdot) = F(\cdot)$. 

First, $\cala$ only ever returns ``pass" or some $\hat{p} \in \{ 0,1\}^*$ since $\mci$ only ever returns elements of $\{ 0,1\}^*$. Second, by \Cref{lem:opt-learner}, for $c \geq \calc^\calr(f_*)$ and $\forall N \geq F(c)$, then $\calr(\mci(D_N(f_*))) = f_*$. Thus, whenever $\cala$ returns an element $\hat{p} \in \{ 0,1\}^*$, it is such that $\calr(\hat{p}) = f_*$, and there is at least one $N$ where $\calr(\mci(D_N(f_*))) = f_*$. % This also gives the third requirement of \Cref{defn:prop1}.

\paragraph{$\calf^\calr$ admits Finite Length Generalization with Complexity Information w.r.t. $\calr$ and $\calc$  $\implies$ $N(\calf_c^\calr)$ computably bounded in $c$.} We need to prove that there exists some computable $F$ which upper bounds $N(\calf_c^\calr)$. Suppose $\cala$ satisfies \Cref{defn:prop1}. Let $F$ be given by \Cref{alg:F-old}.

\begin{algorithm}
  \caption{Algorithm for $F$}\label{alg:F-old}
  \begin{algorithmic}[1]
    \REQUIRE Integer $c\in\mathbb{N}$
    \ENSURE $N_{\max}$

    \STATE $N_{\max} \leftarrow 0$          
    \FORALL{$p \in \{ 0,1\}^*$ with $\mathcal{C}(p) \le c$ (Enumerate with $E$)} 
        \STATE $N \leftarrow 0$
        \WHILE{$\mathcal{A}(D_{N}(\calr(p)),\mathcal{C}(p)) =$ ``pass''}
            \STATE $N \leftarrow N + 1$
        \ENDWHILE
        \STATE $N_{\max} \leftarrow \max(N_{\max}, N)$
    \ENDFOR
    \STATE \textbf{return} $N_{\max}$
  \end{algorithmic}
\end{algorithm}

\Cref{alg:F-old} is computable, due to the guarantee of $\cala$ that for any $f_*$, given $c \geq \calc^\calr(f_*)$, there exists some $N$ where $\calr(\cala(D_N(f_*))) = f_*$, which ensures termination of \Cref{alg:F-old}. 

Regarding correctness, if there exists some $c_* \in \nat$ where $F(c_*) < \min \{ n : \forall p \not\equiv p' \in \{ 0,1\}^*, \calc(p), \calc(p') \leq c_*, \exists x, |x| \leq n, \calr(p)(x) \neq \calr(p')(x)\}$, then there must exist two $p, q \in \{ 0,1\}^*$ with $\calc(p), \calc(q) \leq c_*$, which are not equivalent, but which agree on all inputs of length at most $F(c_*)$. WLOG, suppose that $\calc(p) \leq \calc(q)$. 

There exists $N_q \leq F(c_*)$ where $\cala(D_{N_q}(\calr(q)), \calc(q)) \neq$ ``pass". In particular, $\cala(D_{N_q}(\calr(q)), \calc(q))$ must return a $\hat{p}$ which is equivalent to $q$, where  $q\not\equiv p$. On the other hand, since $N_q \leq F(c_*)$, we have $D_{N_q}(\calr(q)) = D_{N_q}(\calr(p))$. Thus, we have shown that $\cala(D_{N_q}(\calr(p)), \calc(q)) \neq$ ``pass" and is a finite representation which is not equivalent to the ground-truth $p$. Since $\calc(p) \leq \calc(q)$, this yields a contradiction with the fact that whenever $\cala$ is given an upper bound on the ground-truth complexity and $\cala$ does not return ``pass", it must return a finite description $\hat{p}$ which is equivalent to the ground-truth. $\blacksquare$

%% file: sections/appendices/chomsky.tex
\section{Proofs on Length Generalization for DFAs and CFGs.}\label{appen:chomsky}

\dfalgen*

In the following proof, we essentially describe and analyze the State Minimization Algorithm \cite{hopcroft-ullman}. Before proving the proposition, we will need a few concepts and Lemma \ref{lem:suff-cond-for-dfa-learning}.

Suppose there is some ground-truth minimal DFA, $D$, of $n$ states, with language $L := L(D)$. Define $\eps$ as the empty string, of length 0. For two strings $u,v$, denote $uv$ as their concatenation. For any $x \in \{ 0,1\}^*$, denote $f_*(x) = 1 \iff x \in L$. Denote $Q$ as the set of $n$ states for $D$ and $\delta(\cdot, \cdot) : Q \times \{ 0,1\} \to Q$ as the state transition function for $D$, with start state $q_0$. Let $F \subset Q$ be the set of accepting states. We say that states $q, q' \in Q$ are distinguished by string $v \in \{ 0,1\}^*$ if $\delta(q, v) \in F$ and $\delta(q', v) \notin F$, or if $\delta(q, v) \notin F$ and $\delta(q', v) \in F$. 

In general, applying a string from the start state $q_0$ in the DFA will cause the DFA to end up in some state $q \in Q$, and an identification can be made between the  input string and the state it causes the DFA to end up in. Thus, to learn the DFA, we would like to find a mapping from $\{ 0,1\}^* $ to $[n]$, which tells us which strings correspond to which states in the ground-truth DFA. To do this, we claim it is sufficient to consider the equivalence class given by the sets $\{ E(u)\}_{u \in \{ 0,1\}^{\leq n}}$, where:

\begin{align*}
    \forall u \in \{ 0,1\}^{\leq n}, E(u) := \{ v \in \{ 0,1\}^{ \leq n - 2} : uv \in L\}
\end{align*}

If for $u, u' \in \{ 0,1\}^{\leq n}, E(u) = E(u')$, then we will claim that these two strings correspond to the same state. Otherwise, $u, u'$ correspond to different states. This is what is meant by an equivalence class over $\{ E(u)\}_{u \in \{ 0,1\}^{\leq n}}$. 

Once a learning algorithm can make this identification between strings and the states they correspond to, then the learning algorithm can infer the transition function of the DFA by considering, for each $u \in \{ 0,1\}^{\leq n - 1}$, the states $E(u)$, $E(u0)$, and $E(u1)$, via the sets $\{ E(u)\}_{u \in \{ 0,1\}^{\leq n - 1}}$  and $\{ E(u)\}_{u \in \{ 0,1\}^{\leq n}}$. This gives the transition function. $E(\eps)$ corresponds to the start state. Strings $u$ corresponding to accept states will be such that $\eps \in E(u)$. Note that we just need to consider $\{ E(u)\}_{u \in \{ 0,1\}^{\leq n}}$ instead of $\{ E(u)\}_{u \in \{ \eps\} \cup \{ 0,1\}^{*}}$ to characterize the states of $D$, since any $u$ of length larger than $n$ will cause the DFA to reach some state that is also reached by a string of length at most $n$.  

We claim that proving the following property about our sets $E(u)$ suffices to prove that there is a learning algorithm which identifies the ground-truth DFA.

\begin{lemma}\label{lem:suff-cond-for-dfa-learning} 
For any minimal DFA $D$ with $n \geq 2$ states and transition function $\delta$, then with $E(u)$ defined as above for each $u \in \{ 0,1\}^{\leq n}$, we have  $\forall u,u' \in \{ 0,1\}^{\leq n}, \delta(q_0, u) = \delta(q_0, u') \iff E(u) = E(u')$.
\end{lemma}

First, we will show how to prove Proposition \ref{prop:dfa-len-gen} with this Lemma.

\textit{Proof of Proposition \ref{prop:dfa-len-gen}.} Suppose a DFA learning algorithm is given inputs of length $2n - 2$. For $n \geq 2$, Lemma \ref{lem:suff-cond-for-dfa-learning} implies that the learning algorithm can identify the states of $D$ by constructing the sets $\{ E(u)\}_{u \in \{ 0,1\}^{\leq n}}$, using its training data $\{ (x, f_*(x)) : |x| \leq 2n - 2\}$. Each state is identified with an equivalence class of $\{ E(u)\}_{u \in \{ 0,1\}^{\leq n - 1}}$, since each state in an $n$ state DFA can be reached by a string of length at most $n - 1$. The state transitions can be identified from $\{ E(u)\}_{u \in \{ 0,1\}^{\leq n}}$ by considering, for each $u \in \{ 0,1\}^{\leq n - 1}$, the states $E(u)$, $E(u0)$, and $E(u1)$. $E(\eps)$ corresponds to the start state. Strings $u$ corresponding to accept states will be exactly those such that $\eps \in E(u)$. In the case that the learning algorithm only receives inputs of length at most $0$ (i.e. it only receives $(\eps, f_*(\eps))$) or $1$, it will simply output the DFA with language $\emptyset$ if the labels of all inputs received are $0$ and it will output the DFA with language $\{ 0,1\}^*$ if the labels of all inputs received are $1$. This handles the case where $n = 1$.

The learning algorithm above identifies the ground-truth DFA, requiring inputs of length at most $2n - 2$ when the ground-truth DFA has $n$ states. By \Cref{lem:opt-learner}, the Minimum Complexity Interpolator $\mci$ will also require inputs of length at most $2n - 2$ to identify the ground-truth DFA. This proves the Proposition. $\blacksquare$

Now we prove Lemma \ref{lem:suff-cond-for-dfa-learning}.

\textit{Proof of Lemma \ref{lem:suff-cond-for-dfa-learning}.} Showing $\delta(q_0, u) = \delta(q_0, u') \implies E(u) = E(u')$ is easy, since if $\delta(q_0, u) = \delta(q_0, u')$, then the states reached by $u, u'$ from $q_0$ are the same, so no string of any length can distinguish them $\delta(q_0, u), \delta(q_0, u')$, so $E(u) = E(u')$. 

We now show $\delta(q_0, u) \neq \delta(q_0, u') \implies E(u) \neq E(u')$. We are given that the ground truth DFA has $n$ states and is the minimal DFA for its language, so that there are no DFAs of fewer states with the same language. Given this, we claim that there are exactly $n$ equivalence classes among $\{ E(u)\}_{u \in \{ 0,1\}^{\leq n}}$. Each equivalence class must correspond to a unique state, so each of $n$ unique states can be identified with a unique $E(u)$, finishing the proof of Lemma \ref{lem:suff-cond-for-dfa-learning}.

We will now show that there are exactly $n$ equivalence classes among $\{ E(u)\}_{u \in \{ 0,1\}^{\leq n}}$. For $0 \leq i \leq n - 2, u \in \{ 0,1\}^{\leq n}$, define $E_i(u) := \{ v \in \{ 0,1\}^{\leq i} : uv \in L\}$. To do this, we claim that for any $i < n - 2$, if the number of equivalence classes in $\{ E_i(u)\}_{u \in \{ 0,1\}^{\leq n}}$ is less than $n$, then there must be some $u, u' \in \{ 0,1\}^{\leq n}$ and $v_{i + 1} \in \{ 0,1\}^{i + 1}$  such that

\begin{align*}
    E_i(u) &= E_i(u')\\
    \text{ and } f_*(uv_{i + 1}) &\neq f_*(u'v_{i + 1})
\end{align*}

Suppose this were not the case. That is, suppose that for some $i < n - 2$, (1) the number of equivalence classes in $\{ E_i(u)\}_{u \in \{ 0,1\}^{\leq n}}$ is less than $n$. (2) Yet, $\forall u, u' \in \{ 0,1\}^{\leq n}, \forall v_{i + 1} \in \{ 0,1\}^{i + 1},$ $E_i(u) = E_i(u') \implies  f_*(uv_{i + 1}) = f_*(u'v_{i + 1})$. We want to show that together, (1) and (2) contradict the minimality of the ground-truth DFA.

We claim that, by induction, (2) implies that for every $k \geq i$, for every $u, u' \in \{ 0,1\}^{\leq n}$ and $v_{k + 1} \in \{ 0,1\}^{k + 1}$, $E_k(u) = E_k(u') \implies f_*(uv_{k + 1}) = f_*(u'v_{k + 1})$. This is true for base case $k = i$ by (2). 

For the inductive step, suppose the claim is true for $k > i$. Suppose $u, u' \in \{ 0,1\}^{\leq n}$ are such that $E_k(u) = E_k(u')$. Then $\forall v_k \in \{ 0,1\}^{\leq k}$, $f_*(uv_k) = f_*(u'v_k)$. For any such string $v_k$ with length $k$, we can break $v_k$ into a prefix of $k - i$ bits and a suffix of $i$ bits. Thus, 

\begin{align*}
    \forall v_{k - i} \in \{ 0,1\}^{k - i}, \forall v_i \in \{ 0,1\}^i, f_*(uv_{k - i} v_i) = f_*(u'v_{k - i} v_i)
\end{align*}

Each string $uv_{k - i}$ and $u'v_{k - i}$ will correspond to some state reachable by some strings $u'', u''' \in \{ 0,1\}^{\leq n}$, in the sense that $\delta(q_0, uv_{k - i}) = \delta(q_0, u'')$ and $\delta(q_0, u'v_{k - i}) = \delta(q_0, u''')$, as each state in the DFA can be reached by a string of length at most $n - 1$. Thus, $E_i(u'') = E_i(u''')$, so that by (2), $\forall v_{i + 1} \in \{ 0,1\}^{i + 1}, f_*(u'' v_{i + 1}) = f_*(u''' v_{i + 1})$. Since $\delta(q_0, uv_{k - i}) = \delta(q_0, u'')$ and $\delta(q_0, u'v_{k - i}) = \delta(q_0, u''')$ imply that for any such $v_{i + 1}$, $f_*(uv_{k - i} v_{i + 1}) =  f_*(u'' v_{i + 1}) $ and $f_*(u'v_{k - i} v_{i + 1}) =  f_*(u''' v_{i + 1})$, then $\forall v_{i + 1} \in \{ 0,1\}^{i + 1}, f_*(uv_{k - i} v_{i + 1}) = f_*(u'v_{k - i} v_{i + 1})$. In short,

\begin{align*}
    \forall v_{k - i} \in \{ 0,1\}^{k - i}, \forall v_{i + 1} \in \{ 0,1\}^{i + 1}, f_*(uv_{k - i} v_{i + 1}) = f_*(u'v_{k - i} v_{i + 1})
\end{align*}

Thus, $\forall v_{k + 1} \in \{ 0,1\}^{k +1}$, $f_*(uv_{k + 1}) = f_*(u' v_{k + 1})$, completing the induction. We have proved the claim for all $k \geq i$.

However, this will lead to a contradiction if we take $k \to \infty$, since applying the guarantee for all $k \geq i$, we have that 

\begin{align*}
    \forall u, u' \in \{ 0,1\}^{\leq n}, E_i(u) = E_i(u') \implies E_\infty(u) = E_\infty(u')
\end{align*}

Where $E_\infty(u) := \{ v \in \{ \eps\}\cup \{ 0,1\}^* : uv \in L\}$. We know that any two distinct states in a minimal DFA will be distinguished by some finite string (See Theorem 4.24 of \cite{hopcroft-ullman}). Thus, for any $u, u' \in \{ 0,1\}^{\leq n}$, 

\begin{align*}
    E_i(u) &= E_i(u') \\
    \implies E_\infty(u) &= E_\infty(u')\\
    \implies  \delta(q_0, u) &= \delta(q_0, u')
\end{align*}

So $n = \# (\text{Equivalence classes in } \{ E_\infty(u)\}_{u \in \{ 0,1\}^{\leq n}}) = \# (\text{Equivalence classes in } \{ E_i(u)\}_{u \in \{ 0,1\}^{\leq n}}) < n $, where the last inequality is due to (1). This is a contradiction.  

Thus, while the number of equivalence classes in $\{ E_i(u)\}_{u \in \{ 0,1\}^{\leq n}} $ is less than $n,$ there must be some $u, u' \in \{ 0,1\}^{\leq n}$ and $v_{i + 1} \in \{ 0,1\}^{i + 1}$  such that

\begin{align*}
    E_i(u) &= E_i(u')\\
    \text{ and } f_*(uv_{i + 1}) &\neq f_*(u'v_{i + 1})
\end{align*}

This implies that the number of equivalence classes in $\{ E_{i + 1}(u)\}_{u \in \{ 0,1\}^{\leq n}} $ is at least 1 greater than the number of equivalence classes in $\{ E_i(u)\}_{u \in \{ 0,1\}^{\leq n}} $ while the latter is less than $n$.

Finally, $\{ E_{0}(u)\}_{u \in \{ 0,1\}^{\leq n}} $ has 2 equivalence classes for any non-trivial DFA that whose language isn't $\emptyset$ or $\{ 0,1\}^*$, and any $n \geq 2$ state minimal DFA must be non-trivial, as the minimal trivial DFAs have at most one state. The two equivalence classes of $\{ E_{0}(u)\}_{u \in \{ 0,1\}^{\leq n}} $ are given by the states that are accepting and those that are not. For all $i < n - 2$, the number of equivalence classes in $\{ E_{i + 1}(u)\}_{u \in \{ 0,1\}^{\leq n}} $ grows by at least 1 from that of $\{ E_i(u)\}_{u \in \{ 0,1\}^{\leq n}} $. We can have no more equivalence classes than $n$, and the $i$ where $\{E_i(u)\}_{u \in \{ 0,1\}^{\leq n}}$ has $n$ equivalence classes  will be at most $n - 2$. This implies Lemma \ref{lem:suff-cond-for-dfa-learning} is true if we let $E(u) = E_{n - 2}(u), \forall u \in \{ 0,1\}^{\leq n}$. $\blacksquare$

Now, we will prove an impossibility result for linear CFGs.

\cfglgen*
% Denote $\calf_{c} := \{ f \in \calf : \calc(f) \leq c\}$.

\textit{Proof of Proposition \ref{prop:cfg-len-gen}.} This follows directly from \Cref{lem:equiv-decidability} and Theorem \ref{lem:opt-learner}, with $\calr$ being an encoding system which maps string encodings of linear CFGs to the corresponding language, $\calf = \calf^\calr$ being the languages recognized by linear CFGs, and $\calc$ being the complexity measure described in \Cref{defn:cfg}. Because $\calr_{\lcfg}$ satisfies Assumption \ref{asmpt:E-exists}, \Cref{lem:equiv-decidability} and Theorem \ref{lem:opt-learner} imply that encoding system $\calf^{\calr}$ admits non-asymptotic length generalization w.r.t. $\calr, \calc$ iff the language equivalence problem for $\calr$ is decidable. However, by \cite{baker-book-linear-cfg}, the latter is undecidable.  $\blacksquare$

%% file: sections/appendices/crasp1.tex
\subsection{Proof of Theorem \ref{thm:crasp1}}\label{appen:crasp1}

\thmonelayer*

\textit{Proof.} With integer $T$ and integers $a,b,d \in [-T, T]$, $a > 0$, recall $\crasp{}^{1, T}$ is the set of functions of the form:
\begin{align*}
    f_{a,b,d}(x) = \ind [a \cdot \ps (x) -b \cdot n - d > 0]
\end{align*}

By \Cref{lem:opt-learner}, it is sufficient to show that for any tuples of integers $(a,b,d), (a', b', d') \in [-T, T]^3$ with $a, a' > 0$, if there exists an $x \in \{ 0,1\}^*$ with $f_{a,b,d}(x) \neq f_{a', b', d'}(x)$, then there exists an $x'\in \{ 0,1\}^*$ with $|x'| \leq O(T^2)$ such that $f_{a,b,d}(x') \neq f_{a', b', d'}(x')$. 

Suppose there is a string $x$ that distinguishes $f_{a,b,d} $ and $ f_{a', b', d'}$. We can rewrite $f$ and $f'$ as follows.

\begin{align*}
f(x) := f_{a,b,d}(x) &= \ind [\ps (x) - \frac{b}{a} \cdot n - \frac{d}{a} > 0]\\
f'(x) := f_{a',b',d'}(x) &= \ind [\ps (x) - \frac{b'}{a'} \cdot n  - \frac{d'}{a'} > 0]
\end{align*}

Since $f,f'$ differ on $x$, either $\frac{b}{a} \neq \frac{b'}{a'}$; or $\frac{b}{a} = \frac{b'}{a'}$ but $\frac{d}{a} \neq \frac{d'}{a'}$.

\textbf{Case 1: $\frac{b}{a} \neq \frac{b'}{a'}$}

There are 3 categories for the value of a slope $\frac{b}{a}$ (resp. $\frac{b'}{a'}$).

\begin{itemize}
    \item Type (i): $\frac{b}{a} \geq 1$ (resp. $\frac{b'}{a'} \geq 1$)
    \item Type (ii): $\frac{b}{a} \leq 0$ (resp. $\frac{b'}{a'} \leq 0$)
    \item Type (iii): $\frac{b}{a} \in (0,1)$ (resp. $\frac{b'}{a'} \in (0,1)$)
\end{itemize}

Below we will consider how to distinguish two functions $f,f'$ whose slopes $\frac{b}{a}, \frac{b'}{a'}$ are in each of the categories.

\begin{itemize}
    \item $\frac{b}{a}$ is type (iii); $\frac{b'}{a'}$ is either type (iii), (ii), or (i)

% Suppose WLOG that $\frac{b}{a}$ is type (iii) and $\frac{b'}{a'}$ is type (iii), (ii), or (i). 

For any two 2D lines $\yax = \frac{b}{a} \xax + \frac{d}{a} $ and $\yax = \frac{b'}{a'}\xax + \frac{d'}{a'}$, there is a lattice point of $\xax-$coordinate $\xax_* := |aa'| (2\max(|\frac{d}{a}|, |\frac{d'}{a'}|) + 1)$ that lies strictly above the line with smaller slope and un-strictly below the line with greater slope. This follows from the fact that when $\frac{b'}{a'} \neq \frac{b}{a}$, then the smallest that $|\frac{b'}{a'} - \frac{b}{a}|$ can be is $\frac{1}{|aa'|}$. Thus, $|\frac{b'}{a'} - \frac{b}{a}| \cdot |aa'| (2\max(|\frac{d}{a}|, |\frac{d'}{a'}|) + 1) \geq 2\max(|\frac{d}{a}|, |\frac{d'}{a'}|) + 1$ so that $|\frac{b}{a} \xax_* + \frac{d}{a}  - (\frac{b'}{a'} \xax_* + \frac{d'}{a'} )| \geq 1$. Because the vertical gap between the two lines at horizontal coordinate $\xax_*$ is at least 1, there is a lattice point $(\xax_*, \yax_*) \in \mathbb{Z}^2$ with $\yax_* \in (\min(\frac{b}{a}\xax_* + \frac{d}{a}, \frac{b'}{a'}\xax_* + \frac{d'}{a'}), \max(\frac{b}{a}\xax_* + \frac{d}{a}, \frac{b'}{a'}\xax_* + \frac{d'}{a'})]$. In fact, for any $\xax \geq \xax_*$, one can find such a lattice point with horizontal coordinate $\xax$, since the gap between the two lines will continue to grow as $\xax$ increases for $\xax \geq \xax_*$, and so the gap will always be at least $1$.

We want to pick such a lattice point $(\tilde{\xax}, \tilde{\yax})$ subject to four constraints: 

\begin{enumerate}
    \item $\tilde{\yax} \leq \max(\frac{b}{a} \tilde{\xax} + \frac{d}{a}, \frac{b'}{a'}\tilde{\xax} + \frac{d'}{a'})$
    \item $\tilde{\yax} > \min(\frac{b}{a} \tilde{\xax} + \frac{d}{a}, \frac{b'}{a'}\tilde{\xax} + \frac{d'}{a'})$
    \item $\tilde{\yax} \leq  \tilde{\xax}$
    \item $\tilde{\yax} \geq  0$
\end{enumerate}

Since $\frac{b}{a} \in (0,1)$, then it suffices to find such a lattice point either between $\yax = \frac{b}{a} \xax + \frac{d}{a}$ and $\yax = \xax $, between $\yax = \frac{b}{a} \xax + \frac{d}{a}$ and $\yax = 0$, or between $\yax = \frac{b}{a} \xax + \frac{d}{a}$ and $\yax = \frac{b'}{a'}\xax + \frac{d'}{a'}$. For each of the 3 cases, any lattice point between the two lines specified in that case will satisfy all four constraints. 

By the argument in the previous paragraph, for each of the three cases, we can construct such a lattice point with horizontal coordinate $\xax_*$, which will be at most the following, across the three cases.

\begin{align*}
    \tilde{\xax} &\leq \max\Big[|aa'| (2\max(|\frac{d}{a}|, |\frac{d'}{a'}|) + 1), |a\cdot 1| (2\max(|\frac{d}{a}|, 0) + 1), |a\cdot 1| (2\max(|\frac{d}{a}|, 0) + 1)\Big]\\
    &\leq 2\max(|da', |ad'|) + aa'\\
    &\leq 3T^2
\end{align*}

Denote the vertical coordinate of this lattice point as $0 \leq \tilde{\yax} \leq \tilde{\xax} $. Finally, any string $x' \in \{ 0,1\}^*$ with $|x'| = \tilde{\xax}$, consisting of  $\tilde{\yax} \geq 0$ ones and $\tilde{\xax} - \tilde{\yax} \geq 0$ zeros will distinguish $f$ and $f'$. This string will have length at most $3T^2$.

    \item $\frac{b}{a}$ and $\frac{b'}{a'}$ are type (ii)

    % Suppose $\frac{b}{a}$ and $\frac{b'}{a'}$ are type (ii). 

    Since $\frac{b}{a} \neq \frac{b'}{a'}$, say WLOG that $\frac{b}{a} < 0$. 

    If $\frac{b'}{a'} < 0$, then after $\xax$ coordinate at most $T + 1$, then lines $\{ \yax = \frac{b}{a} \xax + \frac{d}{a}, \yax = \frac{b'}{a'}\xax + \frac{d'}{a'}\}$ go below $\yax = 0$ and no lattice point of nonnegative $\yax$ coordinate can be below one but above the other (i.e. both $f,f'$ become the all-ones function on strings of length at least $T + 1$). Thus, any distinguisher of $f,f'$ (in particular, the $x$ presumed in the beginning of the proof of this Theorem) must have length at most $T$, and we take $x' = x$.

    If $\frac{b'}{a'}  = 0$, then in the case $\frac{d}{a} < 0$, the same argument implies that $|x| \leq T$.
    
    In the case that $\frac{d}{a} \geq 0$, then a lattice point which lies between the two lines is $(T + 1, 0)$. Thus, the string $x' = 0_{T + 1}$ distinguishes $f, f'$ and has length $T + 1$.

    \item $\frac{b}{a}$ is type (ii) and $\frac{b'}{a'}$ is type (i)

    The analysis of this case is similar to the Types (iii) versus (iii), (ii), (i) case. One can take the line of slope $\geq 1$ and modify just its slope to be between $(0,1)$, apply the analysis in the Types (iii) versus (iii), (ii), (i) case to attain a lattice point which lies below the original line of slope $\geq 1$ and above the line of slope $\leq 0$. Such a lattice point corresponds to a string of length at most $2T$ which distinguishes $f, f'$.
    
    % WLOG, let $f$ be the line is slope $\geq 1$ and $f'$ the one with slope $\leq 0$. One needs to take $n$ to be sufficiently large so as to be between 3 pairs of lines: $(f, f'), (f, ps(x)_n = 1)$ and $(f', ps(x)_n = n - 1)$.
    
    % As argued before, this is achieved with $n = |aa'| (2\max(|d/a|, |d'/a'|) + 1 \leq O(T^2)$. 

    \item $\frac{b}{a}$ and $\frac{b'}{a'}$ are type (i)
    
    Analogous to the argument about Type (ii) versus (ii). There will be a string of length at most $T + 1$ which distinguishes $f, f'$.
    % Since their slopes are different, at least one function, say $f$ has slope $> 1$. If the two functions are different, they will disagree on a point whose $n$-coordinate is at most the maximum between intersection $f, f'$ with the line $ps(x)_n = n - 1$ and $|aa'| (2\max(|d/a|, |d'/a'|) + 1)$. As the intercept is at most $|\frac{d/a}{b/a - 1}| \leq T^2$, if there is a distinguishing string, then it has length at most $\max(|d/b|, |aa'| (2\max(|d/a|, |d'/a'|) + 1)) \leq O(T^2)$
    
\end{itemize}

\textbf{Case 2: $\frac{b}{a} = \frac{b'}{a'}$ but $\frac{d}{a} \neq \frac{d'}{a'}$} If the slopes are at least $1$ or at most $ 0$, an analogous argument as Case 1, Type (ii) versus (ii) or (i) versus (i) will suffice, where it must be the case that the string $x $ which was presumed to distinguish $f, f'$ has length at most $T + 1$.

Otherwise, if both slopes are in $(0,1)$, then we can take a lattice point $(\xax_*, \yax_*)$ which lies on the one of larger intercept. Such lattice points occur periodically with spacing $a \leq T$, and we need only take one such lattice point where $\xax_* \geq \yax_* \geq 0$. We can find one such lattice point where $\xax_* \leq 3T$. $\blacksquare$

% Suppose the two linear functions disagree at some $x$. If the lines have slope 0, then there is is an $x$ of length 1 that distinguishes them. If the slopes are positive, then there is an $x$ of length at most $T^2$ that distinguishes them, since $x$ with sum and length $(s, n)$ can be transported to $(s - b, n - a)$ until it hits an axis or the line $s = n$. If the lines are negative, then any $x$ distinguishing them must have length at most the maximum $n-$intercept of one of the lines, which is $T$. In all cases, there is an $x$ of length $\leq T^2$ distinguishing them. 

%% file: sections/appendices/main-proof.tex
\paragraph{Notation and Conventions.} For the remaining sections, we will use lower-case $x$ to denote bit-strings in $\{ 0,1\}^*$. 

We define $[k] := \{ 1, 2, \ldots, k - 1, k\}$ as the first $k$ positive integers.

We will use symbol $f$ to denote functions in the relevant hypothesis class, like \crasp{} or DFAs. These will be mappings from $\{ 0,1\}^*$ to $\{ 0,1\}$. We will sometimes refer to functions in $\crasp{}{}^{1}$ and $\crasp{}{}^{2}$ as programs, which is synonymous with ``functions." We will say that two functions are equal if they agree on all strings $\{ 0,1\}^*$, and that they are unequal if they are distinguished by at least one string $x \in \{ 0,1\}^*$. We will call such an $x$ a distinguisher of the two functions. To be clear, the word function here has a distinct meaning and type from test-functions. 

We will use $\calx$ to denote discrete test-functions and $\caly$ to denote continuous test-functions. When we say ``test-function" without specifying whether it is discrete or continuous, assume we mean a continuous test-function. We will use $(B_i(\caly))_{i \in [k]}$ (or $(B_i)_{i \in [k]}$ when clear from context) to represent the activations induced by a continuous test-function (see \Cref{defn:cnts-test-fn}). We will use symbol $Y$ to denote a schema of continuous test-functions (see \Cref{defn:schema}). When we talk about a 2D coordinate system, in the context of test-functions, we will use symbol $\xax$ to denote the horizontal axis of this 2D coordinate system and $\yax$ to denote the vertical axis. To be clear, $\xax$ is distinct from the symbol $x$, which we use to denote bit-strings.

Regarding geometric objects, denote $\clo(S)$ as the closure of a set $S$. Denote $S^c$ as the complement of set $S$. For a convex set $S$, the affine hull of $S$ is the set of linear combinations of points in $S$. Denote $\dim(S)$ as the dimension of the affine hull of $S$. Denote the interior of a convex set $S$ as $\intr(S)$ and the relative interior of convex set $S$ as $\ri(S)$, which is the set of points in $S$ such that there is some non-zero radius such that the Euclidean ball centered at that point with that radius, intersected with the affine hull of $S$, is contained in $S$. Finally, sometimes we will talk about a geometrical set in $\mathbb{R}^k$ and its ``analog" in $\mathbb{R}^M$, where $M > k$ (what ``analog" means, we won't go into here). Notationally, if we use symbol $A \in \mathbb{R}^k$ for the set in $\mathbb{R}^k$, we will use symbol $A^{(M)}$ to denote a set that's analogous to $A$, but in $\mathbb{R}^M$.

A halfspace of $\mathbb{R}^d$ is a subset of $\mathbb{R}^d$ which satisfy a linear inequality over the $d$ coordinates of $\mathbb{R}^d$. A polytope is the intersection of a finite number of halfspaces. The polytopes we will be working with will be restricted to $[0,1]^d$ for some dimension $d$. The faces of a polytope $P \subset [0,1]^d$ refer to the $d - 1$ dimensional polytopes which form the boundary $P$. Each face is associated with a linear inequality (halfspace) which defines the face in the sense that points on the face satisfy the linear inequality with equality. A $d$-dimensional simplex is a $d$-dimensional polytope with $d + 1$ vertices and $d + 1$ faces.

\subsection{Proof of Theorem \ref{thm:main}.}\label{appen:full-pf-thm}

We reiterate the definition of $\crasp{}^{2}$ here.

\twolayercrasp*

We will prove the following length generalization guarantee.

\thmmain*

Note that Theorem \ref{thm:main} is actually  stronger than a result which says that we can learn $\crasp{}^{2,K,T}$ with length $O(T^{O(K)})$. This is because for a fixed $T$ and $K$, $\crasp{}^{2,K,T}$ only contains functions of precision at most $T$ and at most $K$ heads, whereas Theorem \ref{thm:main} also provides length generalization guarantees for $\crasp{}^2$ functions $f$ with $T(f) > T, K(f) < K$ or $T(f) < T, K(f) > K$  such that $T(f)^{K(f)} \leq T^K$. % Phrasing the Theorem statement like this also gives us an end-to-end learning guarantee, where the Minimum Complexity Interpolator equipped with complexity measure $\calc(f) = T(f)^{K(f)}$ can learn $\crasp^2$.
\\
\\
\textit{Proof of Theorem \ref{thm:main}.} By  \Cref{lem:opt-learner}, to upper bound $N_{\cala{\textup{mci}}}(\alpha)$, it suffices to bound:

\begin{align*}
    \min \{ n \in \nat : \forall f \neq f' \in \crasp{}^{2} \text{ with } \calc(f), \calc(f') \leq \alpha, \text{ exists } x \in \{ 0,1\}^{\leq n} \text{ s.t. } f(x) \neq f'(x)\}
\end{align*}

Suppose $f, f' \in \crasp{}^{2}$ are not equal and differ on $x_0 \in \{ 0,1\}^*$. For any $n > 0$, suppose $f$ and $f'$ have the following 2-layer form on an arbitrary input $x \in \{ 0,1\}^n$. 

\begin{align*}
    \forall j \in [n], \forall i \in [K],\quad  h_j^{(i)}(x) &= \ind [\ps(x)_j > \frac{b^{(i)}}{a^{(i)}} j]\\
        f(x) &= \ind [\sum_{i \in [K]} \lambda_i \ps(h^{(i)}(x))_n > z \cdot n]\\
    \forall j \in [n], \forall i \in [K'],\quad (h^{(i)})'_j(x) &= \ind [\ps(x)_j > \frac{(b^{(i)})'}{(a^{(i)})'} j]\\
    f'(x) &= \ind [\sum_{i \in [K']} \lambda_i' \ps((h^{(i)})'(x))_n > z' \cdot n]\\
\end{align*}

Where $f$ has $K$ first-layer heads and integer parameters of precision $T$ and $f'$ has $K'$ first-layer heads and integer parameters of precision $T'$, where $C(f) = T^K \leq \alpha$ and $C(f') = (T')^{K'} \leq \alpha$, and $K \leq T^2$ and $K' \leq (T')^2$. WLOG, suppose that $T \geq T'$. We will find a short string $x_* \in \{ 0,1\}^{O(\alpha^{O(1)})}$ such that $f(x_*) \neq  f'(x_*)$. 

Suppose the set of unique slopes $R := \{ \frac{b^{(i)}}{a^{(i)}}\}_{i \in [K]} \cup \{ \frac{(b^{(i)})'}{(a^{(i)})'}\}_{i \in [K']} \subset (0,1)$ between the first layer of $f$ and $ f'$ has size $k,$ where $\max(K, K') \leq k \leq K + K'$. We will denote $R = \{ s_j\}_{j \in [k]} \subset (0,1)$, where $s_1$ is the largest slope and $s_k$ is the smallest slope, and the slopes are sorted in descending order so that $s_1 > \ldots > s_k$. Let $\textup{ord}(1, i) : [K] \to [k]$ be the index within $R$ of the $i$th slope of $f$, $\frac{b^{(i)}}{a^{(i)}}$. Let $\textup{ord}(2, i) : [K'] \to [k]$ be the index within $R$ of the $i$th slope of $f'$, $\frac{(b^{(i)})'}{(a^{(i)})'}$.  In the following exposition, we will refer to ``line $i$" as the homogeneous, 2D line $\yax = s_i \xax$, with slope $s_i, i \in [k]$.

% changed the lemma to the stronger one; will simply change $M$ in the last part of the proof to O(K).
\Cref{defn:crasp21} requires that $z, z' > 0$, and that $\sum_{i \in [K]}\lambda_i > z$ and $\sum_{i \in [K']}\lambda_i' > z'$. Since $f$ differs from $f'$ on discrete test-function given by $x_0$, then the non-trivial Lemma \ref{lem:stronger-suffcond-for-asmpt} implies that there exists a continuous test-function $\caly_1$ that induces activations $(B_1(\caly_1),\ldots, B_k(\caly_1))$ which satisfies either Case I or Case II.

% \Cref{defn:crasp21} requires that the bias term in the second layer of each function in $\crasp{}^{2,K,T}$ is positive: $z > 0$, and that $\sum_{i \in [K]}\lambda_i > z$. Since $f$ differs from $f'$ on discrete test-function given by $x_0$, then Lemma \ref{lem:suffcond-for-asmpt} implies that there exists a continuous test-function $\caly_1$ that induces activations $(B_1(\caly_1),\ldots, B_k(\caly_1))$ which satisfies:

\begin{align*}
    \text{Case I: }\sum_{i \in [K]} \lambda_i B_{\textup{ord}(1, i)}(\caly_1) > z \text{ and } \sum_{i \in [K']} \lambda_i' B_{\textup{ord}(2, i)}(\caly_1) < z'\\
    \text{Case II: } \sum_{i \in [K]} \lambda_i B_{\textup{ord}(1, i)}(\caly_1) < z \text{ and } \sum_{i \in [K']} \lambda_i' B_{\textup{ord}(2, i)}(\caly_1) > z'\\
\end{align*}

For the remainder of the proof, we will suppose Case I is true. The proof for Case II is entirely analogous to what we will present below, since we will not use the direction of the signs of the two halfspaces in the following proof. Denote the two halfspaces induced by the second-layer of $f$ and $f'$ by $H_1$ and $H_2$ respectively.

\begin{align*}
    H_1 := \{ B \in \mathbb{R}^k : \sum_{i \in [K]} \lambda_i B_{\textup{ord}(1, i)} > z\}\\
    H_2 := \{ B \in \mathbb{R}^k : \sum_{i \in [K']} \lambda_i' B_{\textup{ord}(2, i)} < z'\}\\
\end{align*}

By Corollary \ref{cor:cleaner-basis-test-function} (Completeness of Basis Schema), there exists a continuous test-function $\caly_2$ of a basis test-function schema, $Y$, specified in Corollary \ref{cor:cleaner-basis-test-function}, such that $(B_1(\caly_2), \ldots, B_k(\caly_2)) = (B_1(\caly_1), \ldots, B_k(\caly_1))$. Thus,

\begin{align*}
    (B_1(\caly_2), \ldots, B_k(\caly_2)) \in H_1 \cap H_2
\end{align*}

% New paragraph.
Let $M$ be the number of segments in the basis schema $Y$ of $\caly_2$. Note that our previous application of Lemma \ref{lem:stronger-suffcond-for-asmpt} guarantees that $Y$ is a basis schema of either one or two monotone curves (see \Cref{defn:monotone-curve}), ensuring that

\begin{align}
    M \leq 2k \label{eq:bound-of-M}
\end{align} 

as opposed to the naive bound of $M \leq k^2$ via Corollary \ref{cor:size-of-M}. This fact will be useful at the end for achieving a better final bound. % For now, we keep expressions in terms of $M$.

Denote the lengths of the $M$ segments of schema $Y$ as $(n_1,\ldots ,n_M) \in [0,1]^M$. Let $A^{(M)}(Y)$ be the set of valid segment lengths $(n_1,\ldots,n_M)$ for a continuous test-functions of schema $Y$, where a particular setting of $(n_1,\ldots ,n_M)$ is valid if it obeys the constraints described by Lemma \ref{lem:schema-constraints} for schema $Y$ and $\sum_{i \in [M]} n_i = 1$.

\begin{align*}
    A^{(M)}(Y) &:= \{ (n_1,\ldots ,n_M) : \text{ valid segment lengths of schema $Y$ and } \sum_{i \in [M]} n_i = 1\} \subset [0,1]^M
\end{align*}

Additionally, define $A(Y)$ as the analogous set to $A^{(M)}(Y)$ in the space of activations rather than the space of segment lengths.

\begin{align*}
    A(Y) &:= \{ (B_1(\caly),\ldots ,B_k(\caly)) : \caly \text{ valid test-function of schema $Y$ }\} \subset [0,1]^k
\end{align*}

There exists a linear map $L : \mathbb{R}^M \to \mathbb{R}^k$ which maps points in $A^{(M)}(Y)$ to points in $A(Y)$. $L \in \{ 0,1\}^{k \times M}$ is such that $L_{ij} = 1 \iff $ segment $j$ in schema $Y$ lies above line $i$ (that is, for every $\xax $ in the domain of segment $j$, the $\yax$-value of the segment at $\xax$ is at least $s_i \cdot \xax$) and hence contributes to the $i$th activation $B_i(\caly)$ of any test-function $\caly$ of schema $Y$. Using $L$,  we can rewrite the inequalities which characterize $H_1, H_2$ in terms of $(n_1,\ldots, n_M)$.

\begin{align*}
 H_1^{(M)} := \{(n_1,\ldots,n_M) :  \sum_{i = 1}^{K} \lambda_i B_{\textup{ord}(1, i)}  > z\} &= \{(n_1,\ldots,n_M) :  \sum_{i = 1}^{K} \lambda_i \sum_{j \text{ s.t. } L_{\textup{ord}(1, i), j} = 1} n_j  > z\}\\
   H_2^{(M)} := \{(n_1,\ldots,n_M) :  \sum_{i = 1}^{K'} \lambda_i' B_{\textup{ord}(2, i)}  < z'\} &= \{(n_1,\ldots,n_M) :  \sum_{i = 1}^{K'} \lambda_i' \sum_{j \text{ s.t. }  L_{\textup{ord}(2, i), j} = 1} n_j  < z'\}
\end{align*}

%That is, we can think of $H_1^{(M)}, H_2^{(M)}$ as a halfspace in $\mathbb{R}^M$ given by the two inequalities above for $H_1^{(M)}, H_2^{(M)}$, respectively. 

Since $(\lambda_i)_{i \in [K]}, z$ are integers at most $T$ in magnitude and $(\lambda_i')_{i \in [K']}, z'$  are integers at most $T'$ in magnitude, the coefficients of the linear inequality for $H_1^{(M)}$ (resp. $H_2^{(M)}$), $\sum_{i = 1}^{K} \lambda_i \sum_{j \text{ s.t. } L_{\textup{ord}(1, i), j} = 1} n_j  > z$ (resp. $\sum_{i = 1}^{K'} \lambda_i' \sum_{j \text{ s.t. }  L_{\textup{ord}(2, i), j} = 1} n_j  < z'$) are integers of at most $KT \leq T^3$ (resp. $K'T' \leq (T')^3$) in magnitude, since for each $j \in [M]$,  at most every $i \in [K]$ (resp. $i \in [K']$) can contribute to the $j$th coefficient.

Now, we describe a few more properties of $A^{(M)}(Y)$. Suppose the segment lengths of $\caly_2$ in schema $Y$ is $(n_1(\caly_2),\ldots,n_M(\caly_2 )) \in [0,1]^M$. Then, we have:

\begin{align*}
    (n_1(\caly_2),\ldots,n_M(\caly_2) ) &\in A^{(M)}(Y) \cap H_1^{(M)} \cap H_2^{(M)} \neq \emptyset
\end{align*}

By Lemma \ref{lem:convexity-of-schema-activations} $A^{(M)}(Y)$ is a polytope: the intersection of a finite number of halfspaces. It follows that $A^{(M)}(Y)$ is convex. Moreover, by Lemma \ref{lem:convexity-of-schema-activations}, we have $\dim(A^{(M)}(Y)) = M - 1$. Because $A^{(M)}(Y) \cap H_1^{(M)} \cap H_2^{(M)} \neq \emptyset$ and $H_1^{(M)},$ $H_2^{(M)}$ are open sets of dimension $M$, then by Lemma \ref{lem:open-hs-intersection}, $\dim(A^{(M)}(Y) \cap H_1^{(M)} \cap H_2^{(M)}) = \dim(A^{(M)}(Y)) = M - 1$. 

We have that $\clo(A^{(M)}(Y) \cap H_1^{(M)} \cap H_2^{(M)})$ is the intersection of a finite number of halfspaces. Each face of the polytope $\clo(A^{(M)}(Y) \cap H_1^{(M)} \cap H_2^{(M)})$ is defined by one of these halfspaces. We now discuss the precision of the linear inequalities which define the faces of $\clo(A^{(M)}(Y) \cap H_1^{(M)} \cap H_2^{(M)})$, where precision of an linear inequality with integer coefficients is the maximum magnitude of the integer coefficients, per \Cref{defn:precision}. The linear inequalities of the halfspaces which form the faces of $A^{(M)}(Y)$ are such that there is a subset of at most $6K$ of them with precision at most $T^2$, while the remaining faces of  $A^{(M)}(Y)$ have precision at most $(T')^2$. This is due to the following argument. First, because the $(k,T)$-configuration $\{ s_i\}_{i \in [k]}$ is such that there is a subset of at most $K$ of the $k$ elements of $\{ s_i\}_{i \in [k]}$ which are precision at most $T$, while the rest of $\{ s_i\}_{i \in [k]}$ are precision at most $T'$. Next, referring to Lemma \ref{lem:schema-constraints}, the only faces of $A^{(M)}(Y)$ with $T^2$ precision are ones that correspond to a segment of $Y$ whose start-point or end-point is on one of the $K$ lines of slope whose precision is $T^2$. Third, by Corollary \ref{cor:cleaner-basis-test-function}, any basis schema of one or two monotone curves will cross each of the $k$ lines at most three times. Since each of the (at most) $K$ slopes of precision $T$ correspond to at most $3$ crossing-points of $Y$ with that the line of that slope, and each crossing point is adjacent to at most two segments of $Y$, then there are at most $2 \cdot 3K$ segments of $Y$ such that the start-point or end-point is on a line of slope that is precision $T$. Thus, at most $6K$ of the faces of $A^{(M)}(Y)$ are defined by linear inequalities of precision at most $T^2$, while the remaining faces of  $A^{(M)}(Y)$ are defined by linear inequalities of precision at most $(T')^2$. Note that the square (i.e. in $T^2$ and $(T')^2$) comes from the form of the inequalities defining the faces of $A^{(M)}(Y)$, stated in Lemma \ref{lem:schema-constraints}. Finally, as argued in an earlier paragraph, the face given by $H_1^{(M)}$ has precision at most $T^3$ while that given by $H_2^{(M)}$ has precision at most $(T')^3$. In summary, there is a subset of at most $7K$ faces of $\clo(A^{(M)}(Y) \cap H_1^{(M)} \cap H_2^{(M)})$ such that each face of the subset has precision at most $T^3$ while the faces not in the subset all have precision at most $(T')^3$. In short, we've shown that polytope $\clo(A^{(M)}(Y) \cap H_1^{(M)} \cap H_2^{(M)})$ satisfies the pre-conditions of Corollary \ref{cor:margin-of-C}, which we will apply in the next step.

Now, we return to the process of converting $\caly_2$ into a short distinguisher of $f, f'$. Let $V$ denote the set of vertices of the polytope $\clo(A^{(M)}(Y) \cap H_1^{(M)} \cap H_2^{(M)})$. Let $c \in [0,1]^M$ be the average of the vertices in $V$.

\begin{align*}
    c = \frac{1}{|V|}\sum_{x \in V} x
\end{align*}

Label the coordinates of $c$ as $c := (n_1^{(c)}, \ldots, n_M^{(c)}) $. $c$ is in the relative interior of $\clo(A^{(M)}(Y) \cap H_1^{(M)} \cap H_2^{(M)})$ (which is non-empty by Lemma \ref{lem:nonempty-ri}) so that $c \in A^{(M)}(Y) \cap H_1^{(M)} \cap H_2^{(M)}$. In particular,
% Since $A^{(M)}(Y) \cap H_1^{(M)} \cap H_2^{(M)}$ is an $(M - 1)$-dimensional convex polytope, by applying Lemma \ref{lem:caratheodory} on the set of vertices of the polytope $A^{(M)}(Y) \cap H_1^{(M)} \cap H_2^{(M)}$, there will exist some  $(M - 1)$-simplex, $S \subset [0,1]^M$, whose $M$ vertices are a subset of the vertices of $\clo(A^{(M)}(Y) \cap H_1^{(M)} \cap H_2^{(M)})$ and such that $(n_1(\caly_2),\ldots,n_M(\caly_2)) \in S$. By Lemma \ref{lem:nonempty-ri}, simplex $S$ will have a nonempty relative interior because it is a nonempty $(M-1)$-dimensional convex polytope. As such, denote $c$ as the centroid of $S$, and let $\caly_c$ denote the continuous test-function of schema $Y$ that has segment lengths given by $c$. We have $c := (n_1^{c}, \ldots, n_M^{c}) \in \ri(S) \subset A^{(M)}(Y) \cap H_1^{(M)} \cap H_2^{(M)}$. As such, $c$ will satisfy:

\begin{align*}
    \sum_{i = 1}^{K} \lambda_i \sum_{j \text{ s.t. } L_{\textup{ord}(1, i), j} = 1} n_j^{(c)}  > z\\
    \sum_{i = 1}^{K'} \lambda_i' \sum_{j \text{ s.t. } L_{\textup{ord}(2, i), j} = 1} n_j^{(c)}  < z'
\end{align*}

We will now lower bound the margin (see \Cref{defn:margin}) of $c$ to the faces of $\clo(A^{(M)}(Y) \cap H_1^{(M)} \cap H_2^{(M)})$, which includes the faces given by $H_1^{(M)}, H_2^{(M)}$. Suppose $\clo(A^{(M)}(Y) \cap H_1^{(M)} \cap H_2^{(M)})$ has $N$ faces, and let $\{ L_i\}_{i \in [N]}$ denote the linear inequalities which define each face of $\clo(A^{(M)}(Y) \cap H_1^{(M)} \cap H_2^{(M)})$, so that $L_i(c) \in \mathbb{R}$ is the non-negative margin of $c$ on the $i$th face. Noting that $M \leq 2k \leq 2(K + K')$, we apply Corollary \ref{cor:margin-of-C} on polytope $\clo(A^{(M)}(Y) \cap H_1^{(M)} \cap H_2^{(M)})$ to get the following lower bound on the margin of $c$ on the faces of $\clo(A^{(M)}(Y) \cap H_1^{(M)} \cap H_2^{(M)})$.

\begin{align*}
\forall i \in [N], L_i(c) \geq \Omega(\frac{1}{|V|}\frac{1}{\alpha^{O(1)}})
\end{align*}

Then by Lemma \ref{lem:num-vtxs-of-P}, we have  $|V| \leq 3M^2$, so we deduce that:

\begin{align*}
\forall i \in [N], L_i(c) \geq \Omega(\frac{1}{M^2}\frac{1}{\alpha^{O(1)}})
\end{align*}

In particular, the margins, $\gamma_1, \gamma_2$ of $c$ on the two inequalities defining $H_1^{(M)}$ and $H_2^{(M)}$ will be at least:

\begin{align*}
    \gamma_1 &:= \sum_{i = 1}^{K} \lambda_i \sum_{j \text{ s.t. } L_{\textup{ord}(1, i), j} = 1} n_j^{(c)}  - z\\
    &\geq \Omega(\frac{1}{M^2}\frac{1}{\alpha^{O(1)}})\\
    \gamma_2 &:= z' - \sum_{i = 1}^{K'} \lambda_i' \sum_{j \text{ s.t. } L_{\textup{ord}(2, i), j} = 1} n_j^{(c)}\\
    &\geq \Omega(\frac{1}{M^2}\frac{1}{\alpha^{O(1)}})
\end{align*}

Now that we have shown that $c$ has large margin, we need to augment it one final time before converting it into a discrete test-function. This process will find a nearby point $c_* \in \textup{Ball}^\infty_{r}(c) \cap \{ \sum_{i \in [M]} n_i = 1\} := \{ (n_1, \ldots, n_M) \in \mathbb{R^M} : ||(n_1, \ldots, n_M) - c||_\infty \leq r\} \cap \{ \sum_{i \in [M]} n_i = 1\}$ such that $c_*$ has both large margin to the faces of $\clo(A^{(M)}(Y) \cap H_1^{(M)} \cap H_2^{(M)})$ and low-precision coordinates. With $||L_i||_1$ denoting the sum of the magnitudes of the coefficients of the linear inequality which defines the $i$th face of $\clo(A^{(M)}(Y) \cap H_1^{(M)} \cap H_2^{(M)})$, let:

\begin{align*}
    \gamma_{LB} &:= \frac{1}{\ceil{O(M^2 \alpha^{O(1)})}} \leq \gamma_1, \gamma_2\\
    r &:= \frac{\gamma_{LB}}{2 \cdot \ceil{\max_{i \in [N]} ||L_i||_1}}
\end{align*}

By linearity of the margin of a point in its coordinates $(n_1, \ldots, n_M)$, each point $c' \in \textup{Ball}^\infty_{r}(c) \cap \{ \sum_{i \in [M]} n_i = 1\}$ will have margin at least $\gamma_{LB} - r \cdot \max_{i \in [N]} ||L_i||_1 \geq \frac{\gamma_{LB}}{2}$ to each face of $\clo(A^{(M)}(Y) \cap H^{(M)}_1 \cap H^{(M)}_2)$. This also means that every such $c'$ is contained in $A^{(M)}(Y) \cap H^{(M)}_1 \cap H^{(M)}_2$ as $\gamma_{LB} > 0$.

By Lemma \ref{lem:l-inf-ball-precision}, there exists a low-precision point $c_* \in \textup{Ball}^\infty_{r}(c) \cap \{ \sum_{i \in [M]} n_i = 1\}$, denoted $c_* := (n_1^{(c_*)},\ldots,n_M^{(c_*)})$, such that for all $i \in [M]$ $n_i^{(c_*)}$ is a rational number, and the least common denominator of all elements in the tuple $(n_1^{(c_*)},\ldots,n_M^{(c_*)})$ is $p_{c_*}$, where:

\begin{align*}
    p_{c_*} &\leq \ceil{M \cdot \frac{1}{r}}\\
    &\leq O(M \ceil{\max_{i \in [N]} ||L_i||_1} \cdot \ceil{3M^2 \alpha^{O(1)}})\\
    &\leq O(M^3 \cdot (M\cdot  T^3) \cdot \alpha^{O(1)})
\end{align*}

Let the tuple $c_* = (n_1^{(c_*)},\ldots,n_M^{(c_*)})$ be the segment lengths of the continuous test-function $\caly_*$ of schema $Y$. Note that $c_*$ has positive margin $\frac{\gamma_{LB}}{2}$ to all the faces of $\clo(A^{(M)}(Y) \cap H^{(M)}_1 \cap H^{(M)}_2)$ and is contained in $A^{(M)}(Y) \cap H^{(M)}_1 \cap H^{(M)}_2$, so $c_*$ is a valid setting of segment lengths of schema $Y$, respecting the constraints of Lemma \ref{lem:schema-constraints}.

Finally, applying Lemma \ref{lem:low-prec-activ-to-string} on the continuous test-function of schema $Y$ with segment lengths $c_* = (n_j^{(c_*)})_{j \in [M]}$, we deduce that there exists a $n_0 \leq O(p_{c_*} \cdot \alpha^{O(1)})$ so that for any integer multiple $n$ of $n_0$, there exists a discrete test-function $\calx_*: \{ 0, \ldots, n\} \to \{ 0, \ldots, n\}$, of length $n$, corresponding to string $x_* \in \{ 0,1\}^n$ of length $n$, such that

\begin{align*}
    \forall i \in [k], |B_i(\caly_*) - B_i(\calx_*)| \leq \frac{T^2 + M}{n}
\end{align*}

Where $( B_i(\caly_*))_{i \in [k]}$ are the activations induced by $\caly_*$, a test-function of schema $Y$  with segment lengths $( n_i^{(c_*)})_{i \in [M]}$, and $( B_i(\calx_*))_{i \in [k]}$ the activations induced by $\calx_*$. Because $\forall i, |\lambda_i| \leq T, |\lambda_i'| \leq T' \leq T$, then the difference between the margin of $( B_i(\caly_*))_{i \in [k]}$ and $( B_i(\calx_*))_{i \in [k]}$ on $H_1$ and $H_2$ can be bounded by a term proportional to $\frac{1}{n}$.

\begin{align}
    &| \sum_{i = 1}^{K} \lambda_i B_{\textup{ord}(1, i)}(\caly_*) - \sum_{i = 1}^{K} \lambda_i B_{\textup{ord}(1, i)}(\calx_*)| \\
    &\leq (\max_{i \in [k]} | B_i(\caly_*) - B_i(\calx_*)| ) \cdot (\max_{\forall i \in [k], |\lambda_i| \leq T} \sum_{i = 1}^K \lambda_i)\\
    &\leq \frac{(T^2 + M)KT}{n}\\
    &\leq O(\frac{K M T^3}{n})
\end{align}

\iffalse
&\leq (\frac{T^2}{n}) \cdot KT + 
    |\sum_{i = 1}^{K} \lambda_i \sum_{j \in [M_0, M] \text{ s.t. } L_{\textup{ord}(1, i), j} = 1} n_j^{(c_*)} - \sum_{i = 1}^{K} \lambda_i \sum_{j \in [M_0, M] \text{ s.t. } L_{\textup{ord}(1, i), j} = 1} \frac{n_j(\calx_*)}{n}| \label{eq:bound-away-T2}\\
    &\leq \frac{KT^3}{n} + (\max_{j \in [M_0, M]} |n_j^{(c_*)} - \frac{n_j(\calx_*)}{n}|) \cdot  (\max |\sum_{i = 1}^{K}  \sum_{j \in [M_0, M] \text{ s.t. } L_{\textup{ord}(1, i), j} = 1} \lambda_i |)\\
    &\leq \frac{KT^3}{n} + \frac{T^2}{n} \cdot T \cdot K \cdot M \\
\fi 

% Where Equation \ref{eq:bound-away-T2} follows by first considering the maximum deviation which can be induced by the difference between the (normalized) $\calx_*$ and $\caly_*$ on the interval $[0, \frac{T^2}{n}]$ where there is no approximation guarantee. Because each segment in a test-function can contribute to, at most, every activation $B_1, \ldots, B_k$, then the maximum contribution to the final sum due to the test-functions' segments on $[0, \frac{T^2}{n}]$ is at most the length of the segment times $K$ (for $B_1, \ldots, B_k)$ times $T$ (as $|\lambda_i| \leq T$). This accounts for the first term. The second term just follows by rewriting in terms of the segments $M_0, \ldots, M$ of the schema $Y$ using $L$. 

Using an analogous argument, the difference between $\sum_{i = 1}^{K'} \lambda_i' B_{\textup{ord}(2, i)}(\caly_*)$ and $\sum_{i = 1}^{K'} \lambda_i' B_{\textup{ord}(2, i)}(\calx_*)$ is also bounded by an analogous expression.

\begin{align*}
    | \sum_{i = 1}^{K'} \lambda_i' B_{\textup{ord}(2, i)}(\caly_*) - \sum_{i = 1}^{K'} \lambda_i' B_{\textup{ord}(2, i)}(\calx_*)| \leq O(\frac{K' M T^3}{n})
\end{align*}

% Recall that $\sum_{i = 1}^{K} \lambda_i \sum_{j \text{ s.t. } L_{ord(0, i), j} = 1} n_j^{c} = \sum_{i = 1}^{K} \lambda_i B_{ord(0,i)}^c$ and $\sum_{i = 1}^{K} \lambda_i' \sum_{j \text{ s.t. } L_{ord(1, i), j} = 1} n_j^{c} = \sum_{i = 1}^{K} \lambda_i' B_{ord(1,i)}^c$. 

Together, these imply that for sufficiently large $n$, the difference in the margin of $(B_i(\calx_*))_{i \in [k]}$ and $(B_i(\caly_*))_{i \in [k]}$ on $H_1$ and $H_2$, caused by the discrete test-function approximation, will be smaller than the margin of $(B_i(\caly_*))_{i \in [k]}$ on $H_1$ and $H_2$. The latter equals the margin of $c_*$ on $H_1^{(M)}$ and $H_2^{(M)}$, by the definition of $H_1^{(M)}$ and $H_2^{(M)}$ as the analogous halfspaces to $H_1$ and $H_2$, which is lower bounded by $\frac{\gamma_{LB}}{2}$. More precisely,

% As a result, $( B_i(\calx_*))_{i \in [k]}$ will be in $H_1 \cap H_2$ for sufficiently large $n$, as $( B_i(\caly_*))_{i \in [k]}$ is in $H_1\cap H_2$ with margin at least $\frac{\gamma_{LB}}{2}$ since $\{ n_i^{(c_*)}\}_{i \in [M]}$ has margin at least $\frac{\gamma_{LB}}{2}$ on the equivalent constraints over $(n_1, \ldots, n_M)$ as argued before. In particular,

\begin{align*}
    \sum_{i = 1}^{K} \lambda_i B_i(\calx_*)  &\geq z + \frac{\gamma_{LB}}{2} - O(\frac{K M T^3}{n})\\
    \sum_{i = 1}^{K'} \lambda_i' B_i(\calx_*)  &\leq z' - \frac{\gamma_{LB}}{2} + O(\frac{K' M T^3}{n})
\end{align*}

To this end, since $\gamma_{LB} = \Omega(\frac{1}{M^2\alpha^{O(1)}})$, it suffices for $n$ to be the following value in order for activations $( B_i(\calx_*) )_{i \in [k]}$ induced by $\calx_*$ to be in $H_1 \cap H_2$ (and therefore to cause $f$ and $f'$ to differ, as $\caly_1$ and $\caly_*$ do). 

\begin{align}
    \Omega(\frac{1}{M^2 \alpha^{O(1)}}) - O(\frac{\max(K,K') M T^3}{n}) > 0 \iff n > O(M^3 \max(K,K') T^3 \alpha^{O(1)})\label{eq:overcome-margin}
\end{align}

Let $x_*$ be the string of length $n$ corresponding to $\calx_*$, which is uniquely determined by $\calx_*$. For $n = \max(1 + O(M^3 \max(K,K') T^3 \alpha^{O(1)}), O(p_{c_*} \cdot \alpha^{O(1)}))$, $n$ will be sufficiently large to make the approximation error smaller than the margin of $c_*$ per Equation \ref{eq:overcome-margin}, and also satisfy the condition required to apply Lemma \ref{lem:low-prec-activ-to-string} in the previous part of this proof. Thus, with this value of $n$, $x_*$ will cause $f(x_*) = 1, f'(x_*) = 0$.

We noted previously in Equation (\ref{eq:bound-of-M}) that $M \leq 2k \leq 2(K + K')$ for the basis schema $Y$ as a result of Lemma \ref{lem:stronger-suffcond-for-asmpt}. Plugging this in for $M$, and noting that $\alpha \geq \max(T^K, (T')^{K'})$, we conclude that the length of such an $x'$ distinguishing $f, f'$ need only be at most 

\begin{align*}
    n &= \max(1 + O(M^3 \max(K,K') T^3 \alpha^{O(1)}), O(p_{c_*} \cdot  \alpha^{O(1)} ))\\
    &\leq \max(1 + O(M^3 \max(K,K') T^3 \alpha^{O(1)}), O(M^3 \cdot (MT^3) \cdot \alpha^{O(1)} \cdot  \alpha^{O(1)}) ))\\
    &\leq \boxed{O(\alpha^{O(1)})}
\end{align*}
$\blacksquare$% \footnote{Later: improving the bound on $M$ improves the result.}

%% file: sections/appendices/completeness.tex
\subsection{Lemmas for Completeness of Basis Schema}\label{appen:completeness}

%------------------------------------------------------------
% Completeness of Basis Schema
%------------------------------------------------------------

\paragraph{Goal.}

In the main proof, we will fix two arbitrary unequal $f, f' \in \crasp{}^{2}$ and prove they have a short distinguisher. The goal of this section is to prove that the set of realizable activations $\act (\{ s_i\}_{i \in [k]} )$ equals the union of the set of activations of a small number of \textit{basis} test-function schema. This culminates in Corollary \ref{cor:cleaner-basis-test-function}.

\paragraph{Basic Definitions.} Recall the definition of precision.

\precision*

Note we will say a halfspace defined via a linear inequality with integer coefficients has $p$-precision if each coefficient is at most $p$ in magnitude.

Suppose $f$ and $ f'$ have $K$ and $K'$ heads, respectively, and consider the set of $\max(K, K') \leq k \leq K + K'$ distinct slopes from the parameters of the first layer of $f$ and $f'$: $\{\frac{b^{(i)}}{a^{(i)}}\}_{i \in [K]} \cup \{\frac{(b^{(i)})'}{(a^{(i)})'}\}_{i \in [K']}$. Disregard for now which slope belongs to $f$ or to $f'$, and denote these $k$ slopes as $\{ s_i\}_{i \in [k]} \subset (0,1)$, sorted descending so that $s_1$ is largest and $s_k$ is smallest. We will refer to $\{ s_i\}_{i \in [k]} \subset (0,1)$ as a $(k, T)$-configuration.

\configuration*

These $k$ slopes $\{ s_i\}_{i \in [k]}$ specify $k$ homogeneous, $2$D lines, of the form $\yax = s_i \cdot \xax$. Denote these $k$ lines as $l_1, \ldots, l_k$ with line $l_i$ having slope $s_i$ for all $i \in [k]$.

Recall the definition of Discrete Test-Functions.

\disctestfn*

For a string $x \in \{ 0,1\}^*$, the discrete test-function induced by $x$ is the set of 2D points $\{ (j, \ps(x)_j) \}_{j \in [|x|]}$ where we will associate the $\yax$-axis for $\ps(x)_j$ and the $\xax$-axis for $j$.

Recall two central objects: continuous test-functions and $\act (\{ s_i\}_{i \in [k]} )$.

\cntstestfn*

\begin{definition} ($\act (\{ s_i\}_{i \in [k]} )$)\label{defn:A-activations}
    Given $(k,T)$-configuration $\{ s_i\}_{i \in [k]}$, define $\act (\{ s_i\}_{i \in [k]} )$ as the set of activations induced by continuous test-functions with respect to $\{ s_i\}_{i \in [k]}$.
    \begin{align*}
        \act (\{ s_i\}_{i \in [k]} ) &:=  \{ (B_1(\caly), \ldots, B_k(\caly) ) : \caly \text{ continuous test-function w.r.t. } \{s_i\}_{i \in [k]}\}
    \end{align*} 
\end{definition}

Regarding properties of continuous test-functions, first note that the scaling of $\caly$ can be set WLOG because of the homogeneity of the $k$ lines. Thus, we let their domain be $[0,1]$.

Second, for any continuous test-function $\caly$, we can let the end point of $\caly$ be on one of the lines $\{ l_i\}_{i \in [k]}$. Suppose the last line crossed by $\caly$ is $l_i$. Then we can adjust the segment of $\caly$ between its last crossing point at $l_i$ and its endpoint so that the endpoint is also on line $l_i$. We can make this tweak so that no other lines $\{ l_i\}_{i \in [k]}$ are crossed and so that $(B_1(\caly), \ldots, B_k(\caly))$ remains unchanged by this tweak. In short, the endpoint of any continuous test-function $\caly$ is, WLOG, $(1, s_i)$ where $l_i$ is the last line crossed by $\caly$.

\begin{lemma}\label{lem:end-pt-of-test-fn}
    For any configuration $\{ s_i\}_{i \in [k]}$, for any continuous test-function $\caly$ w.r.t. $\{ s_i\}_{i \in [k]}$, suppose the last line $\{ l_i\}_{i \in [k]}$ crossed by $\caly$ is $l_i$, for $i \in [k]$. Then, WLOG, we can let $\caly$'s end point be $(1, s_i)$ without changing the activations induced by $\caly$.
\end{lemma}

\textit{Proof.} Suppose that the last line $\caly$ crosses is $l_i$ at the point $(\xax, s_i \cdot \xax)$. Then, the portion of $\caly$ on the interval $[\xax, 1]$ is wedged between either the two lines $l_i$ and $l_{i + 1}$ or the two lines $l_i$ and $l_{i - 1}$, since $\caly$ will not cross any other line on the interval. The quantity of interest are the activations with respect to the $k$ lines induced by $\caly$:

\begin{align*}
    \forall i \in [k], B_i(\caly) := \int_0^1 \ind [\caly(j) > s_i \cdot j]dj
\end{align*}

Suppose that $\forall j \in (\xax, 1], s_ij > \caly(j) > s_{i + 1}j$. The $k$ quantities $\{ B_i(\caly)\}_{i \in [k]}$ will be unchanged if we adjust the values of $\caly(j)$ for $j \in (\xax, 1]$ so long as we retain that $\forall j \in (\xax, 1], s_ij > \caly(j) > s_{i + 1}j$ except on a set of measure $0$. With $\caly$ allowed to be any continuous function with slopes in $[0,1]$ and with $s_i \in (0,1)$, we can adjust $\caly(j)$ to stay between lines $l_i$ and $l_{i + 1}$ but closely follow the line  $l_i$ in the sense that $|\caly(j) - s_i j| > 0$ can be made arbitrarily small at all points $j \in (\xax, 1)$, and $\caly(1) = s_i$ (note, this end-point $(1, s_i)$ violates the condition that $s_ij > \caly(j) > s_{i + 1}j$ but only at a single point). This ensures that the modified test-function, call it $\caly'$, is such that $\forall i \in [k], B_i(\caly) = B_i(\caly')$.

Suppose that $\forall j \in (\xax, 1], s_{i - 1}j > \caly(j) > s_{i}j$. Then an analogous adjustment to $\caly(j)$ on the interval $j \in (\xax, 1]$ can be made so that the endpoint is $(1, s_i)$. $\blacksquare$

From now on, assume that each test-function will have starting-point at the origin $(0,0)$ and have end point on some line $l_i \in \{ l_1, \ldots, l_k\}$, at point $(1, s_i)$. This will make the following definitions about segments and schema cleaner.

\begin{figure}[ht]
\vskip 0.2in
\begin{center}
\centerline{\includegraphics[width=\columnwidth]{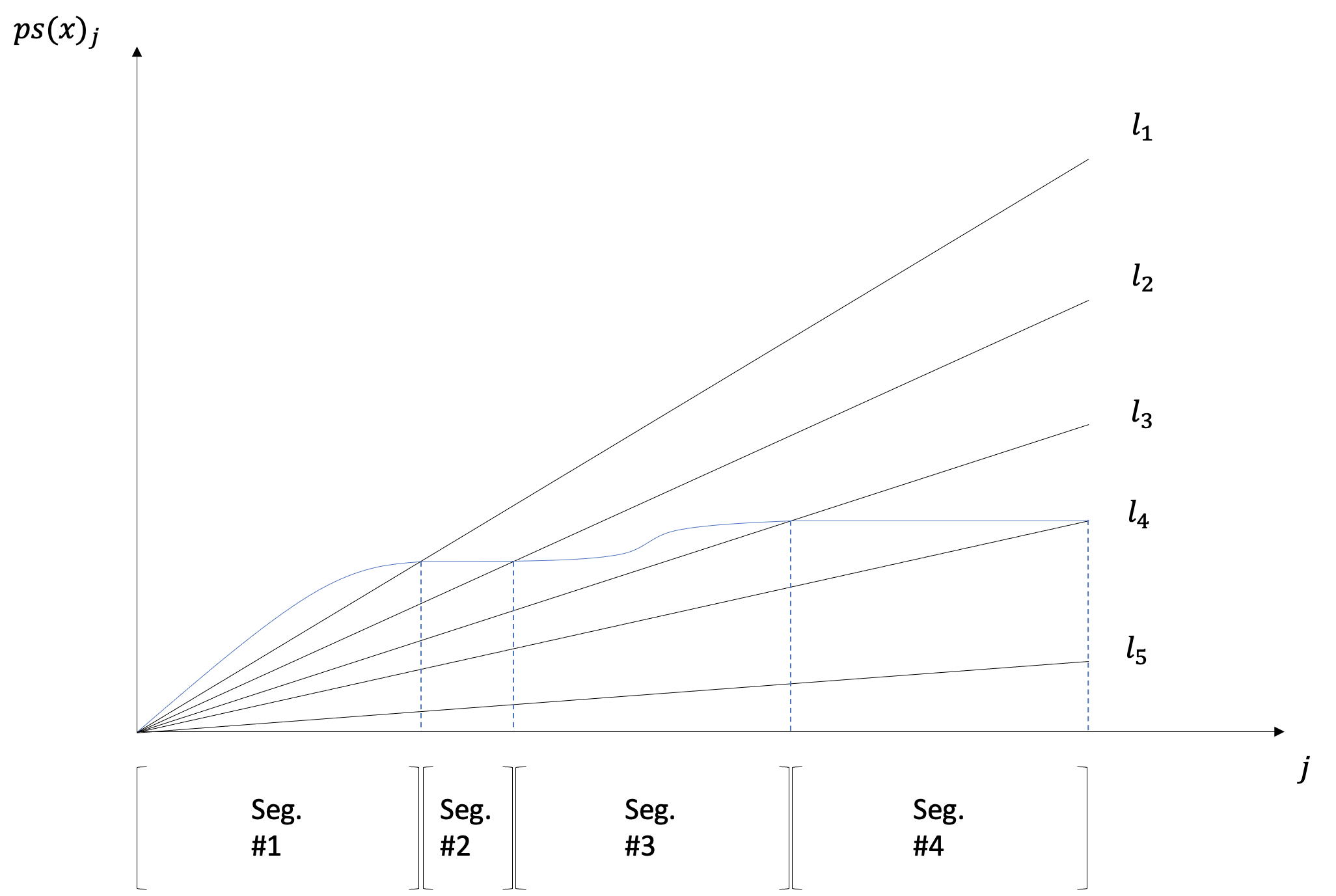}}
\caption{Depiction of a Test-Function consisting of 4 segments. $\yax$-axis shows the prefix sum of input string $x$. $\xax$-axis shows the length of the prefix of input string $x$. The five lines have slopes $\{ s_1, \ldots, s_5\}$. $\textup{Sector}_1$ is the portion of the quadrant which is above $l_1$. $\textup{Sector}_6$ is the portion of the quadrant below $l_5$. For $2 \leq i \leq 5$, $\textup{Sector}_i$ is the portion of the quadrant below $l_{i - 1}$ and above $l_i$. }
\label{fig:test-fn}
\end{center}
\vskip -0.2in
\end{figure}

Define the span of a continuous test-function as the set of lines in $\{ l_i\}_{i \in [k]}$ which the continuous test-function intersects at some point.

\begin{definition}
    The span of a test-function is the set of lines $\{ l_1, \ldots, l_k\}$ that the test-function crosses at least once. That is, $\caly$ crosses $l_i$ if there exists $\xax$ where $\caly(\xax) = s_i \cdot \xax$. Note that the span must be a contiguous subset of $[k]$.
\end{definition}

 We'll also give names to the regions of the positive quadrant of the 2D plane between consecutive lines in $\{ l_i\}_{i \in [k]}$.

\begin{definition}\label{defn:sectors} (Sectors)
    Given a $(k,T)$-configuration $\{ s_i\}_{i \in [k]}$, a sector is a region of the 2D space between two consecutive lines. Define $\textup{Sector}_1$ as the sector strictly above line $l_1$, and  $\textup{Sector}_{k + 1}$ as the sector below line $l_k$. For $i \in \{ 2,\ldots,k\}$ define $\textup{Sector}_i$ as the sector below line $l_{i - 1}$ and strictly above line $l_i$.
\end{definition}

\begin{figure}[ht]
  \centering
  \begin{minipage}[b]{0.475\textwidth}
    \includegraphics[width=\linewidth]{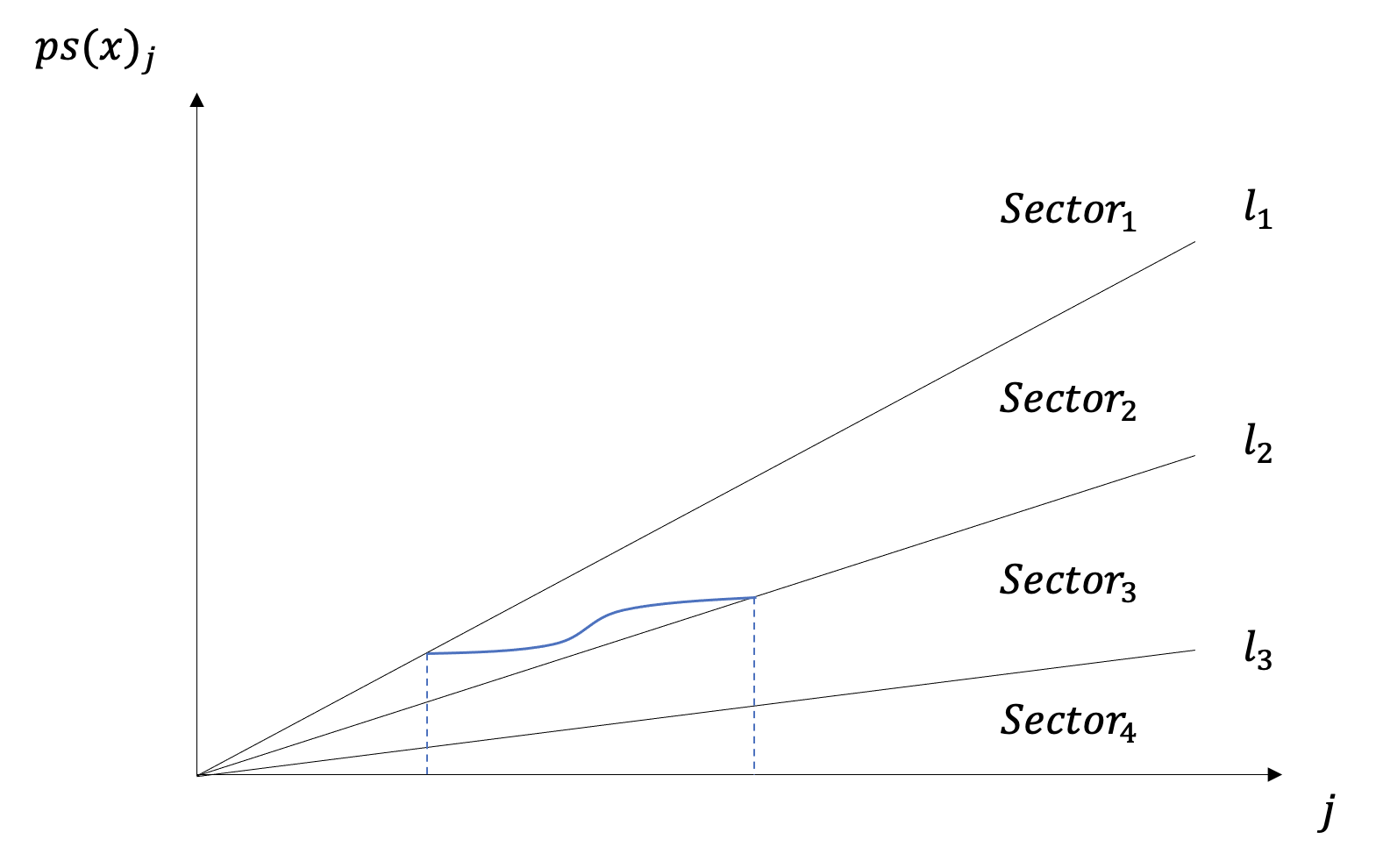}
    \caption*{Start-point on $l_1$, End-point on $l_2$.}\label{fig:a}
  \end{minipage}\hfill
  \begin{minipage}[b]{0.475\textwidth}
    \includegraphics[width=\linewidth]{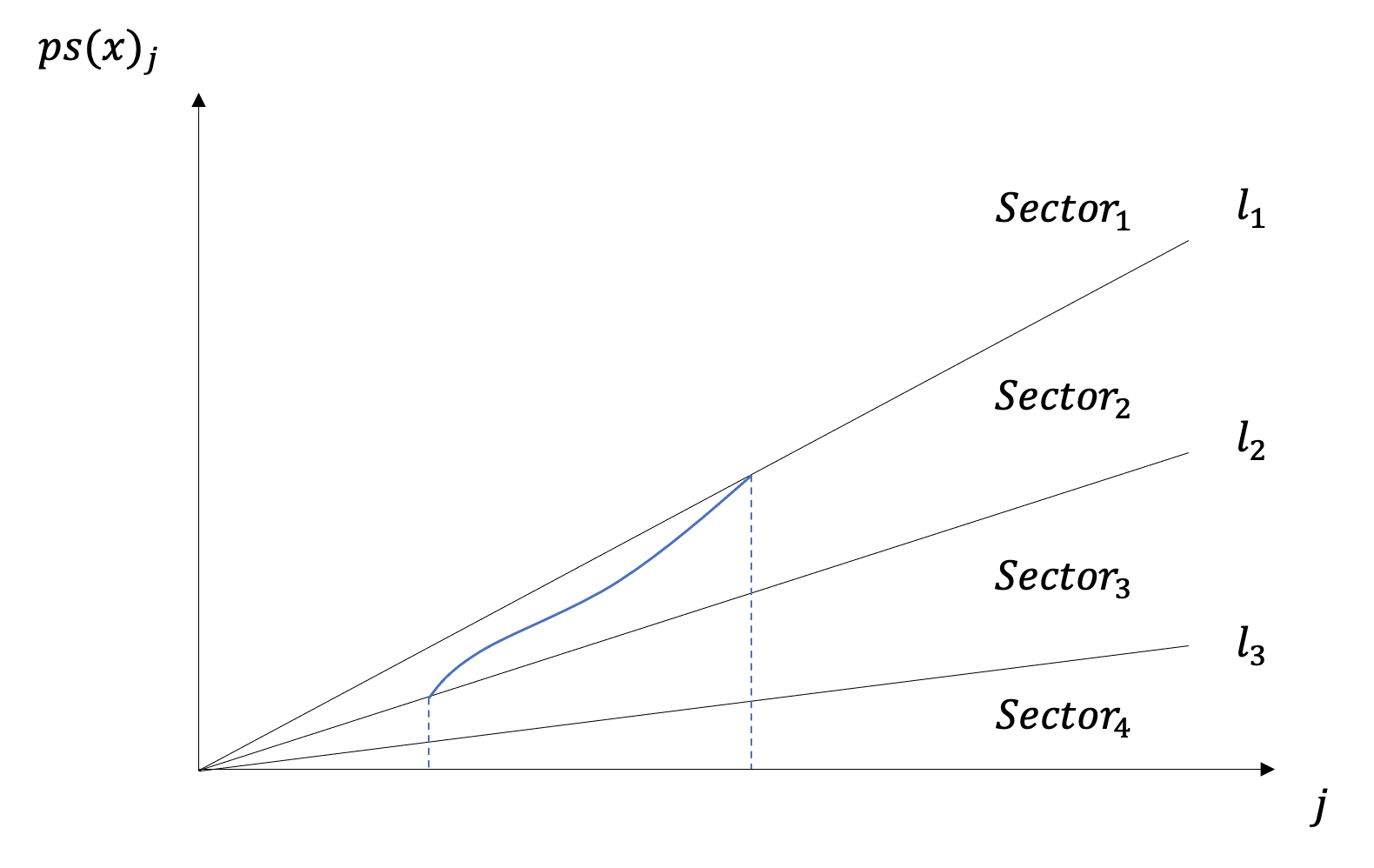}
    \caption*{Start-point on $l_2$, End-point on $l_1$.}\label{fig:b}
  \end{minipage}
  \vskip\baselineskip
  \begin{minipage}[b]{0.475\textwidth}
    \includegraphics[width=\linewidth]{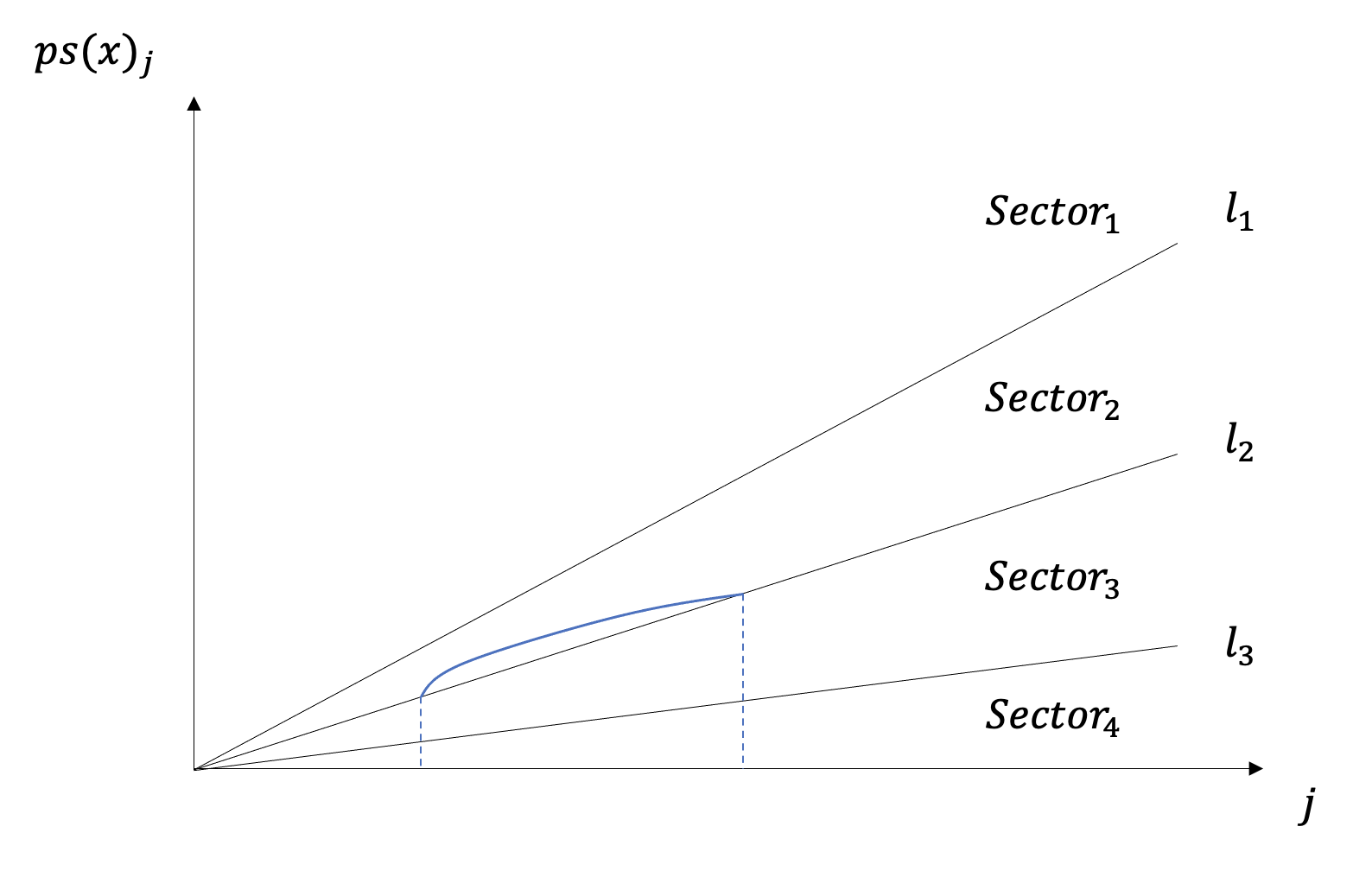}
    \caption*{Start-point and End-point on $l_2$, and in $\textup{Sector}_2$.}\label{fig:c}
  \end{minipage}\hfill
  \begin{minipage}[b]{0.475\textwidth}
    \includegraphics[width=\linewidth]{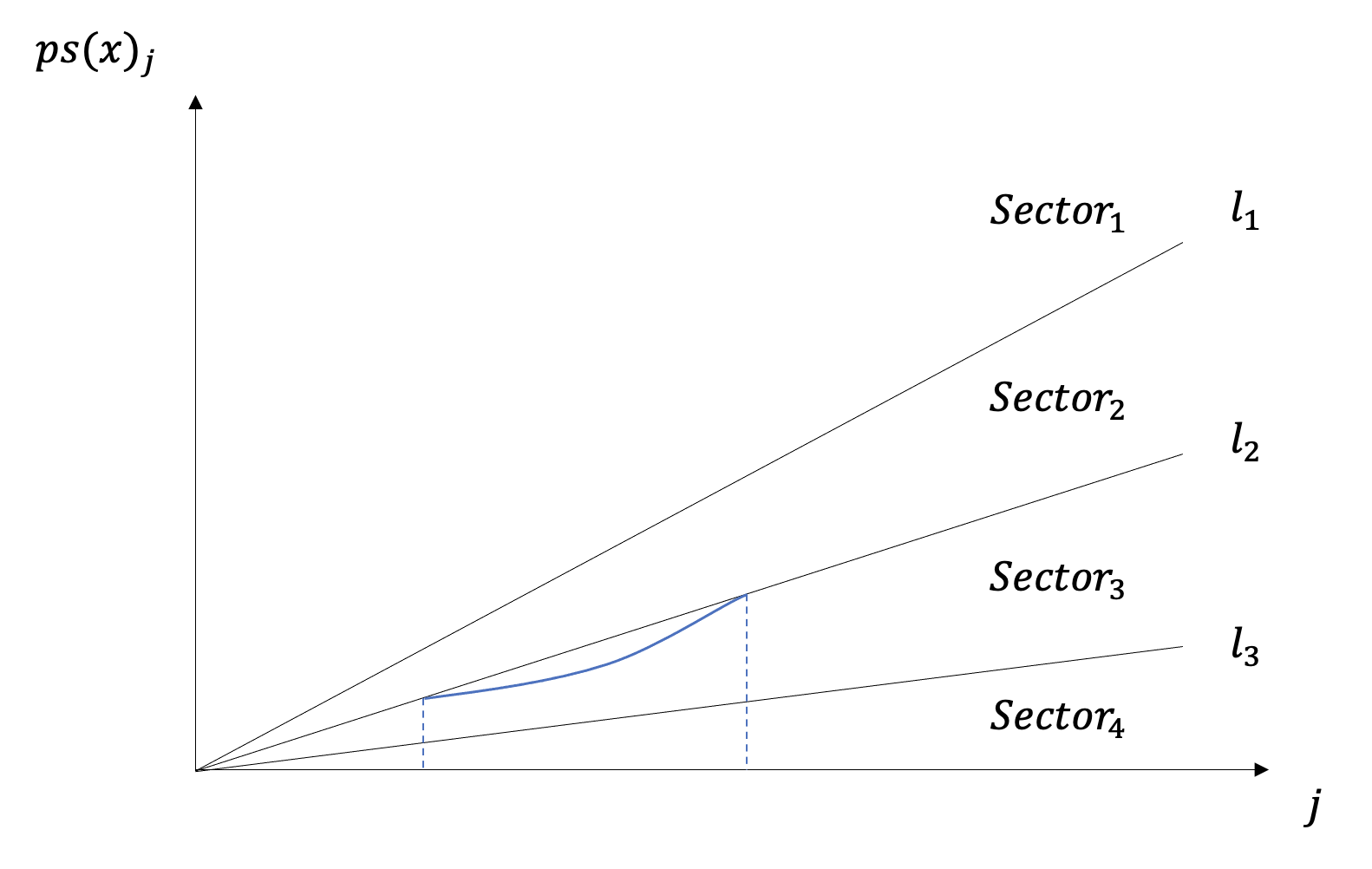}
    \caption*{Start-point and End-point on $l_2$, and in $\textup{Sector}_3$.}\label{fig:d}
  \end{minipage}
  \caption{Four types of Segments, based on which lines their start-point and end-point lie on, and the sector they are in.}\label{fig:four-segments}
\end{figure}

Sectors are depicted in Figure \ref{fig:four-segments}. We'll now define segments and schema. 

\begin{definition} (Segments and Schema)\label{defn:schema} Given any $(k,T)$-configuration $\{ s_i\}_{i \in [k]} \subset (0,1)$, define the following.
\begin{enumerate}
    \item A \textit{segment} is a restricted test-function $S : [a,b] \to [0,1]$ where $[a,b] \subset [0,1]$ which maps a continuous subset $[a,b]$ to $[0,1]$. $S$ is $1$-Lipschitz and monotone non-decreasing. The segment's start-point $(a, S(a)) $ and the end-point $(b,S(b))$ each lie on one of the $k$ lines, in the sense that there exists some $i,j \in [k] $ where $S(a) = s_i \cdot a$ and $S(b) = s_j \cdot b$, where $i = j$ or $|i - j| = 1$. No other points $(x, S(x))$, $x \in (a,b)$ can lie on a line $l_1,\ldots, l_k$. 
    \item A \textit{schema}  $Y$ is a blueprint for a continuous test-function, specifying a sequence of lines $\{ l_i\}_{i \in [k]}$ that any test-function of the schema must cross. It consists of an integer $0 < M < \infty$ and two tuples $\{ \textup{idx}(i)\}_{i \in [M]} \subset [k]^M, \{ \textup{sec}_i\}_{i \in [M]} \subset [k + 1]^M$, where $|\textup{idx}(i) - \textup{idx}(i + 1)| \leq 1$ for all $i \in [M - 1]$. If $|\textup{idx}(i) - \textup{idx}(i + 1)| = 1$, then $\textup{sec}_{i + 1}$ is unique and must be $\max(\textup{idx}(i), \textup{idx}(i + 1))$. If $\textup{idx}(i) = \textup{idx}(i + 1)$, then $\textup{sec}_{i + 1}$ can be either $\textup{idx}(i + 1)$ or $\textup{idx}(i + 1) + 1$. $\textup{sec}_1$ can be either $\textup{idx}(1)$ or $\textup{idx}(1) + 1$.

    Any continuous test-function of schema $Y = (\{ \textup{idx}(i)\}_{i \in [M]}, \{ \textup{sec}_i\}_{i \in [M]})$ consists of exactly $M$ segments $S_1, S_2, \ldots, S_M$ whose domains are a partition of $[0,1]$. For each $i \in [M]$, the $i$th segment $S_i$'s end-point lies on $l_{\textup{idx}(i)}$. For $i > 1$, $S_i$'s start-point lies on line $l_{\textup{idx}(i - 1)}$, and $S_1$'s start-point is the origin, $(0,0)$. In addition, the $i$th segment must be contained in $\textup{Sector}_{\textup{sec}_i}$.

    Notationally, we denote $n_i \in [0,1]$ as the length of the $i$th segment $S_i$, but note that different test-functions of the same schema may have different segment lengths $\{ n_i\}_{i \in [M]}$, subject to some constraints we detail below. For a schema $Y$, denote 
    \begin{align*}
        A(Y) := \{ (B_1(\caly), \ldots, B_k(\caly) ) : \caly \text{ continuous test-function of schema $Y$ w.r.t. } \{s_i\}_{i \in [k]}\}
    \end{align*}

\end{enumerate}
\end{definition}

Figure \ref{fig:test-fn} shows a test-function of a schema with 4 segments. Figure \ref{fig:four-segments} shows four generic types of segments.

We will think of a schema as a list of $M$ segments, for some $M > 0$. We will denote the length of the $i$th segment as $n_i$. Let $\textup{idx}(\cdot)$ be the mapping from $[M]$ to $[k]$ that gives the index of the line that segment $i$'s endpoint is on. 

For continuous test-functions, the first segment of length $n_1$ has start-point at the origin, $(0,0)$. The last segment of length $n_M$ has end-point which lies on the line $l_{\textup{idx}(M)}$. The $i$th segment has start-point $(\sum_{j = 1}^{i - 1} n_j, s_{\textup{idx}(i - 1) } \cdot \sum_{j = 1}^{i - 1} n_j) $ on line $l_{\textup{idx}(i - 1)}$ (as long as $i \geq 2$) and end-point $(\sum_{j = 1}^{i} n_j, s_{\textup{idx}(i) } \cdot \sum_{j = 1}^{i} n_j)$ on line $l_{\textup{idx}(i)}$ and has length $n_i$. There is freedom in choosing $\{ n_i\}_{i \in [M]}$, so long as they satisfy the following constraints.

\begin{lemma} \label{lem:schema-constraints}
    For any segment in any schema, the constraints given in \Cref{defn:schema} exactly characterize the range of values allowed for that segment. For all $i \in [M]$, these constraints are:
    \begin{itemize}
    \item (Segment Starts and Ends on Same Line) If $\textup{idx}(i - 1) = \textup{idx}(i)$ or $i = 1$, the only constraint on the segment's length is $n_i \geq 0$.
    \item (Segment Crosses Down) If $\textup{idx}(i - 1) = \textup{idx}(i) - 1$ and $i \geq 2$, then $n_i \geq (\frac{s_{\textup{idx}(i - 1)}}{s_{\textup{idx}(i)}} - 1) \sum_{j = 1}^{i - 1} n_j$
    \item (Segment Crosses Up) If $\textup{idx}(i - 1) = \textup{idx}(i) + 1$ and $i \geq 2$, then $n_i \geq (\frac{1 - s_{\textup{idx}(i - 1)}}{1 - s_{\textup{idx}(i)}} - 1) \sum_{j = 1}^{i - 1} n_j$
    \end{itemize}
\end{lemma}

\textit{Proof.} \textbf{Case: Segment Starts and Ends on Same Line.} For segment $i$ with length $n_i$, if $\textup{idx}(i - 1) = \textup{idx}(i)$, then $n_i$ can be any nonnegative number. This is because for each $n_1, \ldots, n_{i - 1}$ and for all $n_i \geq 0$, there exists a segment of length $n_i$ that crosses line $\textup{idx}(i)$ at the start-point $(\sum_{j \leq i - 1} n_j, s_{\textup{idx}(i)} \sum_{j \leq i - 1} n_j)$ and end-point $(\sum_{j \leq i } n_j, s_{\textup{idx}(i)} \sum_{j \leq i } n_j)$  and nowhere in between. An example of such a segment is one which stays arbitrarily close to the line $l_{\textup{idx}(i)}$ but doesn't cross it until the end-point $(\sum_{j \leq i } n_j, s_{\textup{idx}(i)} \sum_{j \leq i } n_j)$, which is possible since each line has slope in the range $(0,1)$ while the test-function can have slopes at each point be in the range $[0,1]$.

\textbf{Case: Segment Crosses Down.} If $\textup{idx}(i - 1) = \textup{idx}(i) - 1$, then for any $n_1, \ldots, n_{i - 1}$, the minimum value of $n_i$ in terms of $n_1, \ldots, n_{i - 1}$ is achieved if the segment has slope of $0$ at all points, the ``minimal segment", which will let the segment cross lines $l_{\textup{idx}(i - 1)}$ and $l_{\textup{idx}(i)}$ most efficiently. Such a minimal segment will cross lines $l_{\textup{idx}(i - 1)}$ and $l_{\textup{idx}(i)}$ at start-point $(\sum_{j \leq i - 1} n_j, s_{\textup{idx}(i - 1)} \sum_{j \leq i - 1} n_j)$ and end-point $(\sum_{j \leq i } n_j, s_{\textup{idx}(i)} \sum_{j \leq i } n_j)$, respectively. The value of $n_i$ required for this minimum traversal can be calculated by:

 \begin{align*}
     s_{\textup{idx}(i)}\sum_{j = 1}^{i } n_j &= s_{\textup{idx}(i - 1)}\sum_{j = 1}^{i - 1} n_j\\
     \implies n_i &= (\frac{s_{\textup{idx}(i - 1)}}{s_{\textup{idx}(i)}} - 1) \sum_{j = 1}^{i - 1} n_j
 \end{align*}

 The first equation follows from the fact that  the zero-slope trajectory will enforce that the $\yax$-coordinate of the start-point of the segment $s_{\textup{idx}(i - 1)}\sum_{j = 1}^{i - 1} n_j$ equals that of the end-point $s_{\textup{idx}(i)}\sum_{j = 1}^{i } n_j$. Having $n_i$ be any smaller will not suffice to make the crossing, under the trajectory of the minimal segment (where the slope is $0$ everywhere) and certainly under any other trajectories.
 
 Now, having $n_i > (\frac{s_{\textup{idx}(i - 1)}}{s_{\textup{idx}(i)}} - 1) \sum_{j = 1}^{i - 1} n_j$ is always possible, since the segment can always follow the trajectory of the minimal segment to cross to line $\textup{idx}(i)$, and then use the extra slack, $n_i - (\frac{s_{\textup{idx}(i - 1)}}{s_{\textup{idx}(i)}} - 1) \sum_{j = 1}^{i - 1} n_j > 0$ to closely follow line $\textup{idx}(i)$ until it crosses it at the designated end-point, $(\sum_{j \leq i } n_j, s_{\textup{idx}(i)} \sum_{j \leq i } n_j)$. The latter phase reduces to the case where $\textup{idx}(i - 1) = \textup{idx}(i)$. Thus, the allowed values for $n_i$ are $n_i \geq (\frac{s_{\textup{idx}(i - 1)}}{s_{\textup{idx}(i)}} - 1) \sum_{j = 1}^{i - 1} n_j$.

\textbf{Case: Segment Crosses Up.} The setup of lines and $[0,1]$-slope test-functions has a ``reflection" symmetry about the line $\yax = \frac{1}{2}\xax$. Thus, the argument for the constraint on values of $n_i$ for the $\textup{idx}(i - 1) = \textup{idx}(i) + 1$ case reduces to that of the case $\textup{idx}(i - 1) = \textup{idx}(i) - 1$, except with ``reflected" slopes $1 - s_{\textup{idx}(i - 1)}$ and  $1 - s_{\textup{idx}(i)}$. Plugging in these reflected slopes into the constraint for the Cross Down case yields $n_i \geq (\frac{1 - s_{\textup{idx}(i - 1)}}{1 - s_{\textup{idx}(i)}} - 1) \sum_{j = 1}^{i - 1} n_j$.

Another way to see how to derive the constraint by considering the minimal trajectory as one where the segment has slope $1$ everywhere; and arguing that $n_i$ can be anything larger than the length required by the minimal trajectory.

 \begin{align*}
     (1 - s_{\textup{idx}(i)})n_i &= (s_{\textup{idx}(i)} - s_{\textup{idx}(i - 1)})\sum_{j = 1}^{i - 1} n_j\\
     \implies n_i &= (\frac{1 - s_{\textup{idx}(i - 1)}}{1 - s_{\textup{idx}(i)}} - 1) \sum_{j = 1}^{i - 1} n_j
 \end{align*}

The first equation above is derived as ``Object 1 of relative speed of  $(1 - s_{\textup{idx}(i)})$ to Object 2 takes $n_i$ time to close the initial gap of $(s_{\textup{idx}(i)} - s_{\textup{idx}(i - 1)})\sum_{j = 1}^{i - 1} n_j$." $\blacksquare$

\paragraph{Partial Test-functions and Monotone Curves.} 

We'll now define partial test-functions, which is a slight generalization of continuous test-functions where the start-point does not need to be $(0,0)$, which will be later in the later proofs. 

\begin{definition} (Partial Continuous Test-Function)
    Given a $(k,T)$-configuration $\{ s_i\}_{i \in [k]}$, a continuous test-function $\caly$  is partial if it is allowed to have a start-point at  $ (n_1, s_i, n_1), n_1 \in [0,1), i \in [k]$, instead of $(0,0)$. A partial test-function is undefined on $[0, n_1)$. The induced activations $(B_1(\caly), \ldots, B_k(\caly))$ of partial continuous test-function $\caly$ given $k$ slopes $\{ s_i\}_{i \in [k]} \subset (0,1)$ are defined as: 
    \begin{align*}
        \forall i \in [k], \quad B_i(\caly) := \int_{n_1}^1 \ind [\caly(j) > s_i \cdot j] dj
    \end{align*}
    Schemas of partial test-functions can be thought of as a schema of a continuous test-function. We will still denote the lengths of the segments of the schema as $\{ n_i\}_{i \in [M]}$ for some $M > 1$, except that the test-function is undefined on the first segment's domain $[0, n_1)$. 
\end{definition}

Note that by a re-scaling argument, due to the homogeneity of the $k $ lines $\{ l_i\}_{i \in [k]}$, this definition also captures continuous test-functions where the start-point does not need to be at $(0,0)$ and the end-point need not have $\xax$-coordinate of 1.

We'll define type  (I, $k$) and (II, $k$) partial test-functions, which are particular partial test-functions whose start-point is on a line whose index is either the smallest or largest element in the span of the test-function.

\begin{definition} (Partial test-functions of type (I, $k$), (II, $k$))
    Given a $(k,T)$-configuration $\{ s_i\}_{i \in [k]}$,
    \begin{itemize}
        \item Define a type (I, $k$) partial test-function as a test-function $[0,1] \to [0,1]$ that is undefined on $[0, n_1)$ for some $0 \leq n_1 < 1$. It may span any consecutive subset of lines $\{ a, \ldots, b\} \subset \{ 1, 2, \ldots, k - 1, k\}, a \leq b$. With this span, we require that its start-point is at $(n_1, s_a n_1)$, that it is 1-Lipschitz and monotone non-decreasing, and that it can only intersect the $k$ lines at finitely many points. 
        
        \item Define a type (II, $k$) partial test-function similarly as a test-function $[0,1] \to [0,1]$ undefined on $[0, n_1)$. It may span any consecutive subset of lines $\{ a, \ldots, b\} \subset \{ 1, 2, \ldots, k - 1, k\}, a \leq b$. With this span, we require that its start-point is $(n_1, s_b n_1)$, that it is 1-Lipschitz and monotone non-decreasing, and that it can only intersect the $k$ lines at finitely many points. 
    \end{itemize}
    
\end{definition}

Note that for both type (I, $k$) partial test-functions, if the start-point is on $l_a$ with $a > 1$, then its span cannot contain $l_1$ and it cannot intersect $l_1$ at any point. An analogous observation holds for type (II, $k$) partial test-functions.

% if the $k$ lines $\{ s_i\}_{i \in [k]}$ are a consecutive subset of a larger set of $k' > k$ lines, then each type (I, $k$), (II, $k$) test-function may only span lines in $\{ s_i\}_{i \in [k]}$ and cannot intersect the other $k' - k$ lines.

We'll describe one primitive (partial test-function) schema that will be important for our basis test-function schema later: monotone curves.

\begin{definition}\label{defn:monotone-curve} (Monotone Curve)
    Given a $(k,T)$-configuration $\{ s_i\}_{i \in [k]}$, we define two schemas: Monnotone Up Curve and Monotone Down Curve. Both schema have $k + 2$ segments.
    \begin{itemize}
        \item (Monotone Down) Begins at line $1$ at $(n_1, s_1 \cdot n_1)$, goes above and recrosses line 1 at $(n_1 + n_2, s_1 \cdot [n_1 + n_2])$, crosses each intermediate line  $\{ 2,\ldots, k - 1\}$ once, then crosses line $k$ twice at $(\sum_{i = 1}^{k + 1} n_i, s_k \cdot \sum_{i = 1}^{k + 1} n_i)$ and $(\sum_{i = 1}^{k + 2} n_i, s_k \cdot \sum_{i = 1}^{k + 2} n_i)$. 
        \item (Monotone Up) Begins at line $k$ at $(n_1, s_k \cdot n_1)$, goes below and recrosses line $k$ at $(n_1 + n_2, s_k \cdot [n_1 + n_2])$, crosses each intermediate line once, then crosses line $1$ twice at $(\sum_{i = 1}^{k + 1} n_i, s_1 \cdot \sum_{i = 1}^{k + 1} n_i)$ and $(\sum_{i = 1}^{k + 2} n_i, s_1 \cdot \sum_{i = 1}^{k + 2} n_i)$. 
    \end{itemize}
    Note $n_1 \in [0, 1)$ denotes the length of the first, ``empty" segment on which the test-function is undefined; there are $k + 1$ nonempty segments. WLOG by homogeneity of the $k$ lines that $\sum_{i = 2}^{k + 2}n_i = 1$. 
\end{definition}

\begin{remark}
    Test-functions of the monotone curve schema are monotone in the sense that they don't re-cross lines which they previously crossed, except for line $1$ and line $k$.
\end{remark}

\begin{remark} Type (I, $k$), (II, $k$) test-functions are strictly more general than test-function since we can just set $n_1 = 0$ to recover the usual test-function definition. Type (I, $k$) test-functions start on the top-most line while (II, $k$) test-functions start on the bottom-most line. Monotone Down Curves are Type (I, $k$) while Monotone Up Curves are Type (II, $k$)\end{remark}

\paragraph{Rearrangement Lemma for Monotone Curve.}

\begin{definition} (Equivalence) \label{defn:equiv-test-fn}
    Define two partial test-functions as \textit{equivalent} if both test-functions have the same start-point, end-point, length, and induced activations over the $k$ lines.
\end{definition}

We will now introduce a central Lemma in proving that a certain set of schema is complete. This Lemma is about rearranging a test-function into an equivalent test-function with a simpler schema.

\begin{lemma}\label{lem:monotoneline}
    Given a $(k,T)$-configuration $\{ s_i\}_{i \in [k]}$, suppose a type (I, $k$) test-function (resp. a type (II, $k$)) has its start point on line 1 and end point on line $k$ (resp. start point on line $k$ and end point on line $1$) and spans lines $\{ 1,\ldots, k\}$. Then there is an equivalent, monotone curve of the same length, start point, end point, and that induces the same $(B_1,\ldots, B_k)$.
\end{lemma}

\textit{Proof.} We will only prove the conversion of a type (I, $k$) test-function to a monotone (down) curve. The conversion of a type (II, $k$) test-function to a monotone (up) curve is an analogous argument with effective slopes $s_i' = 1 - s_i, \forall i \in [k]$.

Suppose the type (I, $k$) test-function, $\caly$, has $M$ segments of length $n_1,\ldots,n_M$, where $n_1$ is the $\xax$-coordinate of the starting point of $\caly$. Consider a monotone line $\caly^{(M)}$ that, like $\caly,$ starts at $(n_1, s_1 \cdot n_1)$ and ends at $(\sum_{i = 1}^M n_i, s_k \cdot \sum_{i = 1}^M n_i)$. $\caly^{(M)}$ is comprised of $k + 2$ segments of length $n_1', n_2',\ldots, n_k', n_{k + 1}', n_{k + 2}'$ (where $n_1'$ is an empty segment, denoting the coordinate of the start point), calculated from $n_1,\ldots,n_M$ as follows.

\begin{align*}
    n_1' &= n_1\\
    \forall i \in [k + 1],\quad  n_{i + 1}' &= \sum_{j \in \{ 2,3,\ldots,M\} : n_j \in \textup{Sector}_i} n_j
\end{align*}

First, we must show that this ``rearrangement" forms a valid test-function that meets the constraints that would be enforced on $n_1', n_2',\ldots, n_k', n_{k + 1}', n_{k + 2}'$ described in Lemma \ref{lem:schema-constraints}. 

First, the only constraint on $n_2'$ and $n_{k + 2}'$ is that they must be nonnegative. 

Second, $\forall i \in [3, k + 1], $ the $i$th segment traverses sector $i - 1$ and must cross from $l_{i - 2}$ to $l_{i - 1}$. By Lemma \ref{lem:schema-constraints}, the following constraint is satisfied iff this crossing is possible:

\begin{align*}
    n_i' \geq (\frac{s_{i - 2}}{s_{i - 1}} - 1)\sum_{j = 1}^{i - 1}n_j'
\end{align*}

We would like to show that $(n_i')_{i \in [k + 2]}$ meets these constraints. First, $\forall i \in [3, k + 1],$ define $j_i := \max\{ j \in \{ 2,3,\ldots,M\} : n_j \text{ crosses } \textup{Sector}_{i - 1}\}$ as the index of the last segment where $\caly$ crosses $\textup{Sector}_{i - 1}$, starting on line $l_{i - 2}$ and ending on line $l_{i - 1}$. Then because the curve $\caly$ is continuous and its endpoint is on line $k$, then $j_i$ is monotone in $i$: $j_3 < j_4 < \ldots < j_{k + 1}$. Then,

\begin{align*}
    n_{i}' &= \sum_{j \in \{ 2,3,\ldots,M\} : n_j \in \textup{Sector}_{i - 1}} n_j\\
    &\geq n_{j_i} \text{ \quad just the last segment crossing $\textup{Sector}_{i - 1}$ suffices}\\
    &\geq (\frac{s_{i - 2}}{s_{i - 1}} - 1)\sum_{j = 1}^{(j_i) - 1}n_j\\
    &\geq (\frac{s_{i - 2}}{s_{i - 1}} - 1)\sum_{j = 1}^{j_{i - 1}}n_j \quad \text{ monotonicity of $j_i$}\\
    &\geq (\frac{s_{i - 2}}{s_{i - 1}} - 1)\sum_{j \leq j_{i - 1} \text{ s.t. } n_j \in \bigcup_{m = 1}^{i - 2}\textup{Sector}_m }n_j\\
    &= (\frac{s_{i - 2}}{s_{i - 1}} - 1)\sum_{j = 1}^{i - 1}n_j'
\end{align*}

Thus, the monotone curve $\caly^{(M)}$ is a valid monotone test-function. $\caly^{(M)}$ has the same length as $\caly$ since it just rearranged the segments while preserving their length. $\caly^{(M)}$ starts at line 1 and ends at line $k$, and it starts at $(n_1, s_1 n_1)$ just like $\caly$, so it must also end at the same point as $\caly^{(M)}$ on line $k$. $\caly^{(M)}$ induces the same activations $(B_1,\ldots, B_k)$ as $\caly$ since the rearranged segment lengths stay in their original sectors in $\caly^{(M)}$.

The proof for type (II, $k$) test-function is analogous.  $\blacksquare$

\paragraph{Basis Schema.}

The following is the main result of this section. It says that the following finite set of \textit{basis} schema is complete, in the sense that any continuous test-function is equivalent to some test-function whose schema is one of the basis schema.

Each basis schema described below is indexed by integer $m \in [k]$ and a set of $m$ tuples $\{(y_1^i, y_2^{i})\}_{i \in [m]} \subset \{ 1, \ldots, k\}^m$. These $m$ tuples parameterize $m - 1$ monotone-curves that, when concatenated, yield the schema. For $i \in [m - 1]$, the $i$th tuple $(y_1^i, y_2^i) \in [k]^2$  indicates that the $i$th monotone curve in the schema will have start-point on line $y_1^i$ and end-point on line $y_2^i$. The concatenation of all $m - 1$ monotone curves yield the basis schema.

\begin{corollary}(Completeness of Basis Schema)\label{cor:cleaner-basis-test-function}
    Given a $(k,T)$-configuration $\{ s_i\}_{i \in [k]}$, for any $1 \leq m \leq k$, say that the list of tuples $\{(y_1^i, y_2^{i})\}_{i \in [m]} \subset \{ 1, \ldots, k\}^m$ is valid if they satisfy the following. 
    \begin{align*}
        y_1^m &= y_2^m\\
        \forall i \in [m - 1], y_1^i &\neq y_2^i\\
        \forall i \in [m - 2], y_1^i &< y_2^i \implies y_2^i = y_1^{i + 1} > y_2^{i + 1} > y_1^i\\
        \forall i \in [m - 2], y_1^i &> y_2^i \implies y_2^i = y_1^{i + 1} < y_2^{i + 1} < y_1^i\\
        (y_1^1, y_2^1) &= (1,k) \text{ or } (y_1^1, y_2^1) = (k,1)
    \end{align*}
    For any $m \in [k]$ and valid $\{(y_1^i, y_2^{i})\}_{i \in [m]}$, define the basis schema, $Y_{\{(y_1^i, y_2^{i})\}_{i \in [m]}}$ as the concatenation of $m - 1$ monotone curves, where for $i \in [m - 1]$, the $i$th monotone curve has start-point on line $y_1^i \in [k]$ and end-point on line $y_2^i \in [k]$. The 1st monotone curve has start-point at the origin, $(0,0)$. 
    
    Then, the set of basis schemas over all $m$ and valid  $\{(y_1^i, y_2^{i})\}_{i \in [m]}$ satisfying the above is complete in the following sense.
    \begin{align*}
        \act (\{ s_i\}_{i \in [k]}) = \bigcup_{m \in [k], \text{valid } \{(y_1^i, y_2^{i})\}_{i \in [m]}} A(Y_{\{(y_1^i, y_2^{i})\}_{i \in [m]}})
    \end{align*}
    Where $A(Y_{\{(y_1^i, y_2^{i})\}_{i \in [m]}}) \subset [0,1]^k$ denotes the set of $(B_1,\ldots, B_k)$ induced by any test-function of schema $Y_{\{(y_1^i, y_2^{i})\}_{i \in [m]}}$. Finally, note that the number of basis schema with respect to $\{ s_i\}_{i \in [k]}$ is $N_k = 2^{k - 1}$.
\end{corollary}

\begin{figure}[ht]
\vskip 0.2in
\begin{center}
\centerline{\includegraphics[width=\columnwidth]{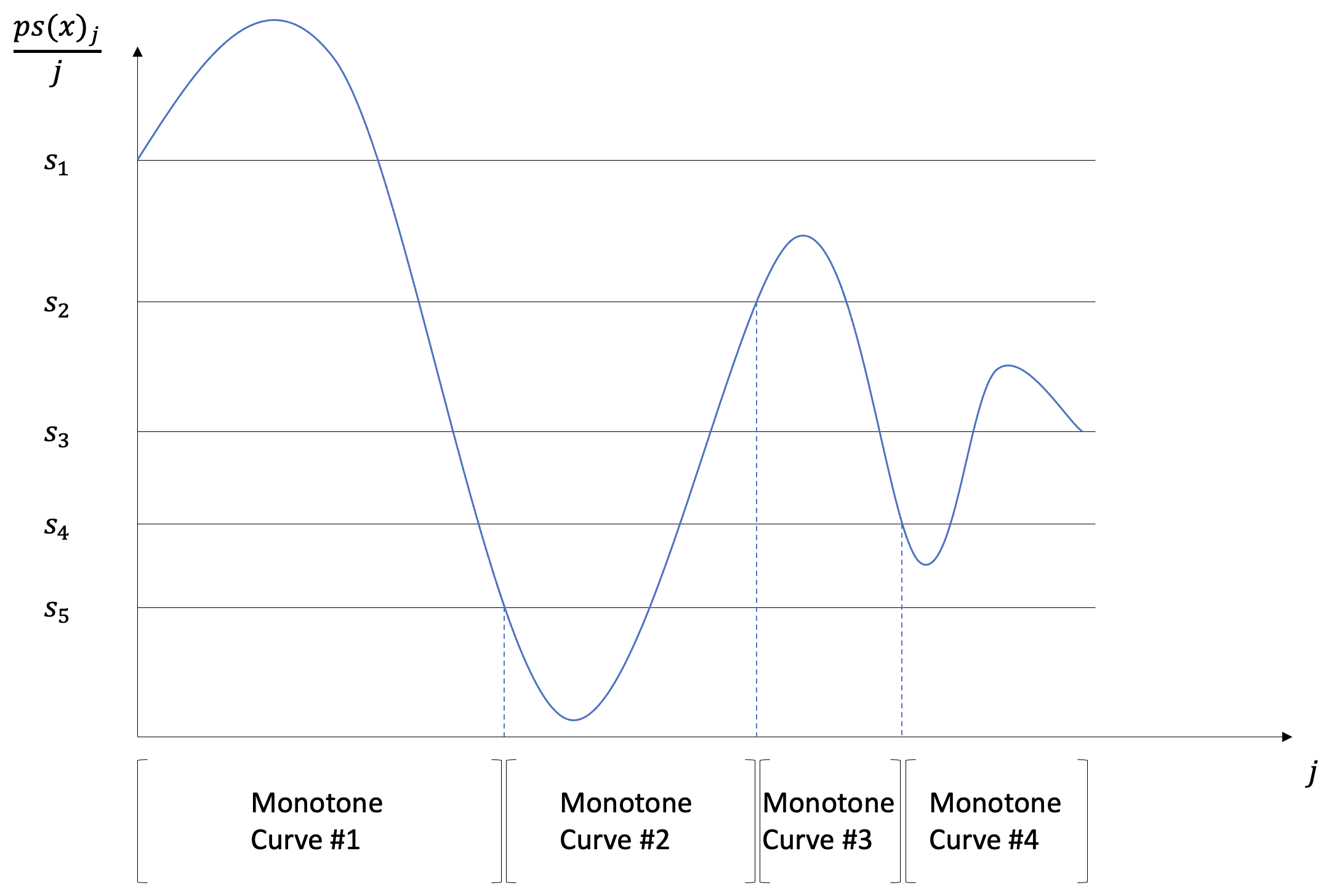}}
\caption{Depiction of a Basis Schema consisting of 4 monotone curves. $\yax$-axis shows the normalized prefix sum of input string $x$. $\xax$-axis shows the length of the prefix of input string $x$. The five horizontal lines correspond to five lines with slopes $\{ s_1, \ldots, s_5\}$. Curve corresponds to Basis Schema with $m = 5$, $(y_1^1, y_2^1) = (1,5), (y_1^2, y_2^2) = (5,2), (y_1^3, y_2^3) = (2,4), (y_1^4, y_2^4) = (4,3)$. Finally, note that each monotone curve consists of multiple segments. The first one has 5 segments, the second one has 4 segments, the third one has 3 segments, and the fourth one has 3 segments.}
\label{fig:basis-schema}
\end{center}
\vskip -0.2in
\end{figure}

\begin{figure}[ht]
\vskip 0.2in
\begin{center}
\centerline{\includegraphics[width=0.8\columnwidth]{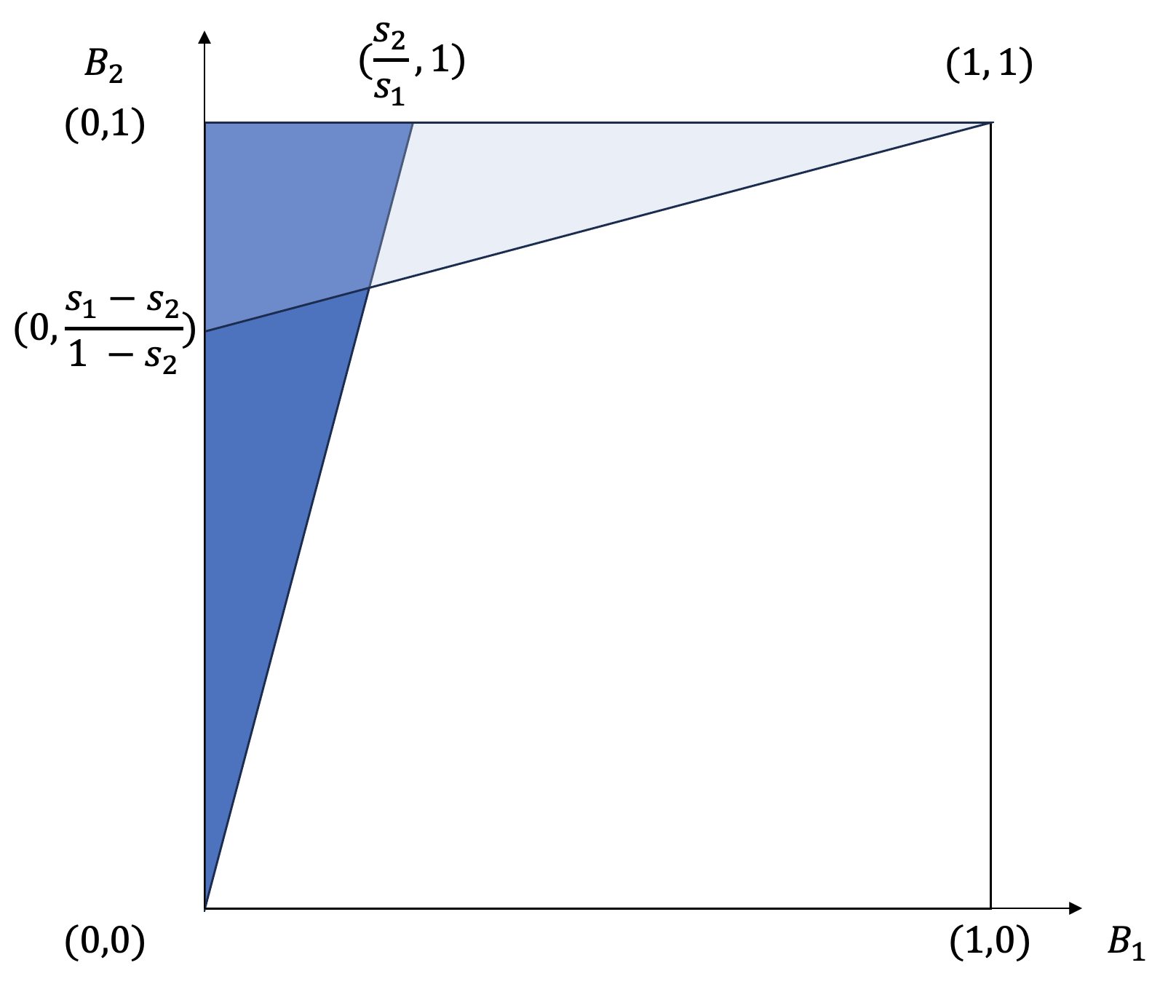}}
\caption{Depiction of $\act (\{ s_i\}_{i \in [k]})$ for $k = 2$ with two slopes $s_1 > s_2$. $B_1$ is on the horizontal axis while $B_2$ is on the vertical axis. When $k = 2$, there are only two basis schema: a single monotone (up) curve and a single monotone (down) curve. The dark blue triangle with vertices $\{ (0,0), (0,1), (\frac{s_2}{s_1}, 1)\}$ is the set of activations induced by test-functions of the monotone (down) curve basis schema. The light blue triangle with vertices $\{ (1,1), (0,1), (0, \frac{s_1 - s_2}{1 - s_2})\}$ is the set of activations induced by test-functions of the monotone (up) curve. The completeness result says that the union of these two triangles equals $\act (\{ s_1, s_2\})$.}
\label{fig:A}
\end{center}
\vskip -0.2in
\end{figure}

Refer to Figure \ref{fig:basis-schema} to get a sense for what the basis schema look like. Note that the $\yax$ axis of the figure shows the \textit{normalized} prefix sum. Also, Figure \ref{fig:A} is a depiction of the Completeness result for the case where $k = 2$.

\textit{Proof of Corollary \ref{cor:cleaner-basis-test-function}.} Recall that $\act (\{ s_i\}_{i \in [k]} )$ and  $A(Y_{\{(y_1^i, y_2^{i})\}_{i \in [m]}})$ for schema $Y_{\{(y_1^i, y_2^{i})\}_{i \in [m]}}$ are defined follows.

\begin{align*}
    A(Y_{\{(y_1^i, y_2^{i})\}_{i \in [m]}}) &:= \{ (B_1(\caly), \ldots, B_k(\caly) ) : \caly \text{ continuous test-function of schema $Y_{\{(y_1^i, y_2^{i})\}_{i \in [m]}}$ w.r.t. } \{s_i\}_{i \in [k]}\}\\
    \act (\{ s_i\}_{i \in [k]} ) &:=  \{ (B_1(\caly), \ldots, B_k(\caly) ) : \caly \text{ continuous test-function w.r.t. } \{s_i\}_{i \in [k]}\}
\end{align*}

First, because every test-function of a basis schema is a test-function,

\begin{align*}
    \act (\{ s_i\}_{i \in [k]}) \supset \bigcup_{m \in [k], \text{valid } \{(y_1^i, y_2^{i})\}_{i \in [m]}} A(Y_{\{(y_1^i, y_2^{i})\}_{i \in [m]}})
\end{align*}

It suffices to show that the converse inclusion holds.

\begin{align*}
    \act (\{ s_i\}_{i \in [k]}) \subset \bigcup_{m \in [k], \text{valid } \{(y_1^i, y_2^{i})\}_{i \in [m]}} A(Y_{\{(y_1^i, y_2^{i})\}_{i \in [m]}})
\end{align*}

To do this, we will prove that any arbitrary continuous test-function can be converted into an equivalent test-function (in the sense of \Cref{defn:equiv-test-fn}), but which follows one of the basis schema. This conversion process is done via Algorithm \ref{alg:decompose-into-mc}, which partitions the input test-function into pieces, and then uses Lemma \ref{lem:monotoneline} to convert each piece into a monotone curve, yielding a test-function of a basis schema.

% suppose its span is given by $\{ a, a+ 1, \ldots, b - 1, b\} \subset [k]$, which is a continuous subset of $[k]$. %  We will describe how to deal with the case when it spans fewer than $k$ lines, later.

% suppose its span is given by $\{ a, a+ 1, \ldots, b - 1, b\} \subset [k]$, which is a continuous subset of $[k]$. Recall that the span of a test-function is the set of lines it crosses at some point (where the lines are indexed from $1, \ldots, k$). The span is a continuous subset of $[k]$ since we assume that $l_i$ has the highest slope $s_i$, where $\{ s_i\}_{i \in [k]}$ is sorted descending.

More precisely, given an arbitrary test-function $\caly$, the main idea is to partition $\caly$ into pieces, such that each piece is either a partial Type (I, $\beta$) or Type (II, $\beta$) test-function for some  $1 \leq \beta \leq k$. Each piece can then be rearranged into a monotone curve via an application of Lemma \ref{cor:cleaner-basis-test-function}. We use Algorithm \ref{alg:decompose-into-mc} to attain $\{ (y_1^\alpha, y_2^\alpha)\}_{\alpha \in [m - 1]}$ which characterize the appropriate basis-schema for which there exists an equivalent test-function to $\caly$.

Recall that the span of a partial test-function is the set of lines $\{ l_1, \ldots, l_k\}$ that it crosses at some point in its domain (where the lines are indexed from $1, \ldots, k$). The span of a test-function is a continuous subset of $[k]$ since we assume that $l_i$ has the highest slope $s_i$ and $\{ s_i\}_{i \in [k]}$ is sorted descending. Regarding notation in Algorithm \ref{alg:decompose-into-mc}, $\textup{Span}_i $ is a tuple of two integers in $[k]^2$, which represent the smallest and largest index of lines in the span of some partial continuous test-function. $\min(\textup{Span}_i)$ is the smaller element in the tuple, and $\max(\textup{Span}_i)$ is the larger element. $\{ s_i\}_{i \in [k]}$ are the $k$ slopes we fixed at the beginning, and $\caly(j) = s_i j$ indicates that test-function $\caly$ intersects the line $\yax = s_i \xax$ at $(j, s_i j)$. Also, when we say that $\caly$ intersects the line $l_i$ on $[t, 1]$ to mean that there is some $j \in [t,1]$ where $\caly(j) = s_i j$, for $t \leq 1$. 

\begin{algorithm}[tb]
\caption{Decompose-Into-Monotone-Curves}
\label{alg:decompose-into-mc}
\begin{algorithmic}[1]
    \STATE \textbf{Initialize:} $\textup{Span}_1 \gets \textup{Span}(\caly),\; t_0 \gets 0,\; b_0 \gets 0$
    \STATE $\textup{YPairs} \gets [\,]$  \quad \textit{\% List of all $(y_1^{\alpha},\, y_2^{\alpha})$ pairs}
    \STATE $\textup{MaxTB} \gets [0\,]$   \quad \textit{\% List of all $\max(t_{\alpha},\, b_{\alpha})$ values}

    \FOR{$\alpha \in \{ 1, 2, \ldots, k - 1\}$}
        \STATE $t_{\alpha} \gets 
               \displaystyle \max\Bigl\{ j \geq \max(t_{\alpha - 1},\, b_{\alpha - 1}) :
               \mathcal{Y}(j) = s_{\min(\textup{Span}_{\alpha})}\,j \Bigr\}$
        \STATE $b_{\alpha} \gets 
               \displaystyle \max\Bigl\{ j \geq \max(t_{\alpha - 1},\, b_{\alpha - 1}) :
               \mathcal{Y}(j) = s_{\max(\textup{Span}_{\alpha})}\,j \Bigr\}$

        \IF{$t_{\alpha} > b_{\alpha}$}
            \STATE $y_1^{\alpha} \gets \max(\textup{Span}_{\alpha}) \quad$ and $\quad
                   y_2^{\alpha} \gets \min(\textup{Span}_{\alpha})$
            \STATE $\textup{Span}_{\alpha+1} \gets \Bigl(
                   \min(\textup{Span}_{\alpha}),\;
                   \max\{\, i < \max(\textup{Span}_{\alpha}) :
                         i \in [k],\,
                         \mathcal{Y} \text{ intersects line } l_i
                         \text{ on } [\max(t_{\alpha},\, b_{\alpha}),\, 1]\}\Bigr)$
        \ELSE
            \STATE $y_1^{\alpha} \gets \min(\textup{Span}_{\alpha}) \quad$ and $\quad
                   y_2^{\alpha} \gets \max(\textup{Span}_{\alpha})$
            \STATE $\textup{Span}_{\alpha+1} \gets \Bigl(
                   \min\{\, i > \min(\textup{Span}_{\alpha}) :
                          i \in [k],\,
                          \mathcal{Y} \text{ intersects line } l_i
                          \text{ on } [\max(t_{\alpha},\, b_{\alpha}),\, 1]\},\,
                   \max(\textup{Span}_{\alpha})\Bigr)$
        \ENDIF

        \STATE $\textup{YPairs} \gets \textup{YPairs} \cup \{(y_1^{\alpha},\, y_2^{\alpha})\}$
        \STATE $\textup{MaxTB} \gets \textup{MaxTB} \cup \{\max(t_{\alpha},\, b_{\alpha})\}$

        \IF{$\max(\textup{Span}_{\alpha}) = \min(\textup{Span}_{\alpha})$}
            \STATE \textbf{break}
        \ENDIF
    \ENDFOR

    \STATE \textbf{return} $(\textup{YPairs},\, \textup{MaxTB})$
\end{algorithmic}
\end{algorithm}

Towards proving completeness of the basis schema, we will prove the following three claims. At the end, we will use these claims to argue that $\act (\{ s_i\}_{i \in [k]}) \subset \bigcup_{m \in [k], \text{valid } \{(y_1^i, y_2^{i})\}_{i \in [m]}} A(Y_{\{(y_1^i, y_2^{i})\}_{i \in [m]}})$.

\begin{enumerate}
    \item Algorithm \ref{alg:decompose-into-mc} terminates.
    \item Algorithm \ref{alg:decompose-into-mc} returns valid $\textup{YPairs} := \{ (y_1^i, y_2^i)_{i \in [m - 1]}\}$ where $m := |\textup{YPairs}| + 1$ with $m \leq k$, and where the valid predicate is defined in the statement of Corollary \ref{cor:cleaner-basis-test-function}.
    \item $\textup{MaxTB} := \{ T_i\}_{0 \leq i \leq m - 1} \subset [0,1]$ is a set of real numbers where for each $i \in \{ 0, 1,, \ldots, m - 2\}$, $\caly$ restricted to the interval $[T_i, T_{i + 1}]$ can be rearranged into an equivalent monotone curve by Lemma \ref{lem:monotoneline}.
\end{enumerate}

% First, by the Definition of Continuous Test-Functions, $\caly$ will intersect the $k$ lines at finitely many points, so each of the quantities we take min or max over is well defined.

\textbf{Proving Termination.} At each iteration $\alpha$, Algorithm \ref{alg:decompose-into-mc} maintains variables $\textup{Span}_{\alpha}$, $t_{\alpha - 1},$ and $ b_{\alpha - 1}$, which we claim satisfy the property that $\textup{Span}_{\alpha}$ holds the smallest and largest index of lines $\{ 1, \ldots , k\}$ which $\caly$ spans when $\caly$ is restricted to interval $[\max(t_{\alpha - 1}, b_{\alpha - 1}), 1]$. 

First, suppose the span of $\caly$ is a single line (recall that the span of $\caly$ is always at least one line, since by Lemma \ref{lem:end-pt-of-test-fn}, we let the endpoint of any test-function be on some line). Then, Algorithm \ref{alg:decompose-into-mc} initializes $\textup{Span}_1$ to be the span of $\caly$, when restricted to interval $[0, 1]$. Algorithm \ref{alg:decompose-into-mc} computes $y_1^1, y_2^1$, then terminates. Our claim holds.

For $\caly$ that span 2 or more lines, we will prove via induction that for each iteration $\alpha$, $\textup{Span}_{\alpha}$, $t_{\alpha - 1},$ and $ b_{\alpha - 1}$ satisfy our claim. As a base case, $\textup{Span}_1 = \textup{Span}(\caly)$ satisfies our claim. As the inductive step, suppose $\textup{Span}_{\alpha}$ holds the smallest and largest index of lines $\{ 1, \ldots , k\}$ which $\caly$ spans, when restricted to interval $[\max(t_{\alpha - 1}, b_{\alpha - 1}), 1]$. Then, in iteration $\alpha$, the variables $t_{\alpha}$ and $b_{\alpha}$ exist (recall that a continuous test-function only intersects the lines at finitely many points, so the max is well defined). We can never have that $t_{\alpha} = b_{\alpha}$ as long as $\min(\textup{Span}_{\alpha - 1}) < \max(\textup{Span}_{\alpha - 1})$, which must be true at the start of iteration $\alpha$, otherwise the algorithm would have terminated in iteration $\alpha - 1$. Thus, either $t_\alpha > b_\alpha$ or $t_\alpha < b_\alpha$. If $t_{\alpha} > b_{\alpha}$, then the last point where $\caly$ intersects line $\min(\textup{Span}_{\alpha})$ is later than the last point where $\caly$ intersects line $\max(\textup{Span}_{\alpha})$. On the interval $[\max(t_{\alpha}, b_{\alpha}), 1]$, $\caly$ never intersects line $\max(\textup{Span}_{\alpha})$ again. Thus, the smallest index of any line which $\caly$ intersects on the interval $[\max(t_{\alpha}, b_{\alpha}), 1]$ is $\min(\textup{Span}_{\alpha})$. The largest index of any line which $\caly$ intersects on the interval $[\max(t_{\alpha}, b_{\alpha}), 1]$ must be less than $\max(\textup{Span}_{\alpha})$ as $t_{\alpha} > b_{\alpha}$. Thus, the largest index line which $\caly$ intersects on the interval $[\max(t_{\alpha}, b_{\alpha}), 1]$ is given by the expression, $\max\{ i < \max(\textup{Span}_\alpha) : i \in [k], \caly \text{ intersects line $l_i$ on } [\max(t_\alpha, b_\alpha), 1]\}$. This shows that line 9 correctly sets $\textup{Span}_{\alpha + 1}$, completing the induction. A similar correctness argument can be made for the case where $t_{\alpha} < b_{\alpha}$. This proves that for all $\alpha$, $\textup{Span}_{\alpha}$ correctly holds the smallest and largest index of lines $\{ 1, \ldots , k\}$ which $\caly$ spans when it is restricted to interval $[\max(t_{\alpha - 1}, b_{\alpha - 1}), 1]$.

% Further, they are at least $\max(t_{\alpha - 1}, b_{\alpha - 1})$.
%Also, on the interval $[\max(t_{\alpha}, b_{\alpha}), 1]$, $\caly$ will only intersect line $\min(\textup{Span}_{\alpha})$ at the point $(t_{\alpha}, s_{\min(\textup{Span}_{\alpha})} \cdot t_{\alpha}) \in \mathbb{R}^2$. 

For all $\alpha$, we have that $\max(\textup{Span}_\alpha) - \min(\textup{Span}_\alpha) \leq \max(\textup{Span}_{\alpha - 1}) - \min(\textup{Span}_{\alpha  - 1}) - 1$. Thus, the Algorithm terminates after at most $|\textup{Span}(\caly)| - 1 \leq k - 1$ iterations since the largest that $|\textup{Span}(\caly)|$ can be is $k$, when $\caly$ spans all $k$ lines (by $|\textup{Span}(\caly)|$, we mean the number of lines which $\caly$ spans). In particular, if we let $m := |\textup{YPairs}| + 1$ returned by Algorithm \ref{alg:decompose-into-mc}, then $m \leq|\textup{Span}(\caly)| \leq  k$.

\textbf{Proving Validity of $\textup{YPairs}$.} Towards the second claim, first note that if $t_{\alpha} > b_{\alpha}$ in iteration $\alpha$ of Algorithm \ref{alg:decompose-into-mc} and the minimum and maximum of $\textup{Span}_{\alpha + 1}$ are not equal, then in the next iteration $\alpha + 1$, it will be the case that $b_{\alpha + 1} > t_{\alpha + 1}$. This is because the last point where $\caly$ crosses line $\min(\textup{Span}_{\alpha})$ has $\xax$-coordinate $t_{\alpha}$, by line 5 in iteration $\alpha$. However, by lines 8 and 9 in iteration $\alpha$, it must also be true that in the next iteration $\alpha  + 1$,
$t_{\alpha + 1}$, defined in line 5 of iteration $\alpha + 1$,  equals $\max(t_{\alpha}, b_\alpha) = t_{\alpha}$. Since we cannot have $t_{\alpha + 1} = b_{\alpha + 1}$ (argued previously) and $t_{\alpha + 1}$, $ b_{\alpha + 1}$ are both at least $\max(t_{\alpha}, b_\alpha)$, we have that $b_{\alpha + 1} > t_{\alpha + 1} = t_\alpha$. By lines 11 and 12 in iteration $\alpha + 1$, Algorithm \ref{alg:decompose-into-mc} will set $y_1^{\alpha} > y_2^{\alpha + 1} > y_1^{\alpha + 1} = y_2^{\alpha}$. Analogously, if $t_{\alpha} < b_{\alpha}$ and $\textup{Span}_{\alpha + 1}$ has a distinct minimum and maximum element, then $b_{\alpha + 1} < t_{\alpha + 1}$, and $y_1^{\alpha} < y_2^{\alpha + 1} < y_1^{\alpha + 1} = y_2^{\alpha}$. This proves that the algorithm returns $\textup{YPairs}$ which is a set of at most $m - 1 \leq |\textup{Span}(\caly)| - 1 \leq k - 1$ pairs $\{ (y_1^{\alpha}, y_2^{\alpha})\}_{\alpha \in [m - 1]}$ where each pair has a distinct minimum and maximum and satisfies the properties in Corollary \ref{cor:cleaner-basis-test-function}, except potentially for the property that $(y_1^1, y_2^1) = (1,k)$ or $(y_1^1, y_2^1) = (k, 1)$.  

The last issue about ensuring $\textup{YPairs}$ satisfies the property that $(y_1^1, y_2^1) = (1,k)$ or $(y_1^1, y_2^1) = (k, 1)$ is simple to deal with. The issue arises when $\textup{Span}(\caly) \subsetneq [k]$, so that $\caly$ does not span all $k$ lines. However, we can note that any test-function which does not span all $k$ lines can be thought of as part of a schema which does span all $k$ lines, except that the segments of the schema which cross lines in $[k] - \textup{Span}(\caly)$ are the first segments of the schema, and their lengths are \textit{set equal to 0} for the particular case of $\caly$ (which is a valid setting of the lengths of the first segments, per Lemma \ref{lem:schema-constraints}). Thus, though $\caly$ does not span $k$ lines, it can be thought of as belonging to a schema which does. We will discuss this more at the end. % Finally, the output $\textup{YPairs}$ of Algorithm \ref{alg:decompose-into-mc} on any $\caly$ which spans all $k$ lines satisfies all the requirements stated in Corollary \ref{cor:cleaner-basis-test-function}.

% so we can effectively  prepend two pairs $(y_1^0, y_2^0)$ and $(y_1^{-1}, y_2^{-1})$ to $\textup{YPairs}$, the output of Algorithm \ref{alg:decompose-into-mc} on $\caly$. These two pairs can be set so that $(y_1^{-1}, y_2^{-1})$ is either $(1,k)$ or $(k,1)$ and they satisfy the properties stated in Corollary \ref{cor:cleaner-basis-test-function}. Renumbering 

\textbf{Converting Each Part into a Monotone Curve, to Convert $\caly$ into a Continuous Test-function of a Basis Schema.} Third, with $\textup{MaxTB} := \{ T_i\}_{0 \leq i \leq m - 1}$, for each $\alpha \in [m - 1]$, the pair $(y_1^{\alpha}, y_2^{\alpha})$ generated by an iteration of Algorithm \ref{alg:decompose-into-mc} corresponds to the interval $[T_{\alpha - 1}, T_{\alpha}]$, in the sense that

\begin{align*}
    (T_{\alpha - 1}, \caly(T_{\alpha - 1})) &= (T_{\alpha - 1}, T_{\alpha - 1} \cdot s_{y_1^{\alpha}})\\
    (T_{\alpha }, \caly(T_{\alpha })) &= (T_{\alpha } , T_{\alpha } \cdot s_{y_2^{\alpha}})
\end{align*}

% at $\xax$ coordinate $T_{\alpha - 1}$, $\caly$ intersects line $y_1^{\alpha}$, while at $\xax$ coordinate $T_{\alpha }$, $\caly$ is on line $y_2^{m'}$. 

We have already argued that the span of the partial test-function given by restricting $\caly$ to $[T_{\alpha}, 1]$ is given exactly by $\textup{Span}_{\alpha}$, where either $y_1^\alpha = \min(\textup{Span}_{\alpha})$ and $ y_2^\alpha = \max(\textup{Span}_{\alpha})$; or $y_1^\alpha = \max(\textup{Span}_{\alpha})$ and $ y_2^\alpha = \min(\textup{Span}_{\alpha})$. Thus, the restriction of $\caly$ on interval $[T_{\alpha - 1}, T_{\alpha}]$ is a Type (I, $\max(\textup{Span}_{\alpha}) - \min(\textup{Span}_{\alpha}) + 1$) or Type (II, $\max(\textup{Span}_{\alpha}) - \min(\textup{Span}_{\alpha}) + 1$) partial test-function, for which Lemma \ref{lem:monotoneline} applies.

By Lemma \ref{lem:monotoneline}, the restriction of $\caly$ on interval $[T_{\alpha - 1}, T_{\alpha}]$ is equivalent to a monotone curve of span $\textup{Span}_{\alpha}$, which has start-point $(T_{\alpha - 1}, s_{y_1^{\alpha}} \cdot T_{\alpha - 1})$ and end-point $(T_{\alpha}, s_{y_2^{\alpha}} \cdot T_{\alpha})$. The monotone curve will have the same start-point, end-point, length, and induced activations as $\caly$ restricted\footnote{Here we are applying Lemma \ref{lem:monotoneline} on restrictions of test-functions where the end-point is not at $1$, but we can do this due to the homogeneity of the setup} to $[T_{\alpha - 1}, T_{\alpha}]$. Thus, we can construct such a monotone curve for each $\alpha \in [m - 1]$ and concatenate these monotone curves together. The resulting test-function is equivalent to $\caly$, as each individual monotone curve is equivalent to the corresponding restriction of $\caly$. 

We claim the resulting test-function is of one of the basis schema described in Corollary \ref{cor:cleaner-basis-test-function}. This essentially follows from the validity of $\textup{YPairs}$ outputted by Algorithm \ref{alg:decompose-into-mc}, which we justified in the previous section. To reiterate a key point, if $\textup{Span}(\caly) = [k]$, then $\textup{YPairs}$ satisfies the conditions of Corollary \ref{cor:cleaner-basis-test-function} exactly. On the other hand, if $\textup{Span}(\caly) \subsetneq [k]$ is a strict (continuous) subset of $[k]$, then still we can view the concatenation of monotone curves above as a test-function of a basis schema in Corollary \ref{cor:cleaner-basis-test-function}. To see this, we observe that for any such $\caly$, there is a basis schema $Y$ such that if we set the lengths of the segments of the monotone curves in schema $Y$ whose span is a strict superset of $\textup{Span}(\caly)$ to $0$, then the remaining monotone curves are given by $\textup{YPairs}$ outputted by Algorithm \ref{alg:decompose-into-mc} on input $\caly$. This basis schema $Y$ would be given as follows. Given $\textup{YPairs}$ outputted by Algorithm \ref{alg:decompose-into-mc} on input $\caly$, we pre-pend either one or two pairs to $\textup{YPairs}$. If $\min(\textup{Span}(\caly)) = 1$ or $\max(\textup{Span}(\caly)) = k$, we pre-pend a single pair to $\textup{YPairs}$: $(k, 1)$ in the first case and $(1,k)$ in the second case. If $\min(\textup{Span}(\caly)) > 1$ and $\max(\textup{Span}(\caly)) < k$, we pre-pend two pairs $(1, k), (k, \min(\textup{Span}(\caly)))$ to $\textup{YPairs}$. The resulting list of pairs with one or two pairs pre-pended, which we call $L$, will have at most $k - 1$ pairs total. $L$ will satisfy all requirements in Corollary \ref{cor:cleaner-basis-test-function}. As we argued that we can simply set the lengths of the monotone curves corresponding to the pre-pended pairs in $L$ to 0, the final test-function we attained by concatenating all the monotone curves together in the previous paragraph is of the basis schema corresponding to the list of pairs $L$.

Thus, the concatenation of monotone curves described above, based on the output by Algorithm \ref{alg:decompose-into-mc}, is of some basis schema specified in Corollary \ref{cor:cleaner-basis-test-function}. This proves that any continuous test-function $\caly$ is equivalent (in particular, induces the same activations) to some test-function of a basis schema specified in Corollary \ref{cor:cleaner-basis-test-function}. This implies that $\act (\{ s_i\}_{i \in [k]}) \subset \bigcup_{m \in [k], \text{valid } \{(y_1^i, y_2^{i})\}_{i \in [m]}} A(Y_{\{(y_1^i, y_2^{i})\}_{i \in [m]}})$, as desired.
\\
\\
Finally, regarding the number of basis schemas, we can first partition up the schema based on which of the $k$ lines the end-point is on. Let $f(k,i)$ be the number of basis schema w.r.t. a configuration of $k$ slopes whose end-point is on line $i \in [k]$. We claim that $f(k,i) = {k - 1 \choose i - 1}$, from which it follows that the total number of basis schema is $\sum_{i \in [k]} {k - 1 \choose i - 1} = 2^{k - 1}$.

For $i \in [k]$, to see that $f(k,i) = {k - 1 \choose i - 1}$, let $I_i := [k + 1] - \{ i, i + 1\}$. Consider all permutations $\sigma(I_i)$ which contain $\{1, 2, \ldots, i - 1\}$  and $\{ k + 1, \ldots, i + 2\}$ as a subsequence. We claim there is a one-to-one correspondence between any such $\sigma(I_i)$ and a basis schema $Y_{\{ (y_1^i, y_2^i)\}_{i \in [m]}} $. Given $Y_{\{ (y_1^i, y_2^i)\}_{i \in [m]}} $, let $g(\{ (y_1^i, y_2^i)\}_{i \in [m]})$ be a permutation of $I_i$ where the ordering of element $j \in I_i$ in the permutation is the relative ordering of sector $j$ based on the last time where schema $Y_{\{ (y_1^i, y_2^i)\}_{i \in [m]}} $ passed through sector $j$. The mapping is injective because for any two $\{ (y_1^i, y_2^i)\}_{i \in [m]} \neq \{ ((y_1^i)', (y_2^i)')\}_{i \in [m']}$, the first pair such that $(y_1^i, y_2^i) \neq ((y_1^i)', (y_2^i)')$ will be such that $y_2^i \neq (y_2^i)'$, and one of these schema will have visited $\textup{Sector}_{y_2^i}$ or $\textup{Sector}_{(y_2^i)'}$ for the last time while the other will return to it later. The surjectivity of the mapping can be checked easily.

% For each $j \in I_i$, the position of $j$ in $\sigma(I_i)$ represents the ``last time" which the schema passed through $\textup{Sector}_j$. Two distinct basis schema with endpoint on line $i$ cannot 

There are $i - 1$ elements in $\{1, 2, \ldots, i - 1\}$ and $k - 1$ elements in $I_i$, so there are ${k - 1 \choose i - 1}$ such permutations. $\blacksquare$

\begin{corollary}\label{cor:size-of-M}
    Given a $(k,T)$-configuration $\{ s_i\}_{i \in [k]}$, each basis test-function schema specified in Corollary \ref{cor:cleaner-basis-test-function} crosses the $k$ lines of slopes $\{ s_i\}_{i \in [k]}$ a total of at most $k^2$ times.
\end{corollary}

\textit{Proof.} By Corollary \ref{cor:cleaner-basis-test-function}, each basis schema consists of at most $k$ monotone curves. The $i$th monotone curve has a span at most the span of that of the previous monotone curve minus 1, so since the first monotone curve spans $k$ lines, there will be at most $k$ monotone curves.
Each monotone curve intersects the $k$ lines each once, except for the two lines at the top and bottom of its span. However, these can be reduced to one intersection when you concatenate alternating monotone lines together as such.

Thus, the number of times the test-functions of any basis schema crosses the $k$ lines is $\sum_{i = 1}^k i = \frac{k(k + 1)}{2} \leq k^2$.
$\blacksquare$

\paragraph{Convexity of Schema.}
% of partial type (I, $k$) or type (II, $k$)
\begin{lemma} (Convexity of activation set of test-functions of a schema)\label{lem:convexity-of-schema-activations}
    Given a $(k,T)$-configuration, $\{ s_i\}_{i \in [k]}$, consider any schema continuous test-functions. Suppose the schema specifies $M$ segments. Let $A^{(M)}$ be the set of valid $(n_1,\ldots,n_M)$ of segment lengths with $\sum_{i = 1}^M n_i = 1$, where a valid setting of $(n_1,\ldots,n_M)$ satisfies the constraints described in Lemma \ref{lem:schema-constraints}. Then $A^{(M)}$ is a convex set of dimension $M - 1$. Moreover, $A^{(M)}$ is a simplex.
\end{lemma}

\textit{Proof.} $A^{(M)}$ is the intersection of the linear subspace over $(n_1, \ldots, n_M)$ given by $\sum_{i = 1}^M n_i = 1$ along with $M$ halfspaces of the form in Lemma \ref{lem:schema-constraints}:

\begin{align*}
    \forall i \in [M], n_i \geq \frac{p_i}{q_i} \sum_{j = 1}^{i - 1} n_j
\end{align*}

Each $p_i, q_i$ will be determined from the slopes of the lines that segment $i$ first intersects and last intersects. By Lemma \ref{lem:schema-constraints}, there are three cases: 

\textbf{Case 1: Segment $i$ Crosses From Line $j$ to Line $j + 1$.} Then, $ n_i \geq (\frac{s_j}{s_{j + 1}} - 1) \sum_{j = 1}^{i - 1} n_j$.

\textbf{Case 2: Segment $i$ Crosses From Line $j + 1$ to Line $j$.} Then, $ n_i \geq (\frac{1 - s_{j + 1}}{1 - s_{j}} - 1) \sum_{j = 1}^{i - 1} n_j$.

\textbf{Case 3: Segment $i$ Crosses From Line $j$ to Line $j$.} Then, $\frac{p_i}{q_i} = 0.$

First, since $A^{(M)}$ is the intersection of convex sets, it is convex. 

Second, regarding dimension of $A^{(M)}$, since all elements $\{ s_i\}_{i \in [k]}$ of the configuration are distinct and in $(0,1)$, each of the $M$ halfspaces acts on a different subset of the variables $\{ n_i\}_{i \in [M]}$: 
\begin{align*}
    \{n_1\}, \{ n_1, n_2\}, \{ n_1, n_2, n_3\}, \ldots, \{ n_1, \ldots, n_{M - 1}\}, \{ n_1, \ldots, n_M\}
\end{align*}

In particular, no combination of inequality constraint can form an equality constraint which implies that the intersection of the $M$ halfspaces has dimension $M$. Thus, $A^{(M)}$ has dimension $M - 1$ due to the additional constraint $\sum_{i = 1}^M n_i = 1$.

Third, $A^{(M)}$ can be thought of as a polytope over $M - 1$ variables, once we substitute in $n_M = 1 - \sum_{i = 1}^{M - 1} n_i$, while it has exactly $M$ faces given by the $M$ halfspaces, with $n_M = 1 - \sum_{i = 1}^{M - 1} n_i$ substituted into the inequalities. The $M$ faces of $A^{(M)}$ remain distinct even after the substitution since the only face whose linear inequality included $n_M$ will have a bias term of $1$ after the substitution, while the other faces' linear inequality remain homogeneous. Since $A^{(M)}$ has $M$ faces and has dimension $M - 1$, it is a simplex. 

$\blacksquare$

%% file: sections/appendices/margin.tex
%------------------------------------------------------------
% Centroid of Simplex
%------------------------------------------------------------

\subsection{Lemmas for Margin of Point in a Polytope.}\label{appen:margin}

\paragraph{Average of Vertices of a Polytope.}

\iffalse 
\begin{lemma}\label{lem:centroid-formula}
    The centroid of a simplex in dimension $k$ is the average of its $k + 1 $ vertices. 
\end{lemma}

\textit{Proof.} The claim is true for the regular simplex with vertices $e_i, i \in [k]$ and $0$. Any simplex is a linear transformation of the regular simplex, and the centroid of resulting simplex will be the same linear transform applied to the centroid of the regular simplex. $\blacksquare$
\fi 

As a motivation for the following definition, recall that the centroid of a $d$-dimensional simplex is the average of its $d + 1$ vertices. Informally speaking, the centroid has the nice property that it is far from each face of the simplex. This property is useful for our proof, though we will need to extend this average of vertices notion to general convex polytopes. 

\begin{definition} (Average of Vertices)\label{defn:avg-of-vtxs}
Given a convex polytope $P$ of $N$ vertices $v_1, \ldots, v_N$, we will define the average-of-vertices of $P$ as $c = \frac{1}{N}\sum_{i \in [N]} v_i \in P$.
\end{definition}

Note that for a general convex polytope $P$, the average of its vertices will not, in general, be its centroid.

%------------------------------------------------------------
% Precision Lemmas 
%------------------------------------------------------------

\paragraph{Margin of Average of Vertices of Polytope with $\poly(T)\cdot \poly(K)$ Precision Faces.}

\margin*

Often, we will mention the notion of margin in the context of a polytope $P$, where we are interested in the margin of a point $x \in P$ onto a face of $P$. For any face $F$ of $P$, $F$ will be defined as the boundary of the halfspace given by some linear inequality $L$. In this case, WLOG, we will define the margin of any point $x$ in $P$ to $F$ to be $L(x)$. Moreover, we will define $L $ so that $L(x)$ is non-negative for $x \in P$. As such, we will sometimes refer to the margin of a point $x \in P$ onto a face of $P$ as the ``positive" margin to emphasize this point.

%------------------------------------------------------------
% Lemma about margin
%------------------------------------------------------------
We start with the following Lemma.

\begin{lemma}\label{lem:margin}
    Consider a nonempty $(M - 1)$-dimensional convex polytope $P \subset \mathbb{R}^{M - 1}$. Suppose the faces of $P$ consists of $N$ halfspaces over variables $\{ n_i\}_{i \in [M - 1]}$, where each halfspace is given by a linear inequality with integer coefficients of magnitude at most $p$. For $j \in [N]$, define $L_j$ as the linear inequality for the $j$th face. 
    
    Then, for any $j \in [N]$, for any vertex $x$ of $P$ which does not lie on the $j$th face of $P$, then the positive margin  $L_j(x)$ is lower bounded as follows.
    \begin{align*}
        % \forall j \in [N], &\forall \text{ vertex } x_j \text{ of $P$ not on $j$th face of $P$},\\ 
        L_j(x) &\geq \frac{1}{(\sqrt{M}p)^M}
    \end{align*}
\end{lemma}

\textit{Proof.} For each $j \in [N],$ represent the linear inequality $L_j$ as a vector in $v_j \in [-p, p]^{M}$ such that $\forall x' \in \mathbb{R}^{M - 1}, v_j^\top \begin{pmatrix}x' \\-1\end{pmatrix}$ is the margin of $x$ on the constraint given by $L_j$. In addition, in our definition of the vector representations $v_j$, we want to ensure that the vectors $v_j$ are such that $\forall j \in [N], v_j^\top \begin{pmatrix}x \\-1\end{pmatrix} > 0$ for all vertices $x$ which do not lie on the $j$th face of $P$. This is always possible due to the convexity of the polytope, which is contained in the intersection of the halfspaces defined by each of its faces.

As an example of vector representations of inequalities,

\begin{align*}
    n_1 + 2n_2 + 3n_3 > 4 \iff (1,2,3,4)\\
    2n_1 - 3n_2 + 10n_3 \leq -9 \iff (-2,3,-10,9)\\
\end{align*}

WLOG, we will prove the statement for the $1$st face of $P$. Pick any vertex $x$ which does not lie on the $1$st face of $P$ (i.e. $|L_1(x)| > 0$). $x$ is the intersection of $M  - 1$ distinct faces of $P$, whose indices we denote as $\{ j_i\}_{i \in [M - 1]} \subset [N] - \{ 1\}$. Let $\overline{A} \in [-p, p]^{M\times M}$ be a matrix such that for all $i \in [M - 1]$, the $i$th row of $\overline{A}$ is $v_{j_i}$ the vector representation for $L_{j_i}$. Let the $M$th row be $v_1$, the vector representation of $L_1$. Write $\overline{A}$ in block form as:

\begin{align*}
\overline{A} &= \begin{bmatrix}
    A       &   b  \\
    c^\top       &   d
\end{bmatrix}
\end{align*}

where $A \in [-p, p]^{(M - 1)\times (M - 1)}$ and $b,c \in [-p, p]^{M - 1}$ and $d \in [-p, p]$, so that $\begin{pmatrix} c \\d\end{pmatrix}$ is the vector representation of $L_{1}$. We have that $\forall i \in [M - 1], L_{j_i}(x) = (\overline{A}_i)^\top \begin{pmatrix} x \\-1\end{pmatrix}$, where $\overline{A}_i$ denotes the $i$th row of $\overline{A}$. Since $x$ lies on all the faces whose indices are in $\{ j_i\}_{i \in [M - 1]}$,

\begin{align*}
    \forall i \in [M - 1], L_{j_i}(x) &=  0\\
    \implies A x &= b\\
    \implies x &= A^{-1}b\\
    L_{1}(x) &= (\overline{A}_{M})^\top \begin{pmatrix} A^{-1}b \\-1\end{pmatrix}\\
    &= \begin{pmatrix} c \\d\end{pmatrix}^\top \begin{pmatrix} A^{-1}b \\-1\end{pmatrix}\\
    &= c^\top A^{-1}b - d
\end{align*}

Next, we note that

\begin{align*}
    |\overline{A}| &= |  \begin{bmatrix}
    A       &   b  \\
    0       &   d - c^\top A^{-1} b
\end{bmatrix}|\\
&= |A| (d - c^\top A^{-1} b)
\end{align*}

Thus,

\begin{align*}
    L_{1}(x) &= |\frac{|\overline{A}|}{|A|}| > 0
\end{align*}

where the margin is strictly positive since $x_1$ does not lie on face $1$. Its margin has a minimum value of $|\frac{1}{|A|}| \geq \frac{1}{(\sqrt{M} p)^M}$ since the numerator $|\overline{A}|$ is an integer while the denominator is upper bounded by $(\sqrt{M} p)^M$. The upper bound on the determinant of $A$ is by the Geometric-Mean-Quadratic-Mean inequality on the eigenvalues of $A$, where $|A| \leq [\frac{1}{M - 1} ||A||_F^2]^{\frac{M - 1}{2}} \leq (\sqrt{M} \cdot p)^{M}$.

Finally, this lower bound holds for any other face of $P$ and any vertex of $P$ which does not lie on that face by an analogous argument. $\blacksquare$

We now provide a Lemma which can improve the margin lower bound in the case that we know some additional information about the aforementioned matrix $A$ defined in the setting of Lemma \ref{lem:margin}.

% there is a subset of at most $7K$ faces of $A^{(M)}(Y) \cap H_1^{(M)} \cap H_2^{(M)}$ such that each face of the subset has precision at most $T^3$ while the faces not in the subset all have precision at most $(T')^3$.

\begin{lemma}\label{lem:improved-det-bound}
    Suppose $(T, K), (T', K') \in \mathbb{N}^2$ are such that $K \leq T^2$ and $K' \leq (T')^2$. WLOG, suppose $T \geq T'$. Suppose $M \in \nat$ such that $\max(K, K') \leq M \leq 2(K + K')$, and suppose constants $c, d = O(1)$.  Suppose matrix $A \in \mathbb{Z}^{M - 1 \times M - 1}$ such that:
    \begin{enumerate}
        \item $\forall 1 \leq i \leq \min(c K, M - 1), A_i \in \{ -O(T^d), \ldots, O(T^d)\}^{M - 1}$
        \item $\forall \min(cK, M - 1) \leq i \leq M - 1, A_i \in \{ -O((T')^d), \ldots, O((T')^d)\}^{M - 1}$
    \end{enumerate}
    Then $|A| \leq O(T^{O(K)} \cdot (T')^{O(K')})$.
\end{lemma}

\textit{Proof.} If $T, T' \geq 2$, by the homogeneity of the determinant, we can factor out a factor of $O(T^d)$ from each of the first $\min(cK, M - 1)$ rows of $A$ and a factor of $O((T')^d)$ from the remaining rows of $A$.

\begin{align*}
    |A| &\leq (O(T^d))^{\min(cK, M - 1)} \cdot (O((T')^d))^{M - 1 - \min(cK, M - 1)} |\tilde{A}|
\end{align*}

Where $\tilde{A}$ is an $(M - 1) \times (M - 1)$ matrix such that the largest magnitude of any entry in $\tilde{A}$ is 1. Since $T, T' \geq 2$, then there is a universal constant $1 < \eps = O(1)$ such that:

\begin{align*}
    |A| &\leq (T^{\eps \cdot d})^{\min(cK, M - 1)} \cdot ((T')^{\eps \cdot d})^{M - 1 - \min(cK, M - 1)} |\tilde{A}|
\end{align*}

By the same Quadratic-Mean-Geometric-Mean argument in Lemma \ref{lem:margin},

\begin{align*}
    |\tilde{A}| &\leq \Big( \frac{1}{M - 1} ||\tilde{A}||_F^2\Big)^{\frac{M - 1}{2}}\\
    &\leq M^{\frac{M}{2}}
\end{align*}

% Thus, $|A| \leq (T^d)^{\min(cK, M - 1)} \cdot ((T')^d)^{M - 1 - \min(cK, M - 1)} M^{\frac{M}{2}}$.

Suppose that $K \leq K'$. Then $M \leq 4K'$ so that $(T^{\eps d})^{\min(cK, M - 1)} \cdot ((T')^{\eps d})^{M - 1 - \min(cK, M - 1)} \leq T^{\eps cdK} \cdot (T')^{4\eps dK'}$. Meanwhile, since $M \leq 4K'$ and  $K' \leq (T')^2$, then $M^{\frac{M}{2}} \leq (4(T')^2)^{2K'}$, so that $|A| \leq (T^{\eps d})^{\min(cK, M - 1)} \cdot ((T')^{\eps d})^{M - 1 - \min(cK, M - 1)} M^{\frac{M}{2}} \leq T^{\eps cdK} \cdot (T')^{4\eps dK'} \cdot (4(T')^2)^{2K'} \leq O(T^{O(K)} \cdot (T')^{O(K')})$.

Now suppose that $K \geq K'$. Then $M \leq 4K$ so that  $(T^{\eps d})^{\min(cK, M - 1)} \cdot ((T')^{\eps d})^{M - 1 - \min(cK, M - 1)} \leq T^{\eps cdK} \cdot (T')^{4\eps dK} \leq T^{\eps cdK} \cdot T^{4\eps dK}$. Meanwhile, since $M \leq 4K$ and  $K \leq T^2$, then $M^{\frac{M}{2}} \leq (4T^2)^{2K}$, so that $|A| \leq (T^{\eps d})^{\min(cK, M - 1)} \cdot ((T')^{\eps d})^{M - 1 - \min(cK, M - 1)} M^{\frac{M}{2}} \leq T^{\eps cdK} \cdot T^{4\eps dK} \cdot (4T^2)^{2K} \leq O(T^{O(K)} \cdot (T')^{O(K')})$.

In the case $T = 1$ and $T' = 1$, then $M = O(1)$, so $|A| \leq O(1)$. In the case $T \geq 2$ but $T' = 1$, then $K' = (T')^2 = 1$, and following a similar argument as the case where $T, T' \geq 2$ and $K \geq K'$ yields the desired bound.
$\blacksquare$

As a consequence, we have the following finer-grained version of Lemma \ref{lem:margin}.

\begin{lemma}\label{lem:margin-improved}
    Suppose $\alpha \in \mathbb{N}$, $(T, K), (T', K') \in \mathbb{N}^2$ are such that $T^K \leq \alpha$ and $(T')^{K'} \leq \alpha$ and $K \leq T^2$ and $K' \leq (T')^2$. WLOG, suppose $T \geq T'$. Suppose $M \in \nat$ such that $\max(K, K') \leq M \leq 2(K + K')$. Suppose $d_1, d_2 = O(1)$.
    
    Consider a nonempty $(M - 1)$-dimensional convex polytope $P \subset \mathbb{R}^{M - 1}$. Suppose the faces of $P$ consists of $N$ halfspaces over variables $\{ n_i\}_{i \in [M - 1]}$. Suppose that there is a subset of at most $d_1\cdot K$ faces of $P$ such that each face of the subset is given by a linear inequality whose coefficients have precision at most $O(T^{d_2})$ while the faces not in the subset all have precision at most $O((T')^{d_2})$. For $j \in [N]$, define $L_j$ as the linear constraint for the $j$th face. 
    
    Then, for any $j \in [N]$, for any vertex $x$ of $P$ which does not lie on the $j$th face of $P$, then the positive margin  $L_j(x)$ is lower bounded as follows.
    \begin{align*}
        % \forall j \in [N], &\forall \text{ vertex } x_j \text{ of $P$ not on $j$th face of $P$},\\ 
        L_j(x) &\geq \Omega(\frac{1}{\alpha^{O(1)}})
    \end{align*}
\end{lemma}

\textit{Proof.} We follow the proof of Lemma \ref{lem:margin} up to the step where we've defined $A$ and $\overline{A}$ and where we deduce that for any vertex $x$ of $P$ which does not lie on the first face of $P$, 

\begin{align*}
    L_1(x) = |\frac{|\overline{A}|}{|A|}| > 0
\end{align*}

Because $A$ satisfies the pre-conditions Lemma \ref{lem:improved-det-bound} once we swap of rows appropriately, then $|A| \leq O(T^{O(K)} (T')^{O(K')}) \leq O(\alpha^{O(1)})$. Thus, $L_1(x) \geq \Omega(\frac{1}{\alpha^{O(1)}})$. An analogous argument holds for all the other faces of $P$. $\blacksquare$

The following corollary applies Lemma \ref{lem:margin-improved} to our setting of interest.

\begin{corollary}\label{cor:margin-of-C}
    Suppose $\alpha \in \mathbb{N}$, $(T, K), (T', K') \in \mathbb{N}^2$ are such that $T^K \leq \alpha$ and $(T')^{K'} \leq \alpha$ and $K \leq T^2$ and $K' \leq (T')^2$. WLOG, suppose $T \geq T'$. Suppose $M \in \nat$ such that $\max(K, K') \leq M \leq 2(K + K')$. Suppose $d_1, d_2 = O(1)$.
    
    Denote the coordinates in an $M$-dimensional Euclidean space $E$ as $(n_1, \ldots, n_M)$. Consider a nonempty $(M - 1)$-dimensional polytope $P \subset E$, such that $\clo(P)$ has vertices $V$, with $|V| \geq M$, and $N$ faces. Suppose $P$ is contained in the $(M - 1)$-dimensional subspace given by $\sum_{i \in [M]} n_i = 1$. Suppose there is a subset of at most $d_1 \cdot K$ faces of $P$ such that each face of the subset is given by a linear inequality whose coefficients have precision at most $O(T^{d_2})$ while the faces not in the subset all have precision at most $O((T')^{d_2})$. For $j \in [N]$, define $L_j$ as the linear constraint for the $j$th face. 
    
    Then the average of the $|V|$ vertices of $P$ will have margin at least $\Omega(\frac{1}{|V|} \frac{1}{\alpha^{O(1)}})$ for all $(M - 2)$-dimensional faces of $P$.
    
    %Then for every $(M - 1)$-dimensional simplex $S$ determined by $M$ vertices in $V$, the centroid of $S$ will have margin at least $\frac{1}{M} \frac{1}{(2\sqrt{M}p)^M}$ for all faces of $C$.
\end{corollary}

\textit{Proof. } To reduce this to the setting of Lemma \ref{lem:margin-improved}, first substitute $n_M = 1 - \sum_{i = 1}^{M - 1}n_i$ in the inequalities for all $N$ faces of $P$ so that each inequality is an equivalent inequality over $(n_1,\ldots,n_{M - 1})$ with integer coefficients increased by a factor of at most 2 (which will ultimately be absorbed by the Big-O notation).

Applying Lemma \ref{lem:margin-improved} on $P$, where $n_M = 1 - \sum_{i = 1}^{M - 1}n_i$ was substituted into the inequalities defining faces of $P$, implies that for every $(M - 2)$-dimensional face $F$ of $\clo(P)$ and every vertex $x$ of $\clo(P)$ which does not lie on $F$, then the margin of $x$ on $F$ is at least $\Omega(\frac{1}{\alpha^{O(1)}})$. All vertices which lie on $F$ have margin $0$.

% Consider any simplex $S$ consisting of $M$ vertices in $\{ x_1, \ldots, x_M\}\subset V$. By Lemma \ref{lem:centroid-formula}, its centroid is given by:

Consider $c$, the average of vertices $\{ x_1, \ldots, x_{|V|}\}$ of $P$.

\begin{align*}
    c &= \frac{1}{|V|} \sum_{i = 1}^{|V|}x_i
\end{align*}

For any face $F$ of $P$ with constraint given by $L_F(\cdot)$, at least one of the $|V|$ vertices $\{ x_1, \ldots, x_{|V|}\}$ does not lie on $F$, as $F$ is $(M - 2)$-dimensional while $P$ is $(M - 1)$-dimensional. By linearity of the margin,

\begin{align*}
    L_F(c) &= \frac{1}{|V|} \sum_{i = 1}^{|V|} L_F(x_i)\\
    &\geq \Omega(\frac{1}{|V|} \frac{1}{\alpha^{O(1)}})
\end{align*}
$\blacksquare$

Finally, we need one more Lemma to bound the number of vertices of the polytope of interest in the main proof.

\begin{lemma} \label{lem:num-vtxs-of-P}
    Denote the coordinates in an $M$-dimensional Euclidean space $E$ as $(n_1, \ldots, n_M)$. Consider a nonempty $(M - 1)$-dimensional polytope $P \subset E$, such that $\clo(P)$ has vertices $V$, with $|V| \geq M$, and $N$ faces. Suppose $P$ is contained in the $(M - 1)$-dimensional subspace given by $\sum_{i \in [M]} n_i = 1$, and $P$ is the intersection of an $(M-1)$-dimensional simplex and two halfspaces. Then, the number of vertices of $P$ is at most $3M^2$.
\end{lemma}

\textit{Proof.} Let $P = A \cap H_1 \cap H_2$, where $A$ is the $(M-1)$-dimensional simplex, and $H_1, H_2$ are the halfspaces. First, $A$ will have $M$ vertices and $M$ faces. 

The new vertices formed by intersecting $H_1$ with $A$ can be upper bounded by counting the number of tuples of $M - 1$ faces of $A$, as each vertex is formed by the intersection of $M$ faces. $H_1$, when intersecting $A$ will be able to form at most ${M \choose M - 1} = M$ new vertices. $H_2$, when intersecting $A \cap H_1$, will be able to form at most ${M + 1 \choose M - 1} = \frac{M (M + 1)}{2}$ new vertices. The total vertices of $A \cap H_1 \cap H_2$ is at most $2M + \frac{M (M + 1)}{2} \leq 3M^2$. $\blacksquare$

%% file: sections/appendices/disc-approx.tex
%------------------------------------------------------------
% Discrete Approximation
%------------------------------------------------------------
\subsection{Lemmas for Discrete Approximation.} \label{appen:disc-approx}

In this section, we are interested in showing that an activation $(B_1, \ldots, B_k)$ realized by some continuous test-function $\caly$ can also be approximately realized by some discrete test-function $\calx$. The difficulty is that discrete test-functions consist of a sequence of lattice points  $\{ (j,ps(x)_j) \}_{j \in [|x|]}$, where successive lattice points differ by a step of $(+1, 0)$ (corresponding to $x_{j} = 0$) or $(+1, +1)$ (corresponding to $x_{j} = 1$).

First, here is an auxiliary Lemma that provides a strategy to construct a trajectory of lattice points where successive lattice points differ by a step of $(+1, 0)$ or $(+1, +1)$ and where the trajectory stays between two lines $\yax = \frac{b}{a} \xax$ and $\yax = \frac{d}{c}\xax$. The strategy offers a way to iteratively find the next lattice point in the trajectory.

\begin{lemma}\label{lem:strategy}
    For any two 2D lines with slopes $1 > \frac{b}{a} > \frac{d}{c} > 0$ where $a,b,c,d \leq T$, there exists an infinite sequence of bits $x \in \{ 0,1\}^{\infty}$ such that for all $n \geq T^2$ we have that $\frac{b}{a}n \geq \ps(x)_n > \frac{d}{c}n$. That is, the discrete test-function given by $x$  is above the lower line and below the upper line.
\end{lemma}

\textit{Proof.} $\min_{0 \leq a,b,c,d \leq T, \frac{b}{a} \neq \frac{d}{c}}(\frac{b}{a} - \frac{d}{c}) \geq \frac{1}{T^2}$, so for $n \geq T^2$, $(\frac{b}{a} - \frac{d}{c})n \geq 1$. Thus, for every $n \geq T^2$, there is at least one lattice point $(n, y_n) \in \mathbb{Z}^2$ with $\frac{b}{a}n \geq y_n > \frac{d}{c}n$.

Now let's consider a strategy to jump from $(n, y_n)$ to $(n + 1, y_{n + 1})$. For all $n \geq T^2$, from any such lattice point, $(n, y_n)$, either $(n + 1, y_{n })$ or $(n + 1, y_{n} + 1)$  will be in $(\frac{d}{c}(n + 1), \frac{b}{a}(n + 1)]$. Thus, there is always a continuation of the trajectory that remains between the two lines; one either takes a step along $(+1, 0)$ (corresponding to $x_{n + 1} = 0$) or along $(+1, +1)$ (corresponding to $x_{n + 1} = 1$) to reach $(n + 1, y_{n })$ or $(n + 1, y_{n} + 1)$, respectively. 

Finally, the trajectory from $(0,0)$ to  $(T^2, y_{T^2})$ can always be made by some trajectory using increments $(+1, +1)$ and $(+1, 0)$ since $y_{T^2} \leq T^2 \frac{b}{a} < T^2$. $\blacksquare$

The following Lemma says that we can approximate any continuous test-function $\caly$ with a length $n$ discrete test-function, for $n$ above some threshold.

% \tc{schema is not well defined for discrete test-fn. Maybe just say the end goal: that B(x) - B(y) is at most $T^2/n$}

\begin{lemma}\label{lem:low-prec-activ-to-string} (Discrete Approximation to Continuous Test-Function) Suppose $ \in \mathbb{N}$, $(T, K), (T', K') \in \mathbb{N}^2$ are such that $T^K \leq \alpha$ and $(T')^{K'} \leq \alpha$ and $K \leq T^2$ and $K' \leq (T')^2$. WLOG, suppose $T \geq T'$. Suppose we are given a $(k,T)$-configuration $\{ s_i\}_{i \in [k]}$, where a subset of $\{ s_i\}_{i \in [k]}$ of size at most $K$ have precision at most $T$, while the remaining entries of $\{ s_i\}_{i \in [k]}$ have precision at most $T'$. For any schema $Y$ of $M$ segments, suppose $\caly$ is any continuous test-function of schema $Y$, with segment lengths $(\overline{n}_1(\caly), \ldots, \overline{n}_M(\caly)) \in [0,1]^M$ (where we assumed WLOG that $\sum_{j} \overline{n}_j(\caly) = 1$). Suppose every $\overline{n}_j(\caly)$ is a rational number and that the common denominator of all $( \overline{n}_j(\caly))_{j \in [M]}$ is $p$.
    
Then there exists an $n_0 \leq O(p \cdot \alpha^2)$ so that for any positive integer multiple $n$ of $n_0$, there exists a discrete test-function $\calx$ of length $n$ so that 

\begin{align*}
    \forall i \in [k], |B_i(\caly) - B_i(\calx)| \leq \frac{T^2 + M}{n}
\end{align*}
    
\end{lemma}

\textit{Proof.} Denote $\textup{idx}(i) \in [k]$ as the line which the end-point of segment $i$ of the schema $Y$ (the schema of $\caly$) lies on.

First, given the $k$ slopes $s_1, s_2, \ldots s_k \in (0,1)$, suppose that the denominators of these slopes are $a^{(1)}, \ldots, a^{(k)}$, where $\forall i \in [k], s_i = \frac{b^{(i)}}{a^{(i)}}$. There is a subset   $U\subset[k]$ of size at most $K$ such that $\{ a^{(i)} : i \in U\}$ are all positive integers of precision at most $T$. Further, $\{ a^{(i)} : i \notin U\}$ is a set of at most $K'$  positive integers of precision at most $T'$. Let $d$ be the least common multiple of $\{ a^{(1)}, \ldots, a^{(k)}\}$. Note that $d \leq T^K \cdot (T')^{K'} \leq \alpha^2$. 

Now, set $n_0 = d \cdot p$, and consider any multiple $n$ of $n_0$. Consider the quantities, $\{ n_i(\caly)\}_{i \in [M]} := \{ n \cdot \overline{n}_i(\caly) \}_{i \in [M]}$, which are the segment lengths $\caly$, rescaled by a factor of $n$. We will refer to this rescaled $\caly$ instead of $\caly$ for convenience of notation in the proof; like a discrete test-function of length $n$, the rescaled $\caly$ will be a map from $[0,n] \to [0, n]$. Denote $\delta_i := \sum_{j = 1}^i  n_j(\caly)$ for all $i \in [M]$ and $\delta_0 = 0$.

% We will find a discrete test-function of length $n$ with segment lengths $\{ \Delta_i^* - \Delta_{i - 1}^* \pm (T^2 + T) \}_{i \in [M]}$. 

We will construct a discrete test-function which follows the same schema as the rescaled $\caly$ on the interval $[T^2, n]$, though on the interval $[0, T^2]$, $\calx$ will have no guarantees. 

First, we will define $M + 1$ lattice points where the discrete test-function will cross over the $k$ lines (``crossing points"). Second, we show how to connect those crossing points with a sequence of lattice points, where consecutive lattice points differ by increments of the 2D vectors, $(+1, +1)$ and $(+1, 0)$ (so that the sequence of lattice points is like a discrete test-function). At the end, we will argue that the constructed discrete test-function $\calx$ and the rescaled $\caly$ will satisfy the guarantee that $\forall i \in [k], |B_i(\caly) - B_i(\calx)| \leq \frac{M + T^2}{n}$.

% will show the approximation error of such an approximation strategy is at most  $\frac{T^2 + T}{n}$.

\paragraph{Step 1: Defining the Crossing Points of the Discrete Test-Function on the $k$ Lines.}

The rescaled continuous test-function $\caly$, where the input domain and the output is scaled by $n$, will cross the $k$ lines $l_1, \ldots, l_k$ at the $M + 1$ points, 

\begin{align*}
    \{ ( \delta_i, s_{\textup{idx}(i)} \delta_i)\}_{i \in [M]} \cup \{ (0,0)\}
\end{align*}
 
Due to the way we have set $n$, these are all lattice points. Since $n = d \cdot p$, and all $\overline{n}_i(\caly)$ have common denominator $p$, then $\overline{n}_i(\caly)$  is an integer for all $i \in [M]$, so $\delta_i$  is an integer for all $i \in [M]$. Second, because of the factor of $d$, $s_{\textup{idx}(i)}\delta_i$ is an integer. They are therefore valid points for the discrete test-function to go through, and we will design a discrete test-function that crosses the $k$ lines at $\{ ( \delta_i, s_{\textup{idx}(i)} \delta_i)\}_{i \in [M]} \cup \{ (0,0)\}$ for all $\delta_i \geq T^2$.

\paragraph{Step 2: Connecting the Crossing Points.}

We'll describe a strategy for constructing the discrete test-function that connects these crossing points. For any segment in the continuous schema, it will either cross between two consecutive lines, or it will cross a line and then recross the same line. We need only show how to do these types of crossings in a way that only crosses the appropriate lines at the start and endpoint of the segment $\{ (\delta_i, s_{\textup{idx}(i)} \delta_i)\}_{i \in [M]}$ and at no point in between. Consider the $(i + 1)$th segment of $\caly$, which has start-point $(\delta_{i }, s_{\textup{idx}(i )} \delta_{i })$ and end-point $(\delta_{i + 1}, s_{\textup{idx}(i + 1)} \delta_{i + 1})$. There are three cases: 

\begin{itemize}
    \item The segment crosses down (segment start-point on line $l_j$ and end-point on line $l_{j + 1}$ for some $j \in [k - 1]$)
    \item The segment crosses up (segment start-point on line $l_{j + 1}$ and end-point on $l_{j}$ for some $j \in [k - 1]$)
    \item The segment crosses and re-crosses the same line (segment start and end-point on $l_j$ for some $j \in [k]$)
\end{itemize}

The strategies below will apply when the segment's start-point's $\xax$-coordinate is at least $T^2$ (i.e. $\delta_i \geq T^2$). After describing these, we will handle the corner case where a segment's start-point's $\xax$-coordinate is less than $T^2$ but its end-point's $\xax$-coordinate is larger than $T^2$.

\textbf{Case 1: Segment Crosses Down.} Suppose that we must connect the following start and end-point with a sequence of lattice points which lie between lines $l_j, l_{j + 1}$:

\begin{align*}
    (\delta_i, s_j \cdot \delta_i), (\delta_{i + 1}, s_{j + 1}\cdot \delta_{i + 1}) \in \mathbb{Z}^2
\end{align*}

The corresponding segment of the rescaled continuous test-function is the $(i + 1)$th segment with length $n_{i + 1}(\caly)) \in \mathbb{Z}$.

First, there exists a sequence (trajectory) of lattice points $\{(\xax_t, \yax_t)\}_{0 \leq t \leq n_{i + 1}(\caly)}$ where $(\xax_0, \yax_0) = (\delta_i, s_j \cdot \delta_i) $, $(\xax_{n_{i + 1}(\caly)}, \yax_{n_{i + 1}(\caly)}) = (\delta_{i + 1}, s_{j + 1}\cdot \delta_{i + 1})$ where for all $0 \leq t < n_{i + 1}(\caly), \xax_{t + 1} = \xax_{t } + 1$; and $\yax_{t + 1} = \yax_{t} + 1$ or $\yax_{t + 1} = \yax_{t}$. This is because the rescaled $\caly$ is a continuous test-function which connects these two lattice points and which is $1$-Lipschitz and non-decreasing. We want to further show that there exists such a trajectory of lattice points $\{(\xax_t, \yax_t)\}$ such that $\forall 1 \leq t \leq n_{i + 1}(\caly) - 1, s_{j} \cdot \xax_t \geq \yax_t > s_{j + 1} \cdot \xax_t$.

% Now we'll describe a strategy to connect the start and end-point with a discrete test-function that only intersects $l_j, l_{j + 1}$ at the start and end-point. 

The strategy to do so constructs a trajectory from the start-point $(\delta_i, s_j \cdot \delta_i)$ to the end-point $ (\delta_{i + 1}, s_{j + 1}\cdot \delta_{i + 1})$ and consists of two phases:

\begin{itemize}
    \item (Phase 1) First, the following iterative procedure is applied, starting with  $(\xax_0, \yax_0) = (\delta_i, s_j \cdot \delta_i)$ and $t = 0$:
    
    \begin{adjustwidth}{2em}{0pt}
    From the current lattice point $(\xax_t, \yax_t)$, choose $(\xax_{t + 1}, \yax_{t + 1}) = (\xax_t + 1, \yax_t)$ or $(\xax_t + 1, \yax_t + 1)$ depending on which one satisfies $s_{j} \cdot \xax_{t + 1} \geq \yax_{t + 1} > s_{j + 1} \cdot \xax_{t + 1}$. At least one of the two will always satisfy this condition if $\xax_0 \geq T^2$, as argued in Lemma \ref{lem:strategy}. If both are possible, then choose either. %, then take $(\xax_t + 1, \yax_t + 1)$. 
    \end{adjustwidth}
    
    This is repeated until, $\yax_t$, the $\yax$-coordinate of the current lattice point equals $s_{j + 1} \delta_{i + 1}$.

\item (Phase 2) The trajectory only takes steps along the direction $(+1,0)$ so that $(\xax_{t + 1}, \yax_{t + 1}) = (\xax_{t} + 1, \yax_{t})$ until the end-point is reached.
\end{itemize}

In Phase 1, we will always be able to find a next point $(\xax_{t + 1}, \yax_{t + 1})$ between $l_j, l_{j + 1}$ from a current point $(\xax_{t}, \yax_{t })$ between $l_j, l_{j + 1}$, if $\xax_0 \geq T^2$ as argued in Lemma \ref{lem:strategy}. By induction, all such lattice points will lie between the two lines $l_j, l_{j + 1}$. Phase 1 will terminate because $\yax_t$ is non-decreasing in $t$ and cannot remain bounded by any constant as $t$ increases to $\infty$, in order to stay between $l_j, l_{j + 1}$, since $s_{j + 1} > 0$. Thus, there exists some minimal $t_*$ such that $\yax_{t_*} = s_{j + 1} \delta_{i + 1}$ for the first time. Because the lattice points always stay between $l_j, l_{j + 1}$, then at time $t_*$, we must have that $\xax_{t_*} < \delta_{i + 1}$. Thus, Phase 1 terminates and Phase 2 can commence. 

It is clear that Phase 2 will yield a set of lattice points that are between $l_j, l_{j + 1}$ with the last lattice point being $ (\delta_{i + 1}, s_{j + 1}\cdot \delta_{i + 1})$ as desired.

This connects the two points with a trajectory that does not intersect $l_j$ and $l_{j + 1}$ at any points besides the start and endpoint, as neither phase crosses the two lines.

\textbf{Case 2: Segment Crosses Up.} 

Suppose that we must connect the following start and end-point with a sequence of lattice points which lie between lines $l_j, l_{j + 1}$:

\begin{align*}
    (\delta_i, s_{j + 1} \cdot \delta_i), (\delta_{i + 1}, s_{j}\cdot \delta_{i + 1})
\end{align*}

% One can reduce Case 2 to precisely Case 1 because of a one-to-one correspondence between sequences of lattice points from $(\delta_i, s_{j + 1} \cdot \delta_i)$ to $ (\delta_{i + 1}, s_{j}\cdot \delta_{i + 1})$ which must stay between lines $l_j$ and $l_{j + 1}$ and sequences of lattice points from $(\delta_i, (1 - s_{j + 1}) \cdot \delta_i)$ to $ (\delta_{i + 1}, (1 - s_{j}) \cdot \delta_{i + 1})$ which must stay between lines with slopes $(1 - s_{j + 1})$ and $(1 - s_{j})$. A path from the former corresponds to one of the latter by flipping all the $(+1, 0)$ moves to $(+1, +1)$ moves and all $(+1, +1)$ moves to $(+1, 0)$ moves. No--there is asymmetric due to the > and <=

The argument is analogous. First, there exists a sequence of lattice points $\{(\xax_t, \yax_t)\}$ where $(\xax_0, \yax_0) = (\delta_i, s_{j + 1} \cdot \delta_i) $, $(\xax_{n_{i + 1}(\caly)}, \yax_{n_{i + 1}(\caly)}) = (\delta_{i + 1}, s_{j}\cdot \delta_{i + 1})$ where for all $0 \leq t < n_{i + 1}(\caly), \xax_{t + 1} = \xax_{t } + 1$ and $\yax_{t + 1} = \yax_{t} + 1$ or $\yax_{t + 1} = \yax_{t}$. This is because the rescaled $\caly$ is a continuous test-function which connects these two lattice points and which is $1$-Lipschitz and non-decreasing. We want to further show that there exists such a trajectory of lattice points $\{(\xax_t, \yax_t)\}$ such that $\forall 1 \leq t \leq n_{i + 1}(\caly) - 1, s_{j} \cdot \xax_t \geq \yax_t > s_{j + 1} \cdot \xax_t$.

The strategy to do so consists of two phases:

\begin{itemize}
    \item (Phase 1) First, the following iterative procedure is applied, starting with  $(\xax_0, \yax_0) = (\delta_i, s_{j + 1} \cdot \delta_i)$ and $t = 0$:
    
    \begin{adjustwidth}{2em}{0pt}
    From the current lattice point $(\xax_t, \yax_t)$, choose $(\xax_{t + 1}, \yax_{t + 1}) = (\xax_t + 1, \yax_t)$ or $(\xax_t + 1, \yax_t + 1)$ depending on which one satisfies $s_{j} \cdot \xax_{t + 1} \geq \yax_{t + 1} > s_{j + 1} \cdot \xax_{t + 1}$. At least one of the two will always satisfy this condition if $\xax_0 \geq T^2$, as argued in Lemma \ref{lem:strategy}. If both are possible, then chose either. % $(\xax_t + 1, \yax_t )$. 
    \end{adjustwidth}
    
    This is repeated until the current lattice point $(\xax_t, \yax_t)$ and the end-point $(\delta_{i + 1}, s_{j}\cdot \delta_{i + 1})$ can be connected with a line of slope 1. That is, $s_{j}\cdot \delta_{i + 1} - \yax_t = \delta_{i + 1} - \xax_t$.

\item (Phase 2) The trajectory only takes steps along the direction $(+1,+1)$ so that $(\xax_{t + 1}, \yax_{t + 1}) = (\xax_{t} + 1, \yax_{t} + 1)$ until the end-point is reached.
\end{itemize}

The analysis of this strategy is analogous to that in Case 1.

\textbf{Case 3: Segment Crosses and Recrosses the Same Line, $l_j$.} 

Suppose that we must connect the following start and end-point with a sequence of lattice points.

\begin{align*}
    (\delta_i, s_{j } \cdot \delta_i), (\delta_{i + 1}, s_{j}\cdot \delta_{i + 1})
\end{align*}

This can always be accomplished by a discrete test-function if $\delta_i \geq T^2$. There are two subcases: (1) if the trajectory needs to cross above line $j$ then recross line $j$, and (2) if the trajectory needs to cross below line $j$ and then recross line $j$. In case (1), we require the sequence of lattice points to lie between lines $l_{j - 1}, l_j$ while in case (2), we require the sequence of lattice points to lie between lines $l_{j}, l_{j + 1}$.

A simple strategy for subcase (1) is to first set $(\xax_1, \yax_1) = (\delta_i + 1, s_j \delta_i + 1)$ (i.e. move one step along to $(+1, +1)$ from the start-point). Note that $(\delta_i + 1, s_j \delta_i + 1)$ will not lie above line $j - 1$ since $\delta_i \geq T^2$, so $(s_{j - 1} - s_j)\delta_i \geq 1$ and $s_j \delta_i + 1 \leq s_{j - 1} (\delta_i + 1)$. The trajectory will now be above line $j$ and below line $j - 1$. Next, use Lemma \ref{lem:strategy} to iteratively find the next lattice point $(\xax_{t + 1}, \yax_{t + 1})$ from the current lattice point $(\xax_{t}, \yax_{t})$ by taking steps that are either $(+1, +1)$ or $ (+1, 0)$ while staying between $l_j, l_{j - 1}$ until $\yax_t = s_{j}\cdot \delta_{i + 1}$. Then, take the remaining steps along $(+1, 0)$ until the end-point is reached.

An analogous strategy exists for subcase (2). Move one step according to $(+1, 0)$ so that $(\xax_1, \yax_1)$ is below line $j$ and above line $j + 1$. Use Lemma \ref{lem:strategy} to move to the next lattice point strictly between line $j$ and $j + 1$ until $(\delta_{i + 1} - \xax_t =  s_{j}\cdot \delta_{i + 1}  - \yax_t)$.

Notice that this strategy is similar to those in Cases 1 and 2, and the analysis of them is the same.

\paragraph{Step 3: Constructing the Final Discrete Test-Function.}

We have proved that for each $i \in [M]$, the two crossing points $(\delta_{i - 1}, s_{\textup{idx}(i - 1)} \delta_{i - 1})$ and $(\delta_i, s_{\textup{idx}(i)} \delta_i)$  can be connected by some sequence of lattice points that only crosses the $k$ lines $\{ l_1, \ldots, l_k\}$ at the crossing points, as long as $\delta_{i - 1} \geq T^2$. The lattice points differ from each other by increments of $(+1, 0)$ and $(+1, +1)$, like a discrete test-function. Each sequence of lattice points can be identified with one segment of the rescaled $\caly$ so that the number of lattice points, excluding $(\delta_{i - 1}, s_{\textup{idx}(i - 1)} \delta_{i - 1})$, equals $n_i(\caly)$.%, and all these lattice points lie in the same sector as the $i$th segment of $\caly$.

% Moreover, the lattice points have the property that $\delta_{i} - \delta_{i - 1}$ is exactly the number of lattice points 

Before constructing the final discrete test-function, we will need to construct the rest of the lattice points of the discrete test-function corresponding to segments of $\caly$ whose start-point is not larger than $T^2$. 

Suppose that the $M_0$th segment of $\caly$ is such that $0 \leq \delta_{M_0 - 1} \leq T^2 - 1 < \delta_{M_0}$, where $M_0 \in [M]$. We split the $M_0$th segment up into two pieces. Consider the restriction of the $M_0$th segment to the following domains: $[\delta_{M_0 - 1}, T^2]$ and $[T^2, \delta_{M_0}]$. We will construct a sequence of lattice points for the restriction of the $M_0$th segment on $[T^2, \delta_{M_0}]$ using similar techniques as in Step 2. We will then construct a sequence of lattice points from $(0,0)$ to $(T^2, \yax_*)$ for some $\yax_*$, to complete the discrete test-function. There are three cases to consider.

\textbf{Case 1: } $\textup{idx}(M_0 - 1) = j, \textup{idx}(M_0) = j + 1$ for some $j \in [k - 1]$.

We have that $(s_j - s_{j + 1})(T^2) \geq 1$, so there exists a lattice point $(T^2, \yax_*)$ such that $\yax_* \in (s_{j + 1}T^2, s_j T^2]$, with $\yax_* \in \mathbb{Z}$. We then construct a sequence of lattice points from $(T^2, \yax_*)$ to $(\delta_{M_0}, s_{\textup{idx}(M_0)} \delta_{M_0})$ using the same strategy as in Step 2, Case 1. This sequence of lattice points will be such that consecutive lattice points differ by an increment of $(+1, +1)$ or $(+1, 0)$ and all lattice points are between $l_j$ and $l_{j + 1}$.

We also construct a sequence of lattice points from $(0,0)$ to $(T^2, \yax_*)$. The only requirement of this sequence of lattice points is that consecutive lattice points differ by  an increment of $(+1, +1)$ or $(+1, 0)$. Since $\yax_* \leq s_j T^2 < T^2$, this will be possible. 

% In this case, the segment is meant to go down, in the sense where its start point is on a line of higher slope and its endpoint on a line of lower slope. Starting from the end point and constructing in reverse, construct any trajectory of lattice points from $(\delta_{i + 1}, s_{j + 1} \delta_{i + 1})$ to  $(T^2 , y)$ for any integer $y \in (s_{j + 1}T^2, s_{j}T^2]$ where consecutive lattice points differ by either $(+1, +1)$ or $(+1, 0)$, and such that the trajectory does not cross $l_j, l_{j + 1}$ anywhere except at $(\delta_{i + 1}, s_{j + 1} \delta_{i + 1})$ and potentially at $(T^2 , y)$. This will always be possible by Lemma \ref{lem:strategy}. 

% In addition, it will be possible to create some trajectory of lattice points from $(0,0)$ to $(T^2 , y)$  using increments of $(+1, +1), (+1, 0)$ since $y \leq s_{j}T^2 < T^2$. The concatenation of all these segments gives the entire discrete test-function.

\textbf{Case 2: } $\textup{idx}(M_0 - 1) = j + 1, \textup{idx}(M_0) = j$ for some $j \in [k - 1]$.

% In this case, the segment is meant to go up, starting on a line of lower slope and ending on a line of higher slope. Construct, in reverse, any trajectory of lattice points from $(\delta_{i + 1}, s_{j + 1} \delta_{i + 1})$ to  $(T^2 , y)$ for any integer $y \in (s_{j + 1}T^2, s_{j}T^2]$ consisting of $(-1, -1)$ and $(-1, 0)$, with all the usual properties, which will always be possible by Lemma \ref{lem:strategy}. In addition, it will be possible to create some discrete trajectory from $(0,0)$ to $(T^2 , y)$ using $(+1, +1), (+1, 0)$ since $y \leq s_{j}T^2 < T^2$. The concatenation of all these segments gives the entire discrete test-function.

Similar to Case 1. We have that $(s_j - s_{j + 1})(T^2) \geq 1$, so there exists a lattice point $(T^2, \yax_*)$ such that $\yax_* \in (s_{j + 1}T^2, s_j T^2]$, with $\yax_* \in \mathbb{Z}$. We then construct a sequence of lattice points from $(T^2, \yax_*)$ to $(\delta_{M_0}, s_{\textup{idx}(M_0)} \delta_{M_0})$ using the same strategy as in Step 2, Case 2. This sequence of lattice points will be such that consecutive lattice points differ by an increment of $(+1, +1)$ or $(+1, 0)$ and all lattice points are between $l_j$ and $l_{j + 1}$.

We also construct a sequence of lattice points from $(0,0)$ to $(T^2, \yax_*)$. The only requirement of this sequence of lattice points is that consecutive lattice points differ by  an increment of $(+1, +1)$ or $(+1, 0)$. Since $\yax_* \leq s_j T^2 < T^2$, this will be possible. 

\textbf{Case 3: } $\textup{idx}(M_0 - 1) = j, \textup{idx}(M_0) = j$ for some $j \in [k]$.

There are two subcases: either segment $M_0$ is contained in $\textup{Sector}_{j}$ or in $\textup{Sector}_{j + 1}$ (Sectors are defined in \Cref{defn:sectors}). We will just consider the first subcase, as the second is analogous. Define $s_0 = 1$ and $l_0$ as the homogeneous, 2D line with slope $s_0$. 

We have that $(s_{j - 1} - s_{j })(T^2) \geq 1$, so there exists a lattice point $(T^2, \yax_*)$ such that $\yax_* \in (s_{j }T^2, s_{j - 1} T^2]$, with $\yax_* \in \mathbb{Z}$. We then construct a sequence of lattice points from $(T^2, \yax_*)$ to $(\delta_{M_0}, s_{\textup{idx}(M_0)} \delta_{M_0})$ using the same strategy as in Step 2, Case 1. This sequence of lattice points will be such that consecutive lattice points differ by an increment of $(+1, +1)$ or $(+1, 0)$ and all lattice points are between $l_{j - 1}$ and $l_{j }$.

We also construct a sequence of lattice points from $(0,0)$ to $(T^2, \yax_*)$. The only requirement of this sequence of lattice points is that consecutive lattice points differ by  an increment of $(+1, +1)$ or $(+1, 0)$. Since $\yax_* \leq s_{j - 1} T^2 \leq T^2$, this will be possible. 
\\
\\ 
We now construct the full discrete test-function by concatenating all these sequences of lattice points into a single sequence of lattice points. For each $i \in [M_0, M]$, use the aforementioned constructions to construct a sequence of lattice points with start-point $(\max(\delta_{i - 1}, T^2), s_{\textup{idx}(i - 1)} \cdot \max(\delta_{i - 1}, T^2))$ and end-point $(\delta_i, s_{\textup{idx}(i)} \delta_i)$. Remove the start-point from each of these sequences of lattice points, and concatenate all the sequences of lattice points for $i \in [M]$.  Finally, concatenate the sequence of lattice points from $(0,0)$ to $(T^2, \yax_*)$, for $\yax_*$ as described in Case 3 above. This sequence of lattice points is a valid discrete test-function since consecutive lattice points differ by an increment of $(+1, +1)$ or $(+1, 0)$. We will refer to the constructed discrete test-function as $\calx$.

\paragraph{Properties of the Final Construction.}

For $i \in [M_0, M]$, there is a correspondence between the $i$th segment of $\caly$  and the sequence of lattice points with start-point $(\max(\delta_{i - 1}, T^2), s_{\textup{idx}(i - 1)} \cdot \max(\delta_{i - 1}, T^2))$ and end-point $(\delta_i, s_{\textup{idx}(i)} \delta_i)$. The difference between the number of lattice points in the latter and the length of the former is at most $T^2$. In addition, all except possibly one of those lattice points (the end-point) lie in the sector in which the $i$th segment of $\caly$ lies in. 

Define $(B_i(\caly))_{i \in [k]}$ and $(B_i(\calx))_{i \in [k]}$ as follows.

\begin{align*}
    \forall i \in [k], B_i(\caly) &:= \int_0^1 \ind [\caly(j) > s_i \cdot j]dj\\
    \forall i \in [k],  B_i(\calx) &:= \frac{1}{n}\sum_{j = 1}^n \ind [\calx(j) > s_i \cdot j]
\end{align*}

For any $i \in [k]$, $B_i(\calx)$ depends on the total number of lattice points (out of $n$) which are strictly above line $i$, while $B_i(\caly)$ depends on the total length of segments which lie above line $i$. The maximum deviation between $B_i(\calx)$ and $B_i(\caly)$ is upper bounded by $\frac{1}{n}$ multiplied by the number of lattice points $(j, \calx(j)), j \in [n],$ which are not in the Sector which their corresponding segment is in, which is upper bounded by $\frac{T^2}{n} + \frac{\text{\# times $\caly$ crosses line $i$}}{n} = \frac{T^2 + M}{n}$. More precisely, for any $i \in [k]$, 

\begin{align*}
    |B_i(\calx) - B_i(\caly)| &\leq \frac{1}{n}\sum_{m \in [M]} \sum_{\delta_{m - 1} < j \leq \delta_m} \ind[\text{$(j, \calx(j))$ is not in the same Sector that segment $m$ of $\caly$ is in} ]\\
    &\leq \frac{1}{n}[T^2 + \#(j \in [0,1] : \caly(j) = s_i \cdot j)]\\
    &\leq \frac{T^2 + M}{n}
\end{align*}

The $\frac{T^2}{n}$ term arises from the lattice points $(j, \calx(j))$ for $j \in [T^2]$, which have no guarantee of being in the Sector of the corresponding segment of $\caly$. The $\frac{\text{\# times $\caly$ crosses line $i$}}{n}$ term arises from the fact that for $j > T^2$, only $(j, \calx(j))$ which lie on line $i$ may potentially lie in a different Sector than that of the corresponding segment of $\caly$. This is upper bounded by $M$ since $\caly$ has only $M$ segments and so can cross line $i$ at most $M$ times. $\blacksquare$

%% file: sections/appendices/aux.tex
\subsection{Auxiliary Lemmas}\label{appen:aux}

\paragraph{Convex Geometry Lemmas.}

\begin{lemma}(Theorem 6.2 of \cite{rockafellar}) \label{lem:nonempty-ri}
    Let $C$ be a nonempty convex set. Then its relative interior, $\ri(C)$, is also nonempty.
\end{lemma}

\iffalse
\begin{lemma}(Caratheodory's Theorem; Prop. 1.15 of \cite{ziegler}, Theorem 17.1 of \cite{rockafellar})\label{lem:caratheodory}
    Let $V$ be a set in $\mathbb{R}^n$. Denote $\conv(V)$ as the convex hull of $V$. Then any point $p \in \conv(V)$ can be expressed as a convex combination of $\dim(V) + 1$ points in $V$.
\end{lemma}
\fi 

\begin{lemma}\label{lem:line-rel-int} (Theorem 6.1 of \cite{rockafellar})
    Let $C$ be a convex set in $\mathbb{R}^n$. Let $x \in \ri(C)$ and $z \in \clo(C)$. Then for all $0 \leq \lambda < 1$, $(1 - \lambda)x + \lambda z \in \ri(C)$.
\end{lemma}

\begin{lemma}\label{lem:open-hs-intersection}
    Let $C$ be a nonempty convex set in $\mathbb{R}^n$. Consider an open halfspace $H := \{x \in \mathbb{R}^n : \sum_{i \in [n]} \lambda_i x_i > \alpha\}$. If $C \cap H \neq \emptyset$, then $\dim(C \cap H) = \dim(C)$.
\end{lemma}

\textit{Proof.} Suppose that $p \in C \cap H$. If $p \in \ri(C)$, then since $H$ is an open set, we are done.

If not, $p \in \partial C := \clo(C) - \ri(C)$. By Lemma \ref{lem:nonempty-ri}, since $C \neq \emptyset$, there exists $x \in \ri(C)$. By Lemma \ref{lem:line-rel-int}, $\forall 0 \leq \lambda < 1$, $p_{\lambda}:= \lambda p + (1 - \lambda)x \in \ri(C)$. 

We claim there exists some setting of $\lambda \approx 1$ such that $p_{\lambda} \in H$. Since $H$ is open, there exists $\delta > 0$ such that $\textup{Ball}_{\delta}(p) := \{ x' \in \mathbb{R}^n : ||x' - p||_2 < \delta\} \subset H$. Taking $\lambda > 1 - \frac{\delta}{||x - p||_2}$ ensures that $||p_\lambda - p||_2 < \delta \implies p_{\lambda} \in H$. Thus, $p_{\lambda} \in H \cap \ri(C)$. $\blacksquare$

\paragraph{Lemmas Regarding Existence of Low Precision Element in $L_\infty$ Ball.}

\begin{lemma}\label{lem:interval-granularity}
    Let $N > 0$ be an integer. For any $x \in \mathbb{R}$, the interval $[x, x + \frac{1}{N}]$ contains at least one element of the form $\frac{m}{N}$ for integer $m$.
\end{lemma}

\textit{Proof.} For $N > 0$, the set $S = \{ \frac{m}{N} : m \in \mathbb{Z}\}$ is such that the minimum distance between any two consecutive elements is $\frac{1}{N}$. Since the length of the interval $[x, x + \frac{1}{N}]$ is $\frac{1}{N}$, it must contain some element of $S$, or else this would imply that there is some pair of consecutive elements in $S$ such that the distance between them is greater than $\frac{1}{N}$. $\blacksquare$

\begin{lemma}\label{lem:l-inf-ball-precision}
    Let $N > 0$ be an integer. For any $x = (x_1, \ldots, x_d) \in \mathbb{R}^d$ with $\sum_{i \in [d]}x_i = 1$, the $L_\infty$ ball $\{ x' \in \mathbb{R}^d : ||x - x'||_\infty \leq \frac{1}{N}\} \cap \{ x' : \sum_{i \in [d]} x_i' = 1\}$ contains a point $x'' \in \mathbb{Q}^d$ whose fractional coordinates have a least common denominator of at most $d\cdot N$. 
\end{lemma}

\textit{Proof.} Consider $\textup{Ball}^\infty_{\frac{1}{dN}}(x) :=  \{ x' \in \mathbb{R}^d : ||x - x'||_\infty \leq \frac{1}{dN}\} \cap \{ x' \in \mathbb{R}^d : \sum_{i \in [d]} x_i' = 1\}$. 

Along each dimension $i \in \{1, 2, \ldots, d - 1\}$, there must exist a number in the interval $[x_i - \frac{1}{dN}, x_i + \frac{1}{dN}]$ of the form $\frac{m_i}{dN}$ for integer $m_i$ by Lemma \ref{lem:interval-granularity}. We can set $x''_i = \frac{m_i}{dN}$ for all $i \in [d - 1]$. The last coordinate $x''_d$ will need to be set to $1 - \sum_{i \in [d - 1]} \frac{m_i}{dN}$ so that $x'' \in \{ x' : \sum_{i \in [d]} x_i' = 1\}$. Finally, $|x''_d - x_d| \leq \sum_{i \in [d - 1]} |x''_i - x_i| \leq \frac{d - 1}{dN} \leq \frac{1}{N}$. Hence, we have that $x'' \in \textup{Ball}^\infty_{\frac{1}{N}}(x)$ and that each coordinate is of the form of $\frac{m_i}{dN}$ for some integer $m_i$. $\blacksquare$

%--------- TODO: clean this part up--------

\paragraph{Relating to Technical Lemma \ref{lem:suffcond-for-asmpt}.}

\begin{lemma}\label{lem:suffcond-for-asmpt}
    If for every $f \in \crasp{}^{2}$ with $K(f)$ heads, we have that $z > 0$ and $\sum_{i \in [K(f)]}\lambda_i > z$, then for any $f, f' \in \crasp{}^2$ that are not equal, with $K$ and $K'$ heads respectively, then $f$ and $f'$ will differ on some continuous test-function, $\caly$, which strictly satisfies the second layer inequalities. That is, with $\max(K, K') \leq k \leq K + K'$ distinct heads between $f$ and $f'$, let $(B_1(\caly),\ldots, B_k(\caly))$ be the activations induced by $\caly$. Then, either:
\begin{align*}
    \sum_{i = 1}^{K} \lambda_i B_{\textup{ord}(1, i)}(\caly)  &> z\\
    \sum_{i = 1}^{K'} \lambda_i' B_{\textup{ord}(2, i)}(\caly)  &< z'
\end{align*}
Or,
\begin{align*}
    \sum_{i = 1}^{K} \lambda_i B_{\textup{ord}(1, i)}(\caly)  &< z\\
    \sum_{i = 1}^{K'} \lambda_i' B_{\textup{ord}(2, i)}(\caly)  &> z'
\end{align*}
\end{lemma}

To prove this, we will need Lemma \ref{lem:properties-of-A} and \ref{lem:connectedness-of-A}. First, define the following notion of connectedness for subsets of Euclidean space.

% Define connectedness as....; show path connectedness --> connectedness
\begin{definition} (Definition 2.4 of \cite{freiwald}) Two subsets $A$ and $B$ of a metric space $X$ are said to be separated if both $\clo(A) \cap B = \emptyset$ and $A \cap \clo(B) = \emptyset$. A set $E \subset X$ is connected if $E$ is not the union of two nonempty, separated sets.
\end{definition}

We will utilize the following Lemma (implicitly) in the proof below to show that a union of sets $\bigcup_{j \in [N]} A_j$ is connected, by iteratively arguing that for each $j \in [N] - \{ 1\}$, $A_j$ is not separated from $\bigcup_{k < j} A_j$ (i.e. they ``share an edge").

\begin{lemma}\label{lem:successive-increase-conn-set} (Theorem 2.9 in \cite{freiwald})
    Suppose $C$ and $\{C_\alpha\}_{\alpha \in I}$ are connected subsets of $X$ and that for each $\alpha,$ $C_\alpha$ and $C$ are not separated. Then  $C \cup \bigcup_{\alpha \in I} C_\alpha$ is connected
\end{lemma}

Next, fix $f, f'$ and the $k$ distinct slopes $\{ s_i\}_{i \in [k]}$ across the first layers of both programs. Let's consider the $k$-dimensional set $\act (\{ s_i\}_{i \in [k]}) \subset [0,1]^k$. $\act (\{ s_i\}_i)$ is equal to the union of a finite number of convex polytopes by Corollary \ref{cor:cleaner-basis-test-function} and Lemma \ref{lem:convexity-of-schema-activations}.

\begin{align*}
    \act (\{ s_i\}_{i \in [k]}) = \bigcup_{m \in [k], \text{valid } \{(y_1^i, y_2^{i})\}_{i \in [m]} \subset \{ 1, \ldots, k\}^m} A(Y_{\{(y_1^i, y_2^{i})\}_{i \in [m]}})
\end{align*} 

Here are two key Lemmas we will need. 

\begin{lemma}(Dimension of Polytopes)\label{lem:properties-of-A} Given a $(k,T)$-configuration $\{ s_i\}_{i \in [k]}$, then for each basis schema $Y_{\{(y_1^i, y_2^{i})\}_{i \in [m]}}$, the set of activations $A(Y_{\{(y_1^i, y_2^{i})\}_{i \in [m]}})$ has dimension $k$.
\end{lemma}

\iffalse
\begin{itemize}
    \item For each basis schema $Y_{\{(y_1^i, y_2^{i})\}_{i \in [m]}}$, the set of activations $A(Y_{\{(y_1^i, y_2^{i})\}_{i \in [m]}})$ has dimension $k$
    \item $\intr(\act (\{ s_i\}_{i \in [k]}))$ is connected.
\end{itemize}
\fi 

\textit{Proof.} Consider any basis schema  $Y$ that spans $k$ lines with $M'$ segments, as described in Corollary \ref{cor:cleaner-basis-test-function}. It suffices to show $k$ independent degrees of freedom in $A(Y)$.

Given $(B_1, \ldots B_k)$ consider the full-rank linear map, $Q : \mathbb{R}^k \to \mathbb{R}^k$

\begin{align*}
    Q(B_1, \ldots, B_k) = (B_1, B_2 - B_1, B_3 - B_2,\ldots, B_k - B_{k - 1})
\end{align*}

Denote $Q \circ A(Y)$ as the set consisting of points attained by applying $Q$ on all points in $A(Y)$. We have $\dim(Q \circ A(Y)) = \dim(A(Y)) - \dim(\ker(Q) \cap A(Y))$ so since $Q$ has rank $k$,  $\dim(\ker(Q) \cap A(Y)) \leq \dim(\ker(Q)) = 0$. So to prove that $\dim(A(Y)) = k$, it suffices to prove $\dim(Q \circ A(Y))$ is $k.$

By Lemma \ref{lem:convexity-of-schema-activations}, schema $Y$ will span all $k$ lines and will have at least one segment in each of the $k + 1$ sectors. First, set the lengths of all the segments that are in sector $k + 1$ to $0$. This can always be done because by Lemma \ref{lem:schema-constraints}, there is no constraint on any segment in $\textup{Sector}_{k + 1}$ since such a segment would not be crossing two distinct lines. Say that there are $M $ non-zero-length segments left that are each in the sectors $\{1, 2, \ldots, k\}$. Let $(n_1, \ldots, n_M)$ be the lengths of these segments. % M' - 1\leq 

There is a linear mapping $QL : \mathbb{R}^M \to \mathbb{R}^k$ that maps $(n_1, \ldots, n_M)$ to a vector $v \in \mathbb{R}^k$, where for all $i \in [k]$, $v_i$ is the sum of the length of segments that are contained in in $\textup{Sector}_i$. This map is the multiplication of $Q$ and $L$, where $L$ is a linear map $L : \mathbb{R}^M \to \mathbb{R}^k$ which maps segment lengths of $Y$, $(n_1, \ldots, n_M)$, to activations $(B_1, \ldots, B_k)$. In matrix form, $L \in \{ 0,1\}^{k \times M}$ is such that $L_{ij} = 1 \iff $ segment $j$ in schema $Y$ lies above line $i$ and hence contributes to activation $B_i$.

Now, for each sector $i \in [k]$, let:

\begin{align*}
    j_i := \max\{ j \in [M] : n_j \in \textup{Sector}_i\}
\end{align*}

be the index of the last segment in the schema that is in Sector $i$. 

Now, for each $i \in [k]$, consider a setting of $(n_1,\ldots, n_M)$ where for all $j < j_i$, $n_j = 0$, but $n_{j_i} > 0$. Denote the setting for $i \in [k]$ as $N_i := (n_1^{(i)},\ldots, n_M^{(i)})$. We claim that the $k$ vectors, $\{QL(N_i)\}_{i \in [k]}$ are linearly independent. Since $\{QL(N_i)\}_{i \in [k]} \subset Q \circ A(Y)$, this would imply the first point of the Lemma.  The claim that $\{QL(N_i)\}_{i \in [k]}$ are linearly independent can be argued by induction.

As a base case, suppose that $i_0 \in [k]$ is the sector where the $M$th segment of the schema lies in. Then $QL(N_{i_0}) = e_{i_0}$. The singleton set is linearly independent.

For each $1 \leq r \leq k$, denote $I_r:= \{ i \in [k] : j_i \text{ is one of the $r$-largest of the set } \{ j_i\}_{i \in [k]}\}$, so $I_1 = \{ i_0\}$. Suppose by the inductive hypothesis that for $1 \leq r \leq k$, the $r$ vectors $\{ QL(N_i) : i \in I_r\}$ are linearly independent. Now, let $i_*$ be the sector such that $j_{i_*}$ is the $(r + 1)$th largest of the set $\{ j_i\}_{i \in [k]}$. $(n_1^{(i_*)},\ldots, n_M^{(i_*)})$ will be such that $n_{j_{i_*}}^{(i_*)}$ is nonzero, which implies that the $i_*$th entry of $QL(N_{i_*})$ is nonzero. In contrast, for all $i \in I_r$, the $i_*$th entry of $QL(N_i)$ is zero, because for all $j > j_{i_*}$, there is no segment $n_j$ that lies in sector $i_*$, by definition of $j_{i_*}$, so by the construction of $N_i$ where all segments of index less than $j_i$ are set to 0, then the $i_*$th entry of $QL(N_i)$ is $0$ for all $i \in I_r$. Thus, the $r + 1$ vectors $\{ QL(N_i) : i \in I_{r + 1}\}$ are linearly independent. It follows that $I_k$ is a set of $k$ linearly independent vectors in $Q \circ A(Y)$, so $\dim(Q \circ A(Y)) = k$. $\blacksquare$

\iffalse
The second point follows from Lemma \ref{lem:connectedness-of-A} and the fact that 

\begin{align*}
    &\bigcup_{m \in [k], \text{valid } \{(y_1^i, y_2^{i})\}_{i \in [m]}} \intr(A(Y_{\{(y_1^i, y_2^{i})\}_{i \in [m]}})) \\
    &\subset \intr(\bigcup_{m \in [k], \text{valid } \{(y_1^i, y_2^{i})\}_{i \in [m]}} A(Y_{\{(y_1^i, y_2^{i})\}_{i \in [m]}}))
\end{align*}
Since the former, smaller set is connected, so is the latter, larger one. 
\fi

% Connectedness of A--------
\
\begin{lemma}(Connectedness of Interiors of Polytopes)\label{lem:connectedness-of-A}
    Given a $(k,T)$-configuration $\{ s_i\}_{i \in [k]}$, the set $\bigcup_{m \in [k], \text{valid }\{(y_1^i, y_2^{i})\}_{i \in [m]} \subset \{ 1, \ldots, k\}^m} \intr(A(Y_{\{(y_1^i, y_2^{i})\}_{i \in [m]}})) $ is connected.
\end{lemma}

\textit{Proof.} Let us first recall the set of basis schemas in Corollary \ref{cor:cleaner-basis-test-function}. For any $1 \leq m \leq k$ and $\{(y_1^i, y_2^{i})\}_{i \in [m]} \subset \{ 1, \ldots, k\}^m$ such that 

\begin{align*}
    y_1^m &= y_2^m\\
    \forall i \in [m - 1], y_1^i &\neq y_2^i\\
    \forall i \in [m - 2], y_1^i &< y_2^i \implies y_2^i = y_1^{i + 1} > y_2^{i + 1} > y_1^i\\
    \forall i \in [m - 2], y_1^i &> y_2^i \implies y_2^i = y_1^{i + 1} < y_2^{i + 1} < y_1^i\\
    (y_1^1, y_2^1) &= (1,k) \text{ or } (y_1^1, y_2^1) = (k,1)
\end{align*}
    
We can the test-function schema, $Y_{\{(y_1^i, y_2^{i})\}_{i \in [m]}}$ as the concatenation of $m - 1$ monotone curves, where the $i$th monotone curve goes from lines $y_1^i$ to $y_2^i$ for $i \in [m - 1]$. The set of these test-function schemas over all valid $\{(y_1^i, y_2^{i})\}_{i \in [m]}$ satisfying the above is complete by Corollary \ref{cor:cleaner-basis-test-function}.

Let $A(Y_{\{(y_1^i, y_2^{i})\}_{i \in [m]}})$ denote the set of activations induced by test-functions of schema $Y_{\{(y_1^i, y_2^{i})\}_{i \in [m]}}$. $A(Y_{\{(y_1^i, y_2^{i})\}_{i \in [m]}})$ is a convex set by Lemma \ref{lem:convexity-of-schema-activations}. Thus, each set $\intr(A(Y_{\{(y_1^i, y_2^{i})\}_{i \in [m]}}))$, by itself, is connected. 

Now, imagine a graph where each basis schema $Y_{\{(y_1^i, y_2^{i})\}_{i \in [m]}}$ is a node. Create an edge between basis schema $Y$ and $ Y'$ if $\intr(A(Y)) \cap \intr(A(Y')) \neq \emptyset$. By Lemma \ref{lem:successive-increase-conn-set}, to prove our desired conclusion, it is sufficient to show that this graph is connected. For this, we will need to prove which edges exist and that these edges suffice to connect the graph.

Categorize the basis schema into levels, where schema of the same levels have the same number of monotone curves (i.e. the parameter $m \in [k]$).

\textbf{Edges Within Same Level.} Consider any two basis test-function schemas, $Y_{\{(y_1^i, y_2^{i})\}_{i \in [m]}}$ and $Y_{\{((y_1^i)', (y_2^{i})')\}_{i \in [m]}}$, with $m$ monotone curves. Suppose the two schemas' $1$st through $(m - 2)$-th monotone curves are the same, but the $(m - 1)$th (i.e. their last) monotone curve's endpoint is different by 1 index. That is, % There are two cases: the second-to-last monotone curve goes up and the second-to-last monotone curve goes down. Suppose the $(m - 1)$th monotone curve goes up, from line $y_1^{m - 1} $ to line $ y_2^{m - 1} < y_1^{m - 1}$. 

\begin{align*}
    \forall i \in [m - 2], y_1^i &= (y_1^i)' \text{ and } y_2^i = (y_2^i)'\\
    y_1^{m - 1} &= (y_1^{m - 1})'\\
    y_2^{m - 1} &= (y_2^{m - 1})' - 1
\end{align*}

%Note that per the definition of these schema, $y_2^{m - 1} = y_1^m = y_2^m$ and $(y_2^{m - 1})' = (y_1^m)' = (y_2^m)'$. 

We claim that the set of activations of these two ``adjacent" schema must share an interior point:

\begin{align*}
    \intr(A(Y_{\{(y_1^i, y_2^{i})\}_{i \in [m]}})) \cap \intr(A(Y_{\{((y_1^i)', (y_2^{i})')\}_{i \in [m]}})) \neq \emptyset
\end{align*}

Suppose $Y_{\{(y_1^i, y_2^{i})\}_{i \in [m]}}$ has $M$ segments and $Y_{\{((y_1^i)', (y_2^{i})')\}_{i \in [m]}}$ has $M'$ segments\footnote{Note: the two schema have the same number of monotone curves, but different numbers of segments}. Because the two schema have the same monotone curves except the last monotone curve, then there are two possibilities: 

\begin{itemize}
    \item Either $y_1^{m - 1} < y_2^{m - 1} $ and $(y_1^{m - 1})' < (y_2^{m - 1})' $. One can interpret these monotone curves as ``moving downward", since the $(m - 1)$th monotone curve of $Y_{\{(y_1^i, y_2^{i})\}_{i \in [m]}}$ (resp. $Y_{\{((y_1^i)', (y_2^{i})')\}_{i \in [m]}}$ has $M'$) has start-point on line $y_1^{m - 1}$ (resp. $(y_1^{m - 1})'$) and end-point on line $y_2^{m - 1}$ (resp. $(y_2^{m - 1})' $). Note the lines with lower index have higher slope by convention (i.e. $s_1$ is largest).
    \item Or $y_1^{m - 1} > y_2^{m - 1} $ and $(y_1^{m - 1})' > (y_2^{m - 1})' $. One can interpret these monotone curves as ``moving upward".
\end{itemize} 

WLOG, consider just the first case (the second one utilizes an analogous argument). Then, since $y_1^{m - 1} < y_2^{m - 1}$, the last monotone curve is moving downward. Suppose WLOG that the monotone curve which starts at line $(y_1^{m - 1})'$ and ends at line $(y_2^{m - 1})'$ moves down one more line than the monotone curve which starts at line $y_1^{m - 1}$ and ends at line $y_2^{m - 1}$. 

Thus, $M' = M + 1$. Moreover, the set of segments which make up the two schema are similar: for all $j \in [M]$, segment $j$ of $Y_{\{(y_1^i, y_2^{i})\}_{i \in [m]}}$ lies in the same sector as segment $j$ of $Y_{\{((y_1^i)', (y_2^{i})')\}_{i \in [m]}}$. Meanwhile, segment $M + 1$ of $Y_{\{((y_1^i)', (y_2^{i})')\}_{i \in [m]}}$ has no counterpart in $Y_{\{(y_1^i, y_2^{i})\}_{i \in [m]}}$.

To show connectedness, we need to exhibit a point $(B_1,\ldots,B_k) \in [0,1]^k$ that is in the interior of  $A(Y_{\{(y_1^i, y_2^{i})\}_{i \in [m]}})$ and $A(Y_{\{((y_1^i)', (y_2^{i})')\}_{i \in [m]}})$. 

For schema $Y_{\{(y_1^i, y_2^{i})\}_{i \in [m]}}$ (resp. $Y_{\{((y_1^i)', (y_2^{i})')\}_{i \in [m]}}$), there is a linear transformation $L$ (resp. $L'$)  that maps the set of valid segment lengths $A^{(M)}(Y_{\{(y_1^i, y_2^{i})\}_{i \in [m]}}) \in [0,1]^M$ (resp. $A^{(M')}(Y_{\{((y_1^i)', (y_2^{i})')\}_{i \in [m]}}) \in [0,1]^{M'}$) of schema $Y_{\{(y_1^i, y_2^{i})\}_{i \in [m]}}$ (resp. $Y_{\{((y_1^i)', (y_2^{i})')\}_{i \in [m]}}$) to the set of activations $A(Y_{\{(y_1^i, y_2^{i})\}_{i \in [m]}}) \in [0,1]^k$ (resp. $A(Y_{\{((y_1^i)', (y_2^{i})')\}_{i \in [m]}}) \in [0,1]^{k}$). Being rank $k$ (see bullet point 1 of Lemma \ref{lem:properties-of-A}), $L$ (resp. $L'$) will map interior points in $A^{(M)}(Y_{\{(y_1^i, y_2^{i})\}_{i \in [m]}})$ (resp. $A^{(M')}(Y_{\{((y_1^i)', (y_2^{i})')\}_{i \in [m]}})$)  to interior points of $A(Y_{\{(y_1^i, y_2^{i})\}_{i \in [m]}})$ (resp. $A(Y_{\{((y_1^i)', (y_2^{i})')\}_{i \in [m]}})$). Thus, it suffices to exhibit two settings of the lengths of the segments of the two schema $(a_1,\ldots,a_M)$ and $(b_1,\ldots,b_{M'}) = (b_1,\ldots,b_M, b_{M + 1})$, that are interior points in their respective polytopes $A^{(M)}(Y_{\{(y_1^i, y_2^{i})\}_{i \in [m]}})$ (resp. $A^{(M')}(Y_{\{((y_1^i)', (y_2^{i})')\}_{i \in [m]}})$), that give rise to the same $(B_1,\ldots,B_k) \in [0,1]^k$.

Towards this goal, we are interested in the regime where $b_{M + 1} = \delta \ll 1$ and $\forall j \in [M], a_j \approx b_j$, and where each segment length has slack $\eps$ with respect to its constraint defined in Lemma \ref{lem:schema-constraints}. Suppose by Lemma \ref{lem:schema-constraints} that the constraints for the first $M$ segments for schema $Y_{\{((y_1^i)', (y_2^{i})')\}_{i \in [m]}}$ are:

\begin{align*}
    b_j \geq \frac{p_j}{q_j}\sum_{i = 1}^{j - 1} b_i, \forall j \in [M]
\end{align*}

We'll now describe how to set $\{ a_j\}_{j \in [M]}$ and $\{ b_j\}_{j \in [M']}$. We will largely ignore the normalization conditions that $\sum_{j \in [M]} a_j = 1$ and $\sum_{j \in [M']} b_j = 1$, but only require that the total length is the same: $\sum_{j \in [M]} a_j = \sum_{j \in [M']}b_j > 0$. % The fact that the sum is not $1$ will just rescale the resulting $(B_1, \ldots, B_k)$ by some constant factor, but won't affect whether it is an interior point. 

% normalizing at the end will not affect whether the segment lengths we describe constitute an interior point.

For $\eps, \delta > 0$, set $b_1 = 1$ and for $j \in \{ 2, \ldots, M\}$, sequentially assign $b_j = \frac{p_j}{q_j}\sum_{i = 1}^{j - 1} b_i + \eps$. Set $b_{M + 1} = \delta$ (where the only constraint for $b_{M + 1}$ is that it is nonnegative, being the last segment). $\{ b_j\}_{j \in [M + 1]}$ is a valid set of segment lengths respecting the constraints imposed by schema $Y_{\{((y_1^i)', (y_2^{i})')\}_{i \in [m]}}$ as per Lemma \ref{lem:schema-constraints}. $\{ b_j\}_{j \in [M + 1]}$ is also an interior point of $A^{(M')}(Y_{\{((y_1^i)', (y_2^{i})')\}_{i \in [m]}})$, since each constraint has slack $\eps > 0$ or $\delta > 0$.

Now, suppose that the $(M + 1)$th segment of $Y_{\{((y_1^i)', (y_2^{i})')\}_{i \in [m]}}$ lies in sector $i_*$. Because each basis schema spans all $k$ lines, then there exists some $j_* \in [M]$ such that segment $j_*$ in $Y_{\{(y_1^i, y_2^{i})\}_{i \in [m]}}$ is also in sector $i_*$. Set $\{ a_i\}$ as follows.

\begin{align*}
    \forall j \in [M] - \{ j_*\}, \text{ set } a_j = b_j \\
    \text{Finally, set } a_{j_*} = b_{j_*} + \delta 
\end{align*}

We now need to argue this is a valid assignment of $\{ a_i\}$, respecting the constraints on segments imposed by the schema $Y_{\{(y_1^i, y_2^{i})\}_{i \in [m]}}$. Because the first $M$ segments of $Y_{\{((y_1^i)', (y_2^{i})')\}_{i \in [m]}}$ are the same as the  $M$ segments of $Y_{\{(y_1^i, y_2^{i})\}_{i \in [m]}}$, then these segments have the same constraints.

It follows that for all $j \leq j_*$, $a_j$ meets its constraint. For subsequent segments, due to the  extra $\delta$ contribution on $a_{j_*}$, we will need to argue that their constraint is still met with some nonzero slack. The $t$th constraint after $j^*$ will have its slack lessened by at most $\delta \cdot [\max_{p,q \in \mathbb{Z}, |p|, |q| \leq T} (\frac{p}{q})^t] \leq \delta T^{M } \leq \delta T^{k^2}$, where we used that $t \leq M \leq k^2$ by Corollary \ref{cor:size-of-M}.

Thus, as long as $\delta < \frac{\eps}{T^{k^2}}$, then this setting of $\{ a_j\}_{j \in [M]}$ will still satisfy each constraint of $Y_{\{(y_1^i, y_2^{i})\}_{i \in [m]}}$ with nonzero slack. 

\begin{align*}
    a_j \geq \frac{p_j}{q_j}\sum_{i = 1}^{j - 1} a_i, \forall j \in [M]
\end{align*}

Thus, $\{ a_j\}_{j \in [M]}$ is an interior point of $A^{(M)}(Y_{\{(y_1^i, y_2^{i})\}_{i \in [m]}})$. 

Finally, the segments $\{ a_j\}_{j \in [M]}$ and $\{ b_j\}_{j \in [M + 1]}$ induce the same $\{ B_i\}_{i \in [k]}$ since the sum of the segment lengths in each sector is the same. This proves that:

\begin{align*}
    \intr(A(Y_{\{(y_1^i, y_2^{i})\}_{i \in [m]}})) \cap \intr(A(Y_{\{((y_1^i)', (y_2^{i})')\}_{i \in [m]}})) \neq \emptyset
\end{align*}

\textbf{Edges Between Levels $L$ and $L + 1$.}  Consider any two schema $Y_{\{(y_1^i, y_2^{i})\}_{i \in [m - 1]}}, Y_{\{((y_1^i)', (y_2^{i})')\}_{i \in [m]}}$ with $m - 2$ and $m - 1$ monotone curves, respectively, where $Y_{\{((y_1^i)', (y_2^{i})')\}_{i \in [m]}}$'s first $m - 2$ monotone curves are the same as $Y_{\{(y_1^i, y_2^{i})\}_{i \in [m - 1]}}$'s $m - 2$ monotone curves. Moreover, let schema $Y_{\{((y_1^i)', (y_2^{i})')\}_{i \in [m]}}$ be such that:

\begin{itemize}
    \item In the case that $(y_1^{m - 1})' < (y_2^{m - 1})'$, then $(y_1^{m - 1})' + 1 =  (y_2^{m - 1})'$.
    \item In the case that $(y_1^{m - 1})' > (y_2^{m - 1})'$, then $(y_1^{m - 1})' - 1 =  (y_2^{m - 1})'$
\end{itemize}

Instead of a schema of $m - 2$ monotone curves, schema $Y_{\{(y_1^i, y_2^{i})\}_{i \in [m]}}$ can also be thought of as a schema of $m - 1$ monotone curves, where the last monotone curves will only cross the line with index $y_1^{m - 2} = y_2^{m - 2}$. Thus, these two schema can be thought of as having their final $(m-1)$-th monotone curves' endpoints differ by one line, which reduces to the case above with two schema in the same level. Using exactly the same argument as in the previous section with two schema in the same level, we can conclude that

\begin{align*}
    \intr(A(Y_{\{(y_1^i, y_2^{i})\}_{i \in [m - 1]}})) \cap \intr(A(Y_{\{((y_1^i)', (y_2^{i})')\}_{i \in [m]}})) \neq \emptyset
\end{align*}

\textbf{Edges Between Two Connected Components.} In our graph where basis schema are nodes, we now have two connected components: 
\begin{itemize}
    \item Schemas where the first monotone curve is a monotone (up) curve from lines $k$ to $1$
    \item Schemas where the first monotone curve is a monotone (down) curve from lines $1$ to $k$.
\end{itemize}

Within each of these connected components, we have proved connectedness with the previous two cases, by connecting schemas of the same prefix (of monotone curves) and the same level, and schemas of the same prefix and different levels. 

The nodes in the first connected component will all connect in a tree to the schema consisting of a single monotone curve from lines $k$ to $1$.  The nodes in the second connected component will all connect in a tree to the schema consisting of a single monotone curve from lines $1$ to $k$.

Let $Y$ be the schema for $m = 2$ with a single monotone (up) curve from line $k$ to $1$, and $Z$ be the schema for $m = 3$ with two monotone curves: one monotone (down) curve from line $1$ to $k$ and then a monotone (up) curve from $k$ to $2$. These are two representatives from the two aforementioned connected components. We claim that:

\begin{align*}
    \intr(A(Y)) \cap \intr(A(Z)) \neq \emptyset
\end{align*}

We will start with an interior point for $Z$. Suppose it has segment lengths $z_1,\ldots, z_{k + 1}, \ldots,z_{2k}$, where $z_1$ is above line 1, $z_{k + 1}$ is below line $k$, and segment $z_{2k}$ lies above line 2 and below line 1, with end-point on line 2. 

Again, ignore the normalization condition that $\sum_{i \in [2k]} z_i = 1$. For $\eps > 0$, we assign values to the segments as follows.

\begin{align*}
    z_1 &= 1\\
    \forall 2 \leq i \leq 2k - 1, z_i &= \frac{p_i}{q_i} \sum_{j = 1}^{i - 1}z_j + \eps
\end{align*}

Note that since $z_{2k}$ is the last segment of the schema, it need not cross a sector and need only cross and re-cross line 2. Therefore, it has no constraints under schema $Z$ except that $z_{2k} \geq 0$. However, the key idea is that we can choose for $z_{2k}$ to be sufficiently large so as to satisfy the constraint needed to cross to line 1 from line 2.

\begin{align*}
    z_{2k} &= \frac{p_{2k}}{q_{2k}} \sum_{j = 1}^{2k - 1}z_j + \eps\\
\end{align*}

Such a $\{ z_i\}_{i \in [2k]}$, after normalization, exists in the (relative) interior of $A^{(M)}(Z)$.

We claim that rearranging the segments $\{ z_i\}_{i \in [2k]}$ into a monotone (up) curve $(y_1,\ldots, y_{k + 1})$ with slack $\eps$ will always be possible for any $\eps > 0$. This essentially follows from Lemma \ref{lem:monotoneline} as we have intentionally set $z_{2k} = \frac{p_{2k}}{q_{2k}} \sum_{j = 1}^{2k - 1}z_j + \eps$ large enough to cross from line 2 to line 1 (and have its end-point on line 1). More precisely, the following constraints are met by $\{ z_i\}_{i \in [2k]}$.

\begin{align*}
    \forall k + 2 \leq i \leq 2k, z_i &= \frac{p_i}{q_i} \sum_{j = 1}^{i - 1}z_j + \eps\\
    &\geq \frac{p_i}{q_i} \sum_{j \leq i - 1 : z_j \text{ in sector } \geq (2k +3- i)} z_j + \eps
\end{align*}

Then, we obtain a valid monotone curve with segment lengths $(y_1, \ldots, y_{k + 1})$ by setting:

\begin{align*}
    y_1 &= z_{k + 1}\\
    y_{k + 1} &= z_{1}\\
    y_i &= z_{i + k } + z_{k + 2 - i}, 2 \leq i \leq k
\end{align*}

Thus, the large value of $z_{2k}$ enables the re-arranged test-function of monotone schema to  cross from line 2 to line 1. Finally, there is no constraint on $y_{k  + 1}$, so all constraints of the schema $Y$ are satisfied. This re-arrangement will have slack $\min(\eps, y_1, y_{k + 1}) \geq \eps$ because each of the original constraints in $Z$ on the segments $z_1, \ldots, z_{2k - 1}$ as well as the ``pseudo-constraint" we imposed on $z_{2k}$ only got looser in this rearrangement, and there is no constraint on $y_{k + 1}$. Thus, the rearranged monotone curve, after normalization of the segment lengths, is in the interior of $A^{(M)}(Y)$. 

$\{ y_i\}$ and $\{ z_i\}$ are both interior in their respective polytopes and induce the same activations $(B_1,\ldots,B_k)$ since the segments of $\{ y_i\}$ are just a re-arrangement of the segments of $\{ z_i\}$, and $\sum_{i \in [2k]}z_i = \sum_{i \in [k + 1]}y_i$. Thus, they give rise to the same interior point $(B_1,\ldots,B_k) \in \intr (A(Y)), (B_1,\ldots,B_k) \in \intr (A(Z))$, so $ \intr (A(Y)) \cap  \intr (A(Z)) \neq \emptyset$.

These edges demonstrate that the graph is connected, proving the Lemma. $\blacksquare$

%--------------------------------------------------------------------
\textbf{Proof of Lemma \ref{lem:suffcond-for-asmpt}. }

\begin{lemma} (Lemma \ref{lem:suffcond-for-asmpt}, restated)
    If for every $f \in \crasp{}^{2}$ with $K(f)$ heads, we have that $z > 0$ and $\sum_{i \in [K(f)]}\lambda_i > z$, then for any $f, f' \in \crasp{}^2$ that are not equal, with $K$ and $K'$ heads respectively, then $f$ and $f'$ will differ on some continuous test-function, $\caly$, which strictly satisfies the second layer inequalities. That is, with $\max(K, K') \leq k \leq K + K'$ distinct heads between $f$ and $f'$, let $(B_1(\caly),\ldots, B_k(\caly))$ be the activations induced by $\caly$. Then, either:
\begin{align*}
    \sum_{i = 1}^{K} \lambda_i B_{\textup{ord}(1, i)}(\caly)  &> z\\
    \sum_{i = 1}^{K'} \lambda_i' B_{\textup{ord}(2, i)}(\caly)  &< z'
\end{align*}
Or,
\begin{align*}
    \sum_{i = 1}^{K} \lambda_i B_{\textup{ord}(1, i)}(\caly)  &< z\\
    \sum_{i = 1}^{K'} \lambda_i' B_{\textup{ord}(2, i)}(\caly)  &> z'
\end{align*}
\end{lemma}

\textit{Proof of Lemma \ref{lem:suffcond-for-asmpt}. } Fix $f, f'$ and their $k$ distinct slopes $\{ s_i\}_{i \in [k]}.$ Suppose $f, f'$ differ on some discrete test-function $\calx$. We want to show that there must be a \textit{continuous} test-function that strictly satisfy the two inequalities strictly and distinguishes $f$ and $f'$.

 Let's consider the $k$-dimensional set $\act (\{ s_i\}_i) \subset [0,1]^k$. $\act (\{ s_i\}_i)$ is equal to the union of a finite number of convex sets by Corollary \ref{cor:cleaner-basis-test-function} and Lemma \ref{lem:convexity-of-schema-activations}.

\begin{align*}
    \act (\{ s_i\}_{i \in [k]}) = \bigcup_{m \in [k], \text{valid } \{(y_1^i, y_2^{i})\}_{i \in [m]} \subset \{ 1, \ldots, k\}^m} A(Y_{\{(y_1^i, y_2^{i})\}_{i \in [m]}})
\end{align*}

Now consider two halfspaces $H_1, H_2$, induced by the second layers of $f$ and $f'$.

\begin{align*}
    H_1 &:= \{ B \in \mathbb{R}^k : \sum_{i = 1}^{K} \lambda_i B_{\textup{ord}(1,i)}  > z\} \\
    H_2 &:= \{ B \in \mathbb{R}^k : \sum_{i = 1}^{K'} \lambda_i' B_{\textup{ord}(2,i)}  > z'\}
\end{align*}

First, we cannot have that $H_1 = H_2$, as then $f$ and $ f'$ must share the same heads in their first layer and also have the same coefficients in the second layer. Thus, $f = f'$, contradicting the original assumption that $f$ and $f'$ are not equal (as witnessed  by $\calx$).

Thus, $H_1 \neq H_2$. Denote 

\begin{align*}
    L_1 &:= \{ B\in \mathbb{R}^k: \sum_{i = 1}^{K} \lambda_i B_{\textup{ord}(1,i)}  = z \} \\
    L_2 &:= \{ B\in \mathbb{R}^k: \sum_{i = 1}^{K'} \lambda_i' B_{\textup{ord}(2,i)}  = z' \}
\end{align*}

$H_1 \neq H_2$ implies $L_1 \neq L_2$. Now, there are two cases, given by whether $L_1$ and $ L_2$ have a nonzero intersection or not.

\textbf{Case 1: $L_1 \cap L_2 \neq \emptyset$}

$H_1, H_2$ divide $\mathbb{R}^k$ into 4 quadrants. Denote the $(+, +)$ quadrant as $H_1 \cap H_2$, $(+, -)$ quadrant as $H_1 \cap H_2^c$, $(-,+)$ quadrant as $H_1^c \cap H_2$ and $(-,-)$ quadrant as $H_1^c \cap H_2^c$.

Towards contradiction, let us consider all possible $ \act (\{ s_i\}_i)$ such that the following Condition I holds.

\begin{itemize}

\item (Condition I) there is no $B \in \act (\{ s_i\}_i)$ in the $(-, +)$ and $(+, -)$  quadrants such that:

\begin{align*}
    \sum_{i = 1}^{K} \lambda_i B_{\textup{ord}(1,i)}  > z\\
    \sum_{i = 1}^{K'} \lambda_i' B_{\textup{ord}(2,i)}  < z'\\
    \text{ or }\\
    \sum_{i = 1}^{K} \lambda_i B_{\textup{ord}(1,i)}  < z\\
    \sum_{i = 1}^{K'} \lambda_i' B_{\textup{ord}(2,i)}  > z'
\end{align*}
\end{itemize}

For ease of notation, index each basis schema of Corollary \ref{cor:cleaner-basis-test-function} with $j \in [N_k]$, where $N_k$ ($= 2^{k - 1}$) is the total number of basis schema, so that $\act (\{ s_i\}_i) = \bigcup_{j = 1}^{N_k} A_j$. Thus, Condition I must also apply to each $A_j, \forall j \in [N_k]$. % In addition, at least one $A_j$ satisfies Condition 1. 

In order that $A_j$ satisfies Condition I, there is only three convex possibilities:

\begin{enumerate}
    \item (Type I) $A_j$ is a convex subset of a $(k - 1)$-dimensional plane that has nonzero intersection with the $(k - 2)$-dimensional intersection of $O = L_1 \cap L_2$ and that is a subset of 
    \begin{align*}
        \big[\clo(H_1) \cap \clo(H_2)\big] \cup \big[H_1^c \cap H_2^c\big]
    \end{align*}
    \item (Type II) $A_j$ is a convex subset of the quadrant $\clo(H_1) \cap \clo(H_2)$, where:
    \begin{align*}
        \sum_{i = 1}^{K} \lambda_i B_{\textup{ord}(1,i)}  \geq z\\
    \sum_{i = 1}^{K'} \lambda_i' B_{\textup{ord}(2,i)}  \geq z'
    \end{align*}
    \item (Type III) $A_j$ is a convex subset of the quadrant, $H_1^c \cap H_2^c$, where:
    \begin{align*}
        \sum_{i = 1}^{K} \lambda_i B_{\textup{ord}(1,i)}  \leq z\\
    \sum_{i = 1}^{K'} \lambda_i' B_{\textup{ord}(2,i)}  \leq z'
    \end{align*}
\end{enumerate}

Lemma \ref{lem:properties-of-A} implies that it is impossible for any $A_j$ to be Type I since the dimension of Type I sets is $k - 1$ but $A_j$ is dimension $k$.

Lemma \ref{lem:connectedness-of-A} implies that either all  $A_j$ are type II or all $A_j$ are type III, which can be seen as follows. Suppose towards contradiction that  there was an $A_j$ of type II and a separate $A_{j'}$ of type III. First, no point $O = L_1 \cap L_2 $  can be an interior point of any $A_j$, because $O$ is a $k - 2$ dimensional subspace, and it is impossible for an $L_2$-ball centered at any point in $O$ of any nonzero radius and dimension $k$ to be completely contained in $[\clo(H_1) \cap \clo(H_2)] \cup [H_1^c \cap H_2^c] \supset \bigcup_{j = 1}^{N_k} A_j$, as such a ball would always have a non-empty intersection with $H_1 \cap \intr(H_2^c)$ and $H_2 \cap \intr(H_1^c)$. Thus, $O$ is disjoint from $\bigcup_{j \in [N_k]}\intr(A_j)$. Because there is an $A_j$ of type II and a separate $A_{j'}$ of type III, part of $\bigcup_{j = 1}^{N_k} \intr(A_j)$ is contained in $\clo(H_1) \cap \clo(H_2)$ and part of it is contained in $H_1^c \cap H_2^c$. However, since no point in $O$ can be in $\bigcup_{j = 1}^{N_k} \intr(A_j)$, then $\bigcup_{j = 1}^{N_k} \intr(A_j)$ is not connected (i.e. it is the union of two non-empty, separated sets). This contradicts Lemma \ref{lem:connectedness-of-A}. Thus, all $A_j$ must be type II or they must all be type III. % Because we know that at least one $A_{j^*}$ satisfies Condition 1, as it contains $B^*$ in the quadrant, $[\clo(H_1) \cap \clo(H_2)]$, then $A_{j^*}$ is Type II. We conclude that all $A_j$ must be type II.

Finally, observe that $(0,\ldots,0) \in \act (\{ s_i\}_i) = \bigcup_{j = 1}^{N_k} A_j$, realized by a test-function which has slope $0$ everywhere. Also, $(1,\ldots,1) \in \act (\{ s_i\}_i) = \bigcup_{j = 1}^{N_k} A_j$, realized by a test-function which has slope $1$ everywhere.

Suppose all $A_j$ were type II. At least one $A_j$ must contain $(0,\ldots, 0)$, but if $(0,\ldots,0)$ is in the quadrant, $\clo(H_1) \cap \clo(H_2)$, this implies:

\begin{align*}
0 = \sum_{i = 1}^{K'} (\lambda_i' \cdot 0) &\geq z'\\
0 = \sum_{i = 1}^{K} (\lambda_i \cdot 0) &\geq z
\end{align*}

This contradicts the assumption that $z,z' > 0$.

Now, suppose all $A_j$ were type III. At least one $A_j$ must contain $(1,\ldots, 1)$, but if $(1,\ldots,1)$ is in the quadrant, $H_1^c \cap H_2^c$, this implies:

\begin{align*}
\sum_{i = 1}^{K'} (\lambda_i' \cdot 1) &\leq z'\\
\sum_{i = 1}^{K} (\lambda_i \cdot 1) &\leq z
\end{align*}

This contradicts the assumption that $\sum_{i = 1}^{K} \lambda_i > z$ and $\sum_{i = 1}^{K'} \lambda_i' > z'$.

In short, $z > 0$ and $\sum_{i \in [K(f)]}\lambda_i > z$ for all functions in $f \in \crasp^{2}$ $\implies$ $\act (\{ s_i\}_i)$ does not satisfy Condition I. 

Thus, there exists $B \in \act (\{ s_i\}_i)$, realized by a continuous test-function, in the $(-, +)$ and $(+, -)$  quadrants such that:

\begin{align*}
    \sum_{i = 1}^{K} \lambda_i B_{\textup{ord}(1,i)}  > z\\
    \sum_{i = 1}^{K'} \lambda_i' B_{\textup{ord}(2,i)}  < z'\\
    \text{ or }\\
    \sum_{i = 1}^{K} \lambda_i B_{\textup{ord}(1,i)}  < z\\
    \sum_{i = 1}^{K'} \lambda_i' B_{\textup{ord}(2,i)}  > z'
\end{align*}

\textbf{Case 2: $L_1 \cap L_2 = \emptyset$}. 

The argument here is similar. We know $L_1, L_2$ are parallel hyperplanes which never intersect. We will argue that it is still true that $\{A_j\}_{j \in [N_k]}$ are either all type II or all type III, and then reach a contradiction with the assumption that $z > 0$ and $\sum_{i \in [K(f)]}\lambda_i > z$ for all functions in $\crasp^{2}$. Given that $L_1, L_2$ are parallel, consider two subcases. 

Let $(+, -), (+, -)$ denote the subcase where halfspace $H_1, H_2$ have the same parity, but their intercept term is different. An example of this are two halfspaces: ``$x + y > 2, x + y > 1$".

Let  $(-, +), (+, -)$ be the subcase where $H_1, H_2$ have opposite parities. An example of this are the two halfspaces:``$x + y < 2, x + y > 1$".

\textbf{Subcase 1: $(+, -), (+, -)$} In this case, in order that the planes are not identical, there is a nonzero difference in the intercept term between $L_1, L_2$. By Lemma \ref{lem:connectedness-of-A}, since $\bigcup_{j \in [N_k]}\intr(A_j)$ is connected, then $\{A_j\}_{j \in [N_k]}$ are either all type II or all type III, via a similar argument as Case 1, which contradicts that $z, z' > 0$, and $\sum_{i = 1}^{K} \lambda_i > z$ and $\sum_{i = 1}^{K'} \lambda_i' > z'$ respectively. Thus, Condition I cannot hold.

\textbf{Subcase 2: $(+, -), (-, +)$} In this case, either $\clo(H_1) \cap \clo(H_2) \neq \emptyset$ or $\intr(H_1^c) \cap \intr(H_2^c) \neq \emptyset$. Once again, $\{A_j\}_{j \in [N_k]}$ are either all type II or all type III, contradicting that $z, z' > 0$, and $\sum_{i = 1}^{K} \lambda_i > z$ and $\sum_{i = 1}^{K'} \lambda_i' > z'$ respectively. Thus, Condition I cannot hold. $\blacksquare$

%--------- END--------

%---------------A_{23} Lemmas-------------

Here is a Corollary of Lemma \ref{lem:connectedness-of-A} which will be useful in strengthening Lemma \ref{lem:suffcond-for-asmpt}, which will let us improve the final bound from $O(T^{O(K^2)})$ ($\alpha^{O(\log \alpha)}$) to $O(T^{O(K)})$ ($\alpha^{O(1)}$).

\begin{corollary}\label{cor:connectedness-of-A23}  (Connectedness of $\intr(\act_{2,3}(\{ s_i\}_{i \in [k]}))$) Define $\act_{2,3}(\{ s_i\}_{i \in [k]})$ as the set of activations by basis schema where $m \in \{ 2,3\}$. That is, there is only either one or two monotone curves in the schema.
\begin{align*}
    \act_{2,3}(\{ s_i\}_{i \in [k]}) := \bigcup_{m \in \{ 2,3\}, \text{valid } \{(y_1^i, y_2^{i})\}_{i \in [m]} \subset \{ 1, \ldots, k\}^m} A(Y_{\{(y_1^i, y_2^{i})\}_{i \in [m]}})
\end{align*}
Then, we claim that $\bigcup_{m \in \{ 2,3\}, \text{valid } \{(y_1^i, y_2^{i})\}_{i \in [m]} \subset \{ 1, \ldots, k\}^m} \intr(A(Y_{\{(y_1^i, y_2^{i})\}_{i \in [m]}}))$ is connected.
\end{corollary}

\textit{Proof.} We have that $\act_{2,3}(\{ s_i\}_{i \in [k]}) := A(Y_{\{ (1, k), (k,k)\}}) \cup \bigcup_{j \in \{2, \ldots, k - 1\}} A(Y_{\{ (1, k), (k, j), (j,j)\}}) \cup A(Y_{\{ (k, 1), (1,1)\}}) \cup \bigcup_{j \in \{2, \ldots, k - 1\}} A(Y_{\{ (k, 1), (1, j), (j,j)\}}) $. The connectedness of $\intr(A(Y_{\{ (1, k), (k,k)\}})) \cup \bigcup_{j \in \{2, \ldots, k - 1\}} \intr(A(Y_{\{ (1, k), (k, j), (j,j)\}})) \cup \intr(A(Y_{\{ (k, 1), (1,1)\}})) \cup \bigcup_{j \in \{2, \ldots, k - 1\}} \intr(A(Y_{\{ (k, 1), (1, j), (j,j)\}}))$ follows from inspecting the edges in the graph construction in Lemma \ref{lem:connectedness-of-A}. 

In the sections where we analyzed edges between nodes in the same level and between consecutive levels, it follows that:

\begin{align*}
    \forall j \in \{2, \ldots, k - 1\}, \text{ node } Y_{\{ (1, k), (k,k)\}} \text{ is connected to } \text{ node } Y_{\{ (1, k), (k, j), (j,j)\}}
\end{align*}

And

\begin{align*}
    \forall j \in \{2, \ldots, k - 1\}, \text{ node } Y_{\{ (k, 1), (1,1)\}} \text{ is connected to } \text{ node } Y_{\{ (k, 1), (1, j), (j,j)\}}
\end{align*}

In the final section of the proof of Lemma \ref{lem:connectedness-of-A}, we showed that

\begin{align*}
    \text{ node } Y_{\{ (k, 1), (1,1)\}} \text{ is connected to } \text{ node } Y_{\{ (1, k), (k, 2), (2,2)\}}
\end{align*}

$\blacksquare$

\iffalse 
It follows that:

\begin{align}
    \intr(\act_{2,3}(\{ s_i\}_{i \in [k]})) &\supset \\
    \intr(A(Y_{\{ (1, k), (k,k)\}})) &\cup \bigcup_{j \in \{2, \ldots, k - 1\}} \intr(A(Y_{\{ (1, k), (k, j), (j,j)\}})) \\
    \cup \intr(Y_{\{ (k, 1), (1,1)\}}) &\cup \bigcup_{j \in \{2, \ldots, k - 1\}} \intr(A(Y_{\{ (k, 1), (1, j), (j,j)\}}) )\label{eq:intr-of-a23}
\end{align}

The latter is a connected set because (1) we demonstrated an edge exists between the nodes corresponding to each of the schema of $2$ or $3$ monotone curves, so that the entire graph is connected like a tree. Recall that we defined an edge between the nodes for two schema $Y, Z$ if $\intr(A(Y)) \cap \intr(A(Z)) \neq \emptyset$. (2) Each set $A(Y_{\{(y_1^i, y_2^{i})\}_{i \in [m]}})$ is a convex poytope by Lemma \ref{lem:convexity-of-schema-activations}, so the interior of each nodes' set of activations $\intr(A(Y_{\{(y_1^i, y_2^{i})\}_{i \in [m]}}))$ is connected within itself. 

It follows that $\intr(\act_{2,3}(\{ s_i\}_{i \in [k]}))$, which is larger than the set on the RHS of Equation \ref{eq:intr-of-a23}, is connected. 
\fi 

As a corollary, we can strengthen Lemma \ref{lem:suffcond-for-asmpt} to the following.

\begin{lemma}(Stronger version of Lemma \ref{lem:suffcond-for-asmpt})\label{lem:stronger-suffcond-for-asmpt}
    If for every $f \in \crasp{}^{2}$ with $K(f)$ heads, we have that $z > 0$ and $\sum_{i \in [K(f)]}\lambda_i > z$, then for any $f, f' \in \crasp{}^2$ that are not equal, with $K$ and $K'$ heads respectively, then $f$ and $f'$ will differ on some continuous test-function, $\caly$, which strictly satisfies the second layer inequalities. That is, with $\max(K, K') \leq k \leq K + K'$ distinct heads between $f$ and $f'$, let $(B_1(\caly),\ldots, B_k(\caly))$ be the activations induced by $\caly$. Then, either:
\begin{align*}
    \sum_{i = 1}^{K} \lambda_i B_{\textup{ord}(1, i)}(\caly)  &> z\\
    \sum_{i = 1}^{K'} \lambda_i' B_{\textup{ord}(2, i)}(\caly)  &< z'
\end{align*}
Or,
\begin{align*}
    \sum_{i = 1}^{K} \lambda_i B_{\textup{ord}(1, i)}(\caly)  &< z\\
    \sum_{i = 1}^{K'} \lambda_i' B_{\textup{ord}(2, i)}(\caly)  &> z'
\end{align*}
In addition, $\caly$ will be of a basis schema $Y$ with either one or two monotone curves: $Y \in \{ Y_{\{(y_1^i, y_2^{i})\}_{i \in [m]}} : m \in \{ 2,3\}, \text{valid } \{(y_1^i, y_2^{i})\}_{i \in [m]} \subset \{ 1, \ldots, k\}^m\}$. In particular, the number of segments $M$ in the schema of $\caly$ will be at most $2 \cdot k$ instead of $k^2$ for general basis schema.
\end{lemma}

\textit{Proof.} We repeat the proof of Lemma \ref{lem:suffcond-for-asmpt}, but with $\intr(\act_{2,3}(\{ s_i\}_{i \in [k]}))$ substituted for $\act(\{ s_i\}_{i \in [k]})$ and Corollary \ref{cor:connectedness-of-A23} substituted for Lemma \ref{lem:connectedness-of-A}. Note that  $\intr(\act_{2,3}(\{ s_i\}_{i \in [k]}))$ still contains the points $(0,\ldots, 0) \in [0,1]^k$ and $(1, \ldots, 1) \in [0,1]^k$ as well, as both of these are realized by a test-function of where the slope everywhere is equal to $0$ everywhere and $1$ everywhere, respectively, which are both of schemas with just one monotone curve. $\blacksquare$

%% file: main-arxiv.bbl
\begin{thebibliography}{27}
\providecommand{\natexlab}[1]{#1}
\providecommand{\url}[1]{\texttt{#1}}
\expandafter\ifx\csname urlstyle\endcsname\relax
  \providecommand{\doi}[1]{doi: #1}\else
  \providecommand{\doi}{doi: \begingroup \urlstyle{rm}\Url}\fi

\bibitem[Abbe et~al.(2024)Abbe, Bengio, Lotfi, and Rizk]{abbe2024generalizationunseenlogicreasoning}
Abbe, E., Bengio, S., Lotfi, A., and Rizk, K.
\newblock Generalization on the unseen, logic reasoning and degree curriculum, 2024.
\newblock URL \url{https://proceedings.mlr.press/v202/abbe23a/abbe23a.pdf}.

\bibitem[Anil et~al.(2022)Anil, Wu, Andreassen, Lewkowycz, Misra, Ramasesh, Slone, Gur-Ari, Dyer, and Neyshabur]{anil2022exploringlengthgeneralizationlarge}
Anil, C., Wu, Y., Andreassen, A., Lewkowycz, A., Misra, V., Ramasesh, V., Slone, A., Gur-Ari, G., Dyer, E., and Neyshabur, B.
\newblock Exploring length generalization in large language models, 2022.
\newblock URL \url{https://openreview.net/forum?id=zSkYVeX7bC4}.

\bibitem[Baker \& Book(1974)Baker and Book]{baker-book-linear-cfg}
Baker, B.~S. and Book, R.~V.
\newblock Reversal-bounded multipushdown machines, 1974.
\newblock URL \url{https://doi.org/10.1016/S0022-0000(74)80027-9}.

\bibitem[Brown et~al.(2020)Brown, Mann, Ryder, Subbiah, Kaplan, Dhariwal, Neelakantan, Shyam, Sastry, Askell, Agarwal, Herbert-Voss, Krueger, Henighan, Child, Ramesh, Ziegler, Wu, Winter, Hesse, Chen, Sigler, Litwin, Gray, Chess, Clark, Berner, McCandlish, Radford, Sutskever, and Amodei]{brown2020languagemodelsfewshotlearners}
Brown, T.~B., Mann, B., Ryder, N., Subbiah, M., Kaplan, J., Dhariwal, P., Neelakantan, A., Shyam, P., Sastry, G., Askell, A., Agarwal, S., Herbert-Voss, A., Krueger, G., Henighan, T., Child, R., Ramesh, A., Ziegler, D.~M., Wu, J., Winter, C., Hesse, C., Chen, M., Sigler, E., Litwin, M., Gray, S., Chess, B., Clark, J., Berner, C., McCandlish, S., Radford, A., Sutskever, I., and Amodei, D.
\newblock Language models are few-shot learners, 2020.
\newblock URL \url{https://papers.nips.cc/paper/2020/hash/1457c0d6bfcb4967418bfb8ac142f64a-Abstract.html}.

\bibitem[Delétang et~al.(2023)Delétang, Ruoss, Grau-Moya, Genewein, Wenliang, Catt, Cundy, Hutter, Legg, Veness, and Ortega]{delétang2023neuralnetworkschomskyhierarchy}
Delétang, G., Ruoss, A., Grau-Moya, J., Genewein, T., Wenliang, L.~K., Catt, E., Cundy, C., Hutter, M., Legg, S., Veness, J., and Ortega, P.~A.
\newblock Neural networks and the chomsky hierarchy, 2023.
\newblock URL \url{https://openreview.net/pdf?id=WbxHAzkeQcn}.

\bibitem[Freiwald(2014)]{freiwald}
Freiwald, R.~C.
\newblock An introduction to set theory and topology, 2014.
\newblock URL \url{https://openscholarship.wustl.edu/books/20/}.

\bibitem[Gold(1967)]{gold}
Gold, E.~M.
\newblock Language identification in the limit, 1967.
\newblock URL \url{https://www.sciencedirect.com/science/article/pii/S0019995867911655}.

\bibitem[Hopcroft \& Ullman(1979)Hopcroft and Ullman]{hopcroft-ullman}
Hopcroft, J. and Ullman, J.
\newblock Introduction to automata theory, languages, and compuation, 1979.

\bibitem[Huang et~al.(2024)Huang, Yang, Bhattamishra, Sarrof, Krebs, Zhou, Nakkiran, and Hahn]{huang2024formalframeworkunderstandinglength}
Huang, X., Yang, A., Bhattamishra, S., Sarrof, Y., Krebs, A., Zhou, H., Nakkiran, P., and Hahn, M.
\newblock A formal framework for understanding length generalization in transformers, 2024.
\newblock URL \url{https://openreview.net/forum?id=U49N5V51rU}.

\bibitem[Jelassi et~al.(2023)Jelassi, d'Ascoli, Domingo-Enrich, Wu, Li, and Charton]{jelassi2023lengthgeneralizationarithmetictransformers}
Jelassi, S., d'Ascoli, S., Domingo-Enrich, C., Wu, Y., Li, Y., and Charton, F.
\newblock Length generalization in arithmetic transformers, 2023.
\newblock URL \url{https://arxiv.org/abs/2306.15400}.

\bibitem[Li \& Vitányi(2008)Li and Vitányi]{k-complexity}
Li, M. and Vitányi, P.
\newblock An introduction to kolmogorov complexity and its applications, 2008.
\newblock URL \url{https://link.springer.com/book/10.1007/978-0-387-49820-1}.

\bibitem[Mahankali et~al.(2023)Mahankali, Haochen, Dong, Glasgow, and Ma]{mahankali2023ntkvanillagradientdescent}
Mahankali, A., Haochen, J.~Z., Dong, K., Glasgow, M., and Ma, T.
\newblock Beyond ntk with vanilla gradient descent: A mean-field analysis of neural networks with polynomial width, samples, and time, 2023.
\newblock URL \url{https://openreview.net/forum?id=Y2hnMZvVDm&noteId=SzifoxR0by}.

\bibitem[Marsden et~al.(2024)Marsden, Dogariu, Agarwal, Chen, Suo, and Hazan]{marsden2024provablelengthgeneralizationsequence}
Marsden, A., Dogariu, E., Agarwal, N., Chen, X., Suo, D., and Hazan, E.
\newblock Provable length generalization in sequence prediction via spectral filtering, 2024.
\newblock URL \url{https://arxiv.org/abs/2411.01035}.

\bibitem[Nogueira et~al.(2021)Nogueira, Jiang, and Lin]{nogueira2021investigatinglimitationstransformerssimple}
Nogueira, R., Jiang, Z., and Lin, J.
\newblock Investigating the limitations of transformers with simple arithmetic tasks, 2021.
\newblock URL \url{https://arxiv.org/abs/2102.13019}.

\bibitem[Nye et~al.(2021)Nye, Andreassen, Gur-Ari, Michalewski, Austin, Bieber, Dohan, Lewkowycz, Bosma, Luan, Sutton, and Odena]{nye2021workscratchpadsintermediatecomputation}
Nye, M., Andreassen, A.~J., Gur-Ari, G., Michalewski, H., Austin, J., Bieber, D., Dohan, D., Lewkowycz, A., Bosma, M., Luan, D., Sutton, C., and Odena, A.
\newblock Show your work: Scratchpads for intermediate computation with language models, 2021.
\newblock URL \url{https://arxiv.org/abs/2112.00114}.

\bibitem[Pitt \& Warmuth(1993)Pitt and Warmuth]{min-interpolator-np-complete}
Pitt, L. and Warmuth, M.~K.
\newblock The minimum consistent df a problem cannot be approximated within any polynomial, 1993.
\newblock URL \url{https://dl.acm.org/doi/pdf/10.1145/138027.138042}.

\bibitem[Rissanen(1978)]{Rissanen1978ModelingBS}
Rissanen, J.
\newblock Modeling by shortest data description*.
\newblock \emph{Autom.}, 14:\penalty0 465--471, 1978.
\newblock URL \url{https://api.semanticscholar.org/CorpusID:30140639}.

\bibitem[Rockafellar(1970)]{rockafellar}
Rockafellar, R.~T.
\newblock Convex analysis, 1970.

\bibitem[Ruoss et~al.(2023)Ruoss, Delétang, Genewein, Grau-Moya, Csordás, Bennani, Legg, and Veness]{ruoss2023randomizedpositionalencodingsboost}
Ruoss, A., Delétang, G., Genewein, T., Grau-Moya, J., Csordás, R., Bennani, M., Legg, S., and Veness, J.
\newblock Randomized positional encodings boost length generalization of transformers, 2023.
\newblock URL \url{https://aclanthology.org/2023.acl-short.161/}.

\bibitem[Shaw et~al.(2021)Shaw, Chang, Pasupat, and Toutanova]{shaw2021compositionalgeneralizationnaturallanguage}
Shaw, P., Chang, M.-W., Pasupat, P., and Toutanova, K.
\newblock Compositional generalization and natural language variation: Can a semantic parsing approach handle both?, 2021.
\newblock URL \url{https://aclanthology.org/2021.acl-long.75/}.

\bibitem[Shaw et~al.(2024)Shaw, Cohan, Eisenstein, Lee, Berant, and Toutanova]{shaw2024altacompilerbasedanalysistransformers}
Shaw, P., Cohan, J., Eisenstein, J., Lee, K., Berant, J., and Toutanova, K.
\newblock Alta: Compiler-based analysis of transformers, 2024.
\newblock URL \url{https://openreview.net/forum?id=h751wl9xiR}.

\bibitem[Solomonoff(1964)]{solomonoff}
Solomonoff, R.~J.
\newblock A formal theory of inductive inference. part i, 1964.
\newblock URL \url{https://www.sciencedirect.com/science/article/pii/S0019995867911655}.

\bibitem[Vaswani et~al.(2017)Vaswani, Shazeer, Parmar, Uszkoreit, Jones, Gomez, Kaiser, and Polosukhin]{vaswani2023attentionneed}
Vaswani, A., Shazeer, N., Parmar, N., Uszkoreit, J., Jones, L., Gomez, A.~N., Kaiser, L., and Polosukhin, I.
\newblock Attention is all you need, 2017.
\newblock URL \url{https://proceedings.neurips.cc/paper_files/paper/2017/file/3f5ee243547dee91fbd053c1c4a845aa-Paper.pdf}.

\bibitem[Weiss et~al.(2021)Weiss, Goldberg, and Yahav]{weiss2021thinkingliketransformers}
Weiss, G., Goldberg, Y., and Yahav, E.
\newblock Thinking like transformers, 2021.
\newblock URL \url{https://proceedings.mlr.press/v139/weiss21a/weiss21a.pdf}.

\bibitem[Yang \& Chiang(2024)Yang and Chiang]{yang2024countingliketransformerscompiling}
Yang, A. and Chiang, D.
\newblock Counting like transformers: Compiling temporal counting logic into softmax transformers, 2024.
\newblock URL \url{https://openreview.net/forum?id=FmhPg4UJ9K#discussion}.

\bibitem[Zhou et~al.(2023)Zhou, Bradley, Littwin, Razin, Saremi, Susskind, Bengio, and Nakkiran]{zhou2023algorithmstransformerslearnstudy}
Zhou, H., Bradley, A., Littwin, E., Razin, N., Saremi, O., Susskind, J., Bengio, S., and Nakkiran, P.
\newblock What algorithms can transformers learn? a study in length generalization, 2023.
\newblock URL \url{https://openreview.net/forum?id=AssIuHnmHX}.

\bibitem[Zhou et~al.(2024)Zhou, Alon, Chen, Wang, Agarwal, and Zhou]{zhou2024transformersachievelengthgeneralization}
Zhou, Y., Alon, U., Chen, X., Wang, X., Agarwal, R., and Zhou, D.
\newblock Transformers can achieve length generalization but not robustly, 2024.
\newblock URL \url{https://openreview.net/pdf?id=DWkWIh3vFJ}.

\end{thebibliography}
